\documentclass[pdflatex,sn-mathphys-num]{sn-jnl}
\usepackage{graphicx}%
\usepackage{multirow}%
\usepackage{amsmath,amssymb,amsfonts}%
\usepackage{amsthm}%
\usepackage{mathrsfs}%
\usepackage[title]{appendix}%
\usepackage{xcolor}%
\usepackage{textcomp}%
\usepackage{manyfoot}%
\usepackage{booktabs}%
\usepackage{algorithm}%
\usepackage{algorithmicx}%
\usepackage{algpseudocode}%
\usepackage{listings}%
\usepackage{bm}
\usepackage{caption}

\theoremstyle{thmstyleone}%
\theoremstyle{thmstyletwo}%

\theoremstyle{thmstylethree}%

\begin{document}
    \title{\begin{centering}
Chemical Substructure Representations for Hierarchical Learning in Metal-Organic Frameworks\end{centering}}

\author[1,2]{\fnm{Micha\l} \sur{Sanocki}}

\author*[1,2]{\fnm{Julija} \sur{Zavadlav}}\email{julija.zavadlav@tum.de}

\affil[1]{\orgdiv{Multiscale Modeling of Fluid Materials, Department of Engineering Physics and
Computation}, \orgname{TUM School of Engineering and Design, Technical University of Munich}, \country{Germany}}

\affil[2]{\orgdiv{Atomistic Modeling Center, Munich Data Science Institute}, \orgname{Technical University of
Munich}, \country{Germany}}
\date{\vspace{20pt} \today}

\maketitle

\section{Abstract}


Machine-learning models for metal-organic frameworks operate primarily at the atomic level, often overlooking their layered chemical and structural organization, spanning atoms and functional groups, organic linkers, inorganic nodes, and the three-dimensional porous framework.  Here, we introduce hierMOF, a framework that captures this hierarchy by decomposing MOFs into their constituent building blocks and learning transferable representations at each structural level. These pretrained representations are integrated with atomic information through message passing and cross-level information exchange to predict MOF properties. Across band-gap, gas adsorption, and water stability, hierMOF achieves state-of-the-art predictive performance, consistently outperforming the corresponding atom-only baselines and reducing prediction errors by up to 55\%. Furthermore, our explainable AI approach attributes model predictions to individual chemical building blocks, providing insights into their contributions to MOF properties. Our results establish chemically informed hierarchical learning as a promising approach for accurate, data-efficient, and interpretable property prediction for the design of porous materials and computational materials discovery. 

\section{Introduction}

Metal-organic frameworks (MOFs) are porous, crystalline materials composed of nodes connected by organic linkers \cite{EXP, Burner2020-xp}. Their modular structure enables broad control over pore geometry, topology, and chemical functionality, making them promising materials for gas separation, gas storage, catalysis, and many other applications~\cite{EXP, Burner2020-xp,li2024advances,jia2022progress,bavykina2020metal}. At the same time, this tunability creates a vast design space through the many possible combinations of metal nodes, organic linkers, topologies, and functional groups~\cite{gibaldi2025mosaec}. Since this space cannot be explored through empirical methods alone \cite{ARCMOF, D3SC05612K, Sriram2024-sf}, efficient property-prediction methods are needed to guide MOF design.

Recently, machine learning has become an increasingly important tool for MOF property prediction \cite{D3SC05612K,Cao2023-pm, Niyongabo_Rubungo2025-bb, Nandy2022-ud, https://doi.org/10.1002/smll.202405087, https://doi.org/10.1002/anie.202200242,Batra2020-im, han2026multimodaltransformersampleawareprediction, D4SC03609C,Park2023-tf,Sezgin2026-qk,Ozcan2025-dk,Pruyn2025-rr}, as it allows for screening of large MOF databases at low computational cost. Early approaches relied on models based on predefined descriptors~\cite{Shi2023-uv,Li2023-pb}. However, such representations may not fully capture spatial configurations and interatomic interactions~\cite{Li2025-oh}. Thus, graph neural networks were subsequently adopted as they can learn structural representations directly from atomic connectivity and local chemical environments~\cite{PhysRevLett.120.145301,Li2025-oh}. More recently, large language model-based approaches~\cite{Pruyn2025-rr,Niyongabo_Rubungo2025-bb,Badrinarayanan2025-zr} and dedicated transformer architectures, including MOFormer~\cite{Cao2023-pm} and MOFTransformer~\cite{Kang2023-zy}, have been introduced. However, these high-capacity models often require large pretraining datasets to achieve accurate predictions \cite{Cao2023-pm, han2026multimodaltransformersampleawareprediction}, and often perform worse than much simpler graph-based models \cite{Cao2023-pm}.

Several MOF-specific datasets have supported data-driven modeling, including QMOF, hMOF, CoRE MOF, MOSAEC-DB, MOFX-DB, WS24, and the CSD MOF subset~\cite{QMOF,Wilmer2012-dy,CoreMOF,gibaldi2025mosaec,Bobbitt2023-mz,Moghadam2017-ab, Terrones2024-ef}. Despite the availability of more than 100,000 reported MOF structures~\cite{Moghadam2017-ab}, only a fraction are associated with consistent property labels. The three datasets considered in this work illustrate how the availability of labeled MOF data varies across target properties: WS24~\cite{Terrones2024-ef} contains approximately 1,000 MOFs with experimentally derived water-stability labels, QMOF~\cite{QMOF} provides DFT-calculated band gaps for approximately 20,000 structures, and hMOF~\cite{Wilmer2012-dy} includes approximately 138,000 hypothetical frameworks with simulated gas-adsorption properties. In addition to covering different properties and dataset sizes, QMOF and hMOF have been used in previous machine-learning studies \cite{Choudhary2021-rn,10.1039/d4dd00014e, Kang2023-zy}, enabling direct comparisons with established architectures. The uneven availability of labeled data also motivates the development of data-efficient models that can exploit different datasets. One way to achieve this is to leverage larger organic molecular datasets, such as QM9 and ZINC~\cite{Ramakrishnan2014-rd,Tingle2023-qk,10.1021/acscentsci.7b00572}, to learn transferable chemical representations. Since MOF linkers share chemical substructures with organic molecules, these datasets can be used to pretrain transferable representations of recurring chemical components. This could reduce the need for downstream models to learn chemical features directly from limited MOF-labeled data, allowing them to focus on how nodes, linkers, and motifs are organized within the framework and relate to the target property. However, such approaches are still underexplored.

Hierarchical graph neural networks are ideally suited to exploit such knowledge transfer by representing structures at multiple resolutions. Rather than operating only on atoms, they group lower-level features into larger chemical or structural units and exchange information within and across these levels~\cite{Zang2023-iv,Ying2018DiffPool}. This could allow models to combine local chemical environments with higher-level structural organization, extend their receptive field, and shorten message-passing paths between distant regions~\cite{Rampasek2021Hierarchical}. Although this coarse-graining removes some explicit atomistic detail, learned latent representations can retain chemical and structural information from the underlying atomic environments~\cite{Wang2019-on}, while improving data efficiency by encoding recurring chemical units ~\cite{Han2025-gv,Wan2025-km}. Thus, this approach is particularly suitable for MOFs because: 1) their structures are inherently hierarchical~\cite{Feng2020-vq}, 2) they often contain recurring chemical motifs~\cite{CoreMOF}, and 3) explicit representation of different levels may help the model learn both local and global structural features. Multiscale representations have shown promise for MOF modeling by capturing interactions across different spatial scales~\cite{Li2025-oh, Cui2023-ta}. However, previous efforts used fixed distance thresholds rather than chemically defined substructures for scale separation, and did not leverage pretrained structural representations.

Here, we introduce hierMOF, a unified framework for hierarchical MOF representation, property prediction, and interpretation, consisting of four components: structural segmentation, representation learning, property prediction, and hierarchical attribution (Fig.~\ref{fig:fig1}). First, we decompose MOF structures into nodes and organic linkers (hierMOF-seg), with linker atoms further assigned to chemical motifs. Unlike existing string-based MOF representations, we construct a non-overlapping segmentation into chemically defined substructures with explicit atom-level correspondence to the original MOF. Next, hierMOF-enc maps linkers and chemical motifs into latent spaces using a hierarchical attention-based graph autoencoder and a separate graph autoencoder for nodes. We then combine the segmented structure with pretrained representations to predict MOF properties using a geometrical hierarchical graph neural network. Finally, hierMOF-int exploits the same chemically defined hierarchy to attribute model predictions directly to motif, linker, and node representations, enabling interpretation at the level of chemically meaningful framework components. The framework enables transfer of information from larger chemical spaces to MOFs, reducing the amount of labeled data required for downstream tasks while retaining a direct connection between learned representations and the underlying chemical structure. We evaluate the framework on band-gap, gas-adsorption, and water-stability prediction, achieving state-of-the-art performance on all three tasks.

\section{Results}

\begin{figure*}[ht]
    \centering
    \includegraphics[width=0.99\linewidth]{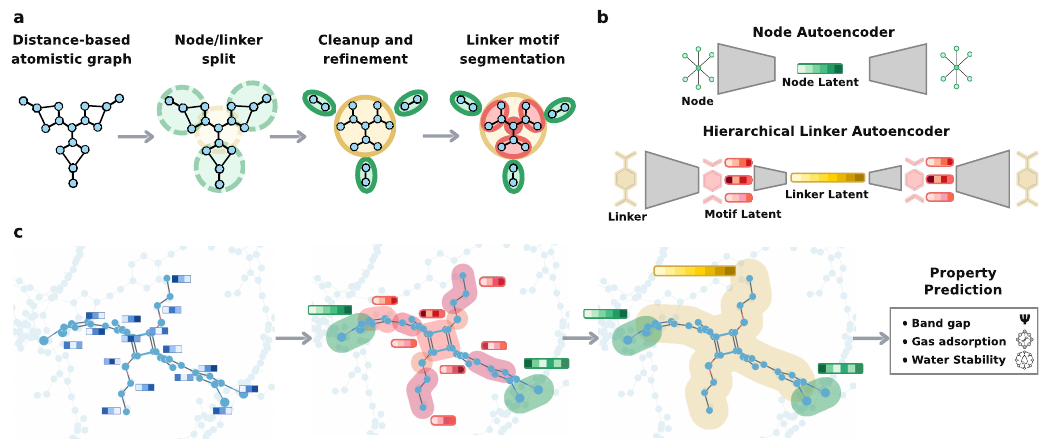}
    \caption{
        \textbf{hierMOF workflow.}
        \textbf{a} hierMOF-seg constructs a periodic atomistic graph, partitions the structure into inorganic nodes and organic linkers, refines the assignments, and further decomposes linkers into chemically defined motifs.
        \textbf{b} hierMOF-enc learns pretrained representations using a separate node autoencoder and a hierarchical linker autoencoder that produces both linker- and motif-level latent representations.
        \textbf{c} hierMOF combines atomic features with pretrained substructure representations through hierarchical message passing across atomistic, motif-node, and linker-node resolutions for downstream prediction of band gaps, gas adsorption, and water stability.
    }
    \label{fig:fig1}
\end{figure*}

The hierMOF workflow (Fig.~\ref{fig:fig1}) combines chemically defined MOF segmentation, pretraining of substructure representations, and hierarchical graph neural networks for property prediction. We examine these components in turn, followed by an evaluation of predictive performance and an analysis of learned structure-property relationships through substructure-level attribution.

\paragraph{hierMOF-seg: Segmentation of MOFs}

Although existing approaches, such as MOFid~\cite{Bucior2019-ps} and MOFSynth~\cite{Livas2024-su}, can identify molecular components within MOFs, they do not establish atom-level correspondence between these components and the original structure. Such correspondence is essential for integrating chemical substructure representations with atomic-level information, developing hierarchical architectures, and attributing model predictions to specific structural components. To address this limitation, we introduce hierMOF-seg, which explicitly maps identified linkers, nodes, and motifs to their constituent atoms in the input MOF. We represent each MOF as a periodic atomistic graph and partition it into node fragments and organic linker fragments using chemical priors and local connectivity (Fig.~\ref{fig:fig1}a and Supplementary Information Fig.~S1). Metal atoms and their directly connected neighbors are first assigned to nodes. Connected components in the remaining subgraph are then labeled as organic linkers, while isolated atoms and water- or hydroxide-like fragments are reassigned to the nearest node using distance-based rules. Following spatial merging of node candidates, disconnected groups are split into connected node fragments. Because bond orders in coordination environments are often ambiguous, we represent nodes using connectivity alone, while inferring bond orders for the organic linker fragments. Finally, each motif, linker, and node is assigned a center of mass for future neighbor list construction. Detailed bonding criteria, fragment reassignment rules, and atom mapping are provided in the Supplementary Information.

Each linker is first unwrapped across the periodic boundary and then reconstructed as an RDKit molecular graph~\cite{Landrum2016RDKit2025093}. Bond orders and formal charges are inferred from the three-dimensional geometry, using an approach developed by \citeauthor{https://doi.org/10.1002/bkcs.10334}~\cite{https://doi.org/10.1002/bkcs.10334} implemented in RDKit ~\cite{Landrum2016RDKit2025093}. Linkers whose reconstructed graphs pass valence, sanitization, and connectivity checks are then partitioned into motifs, which we define as small, recurring molecular graph substructures that do not necessarily correspond to complete functional groups. To retain finer structural detail, multi-ring systems are divided into ring-derived segments and shared junctions rather than treated as a single larger motif, thereby simplifying neighborhood construction (see Supplementary Information Fig.~S2). We assign motifs based on local bonding environments, while remaining atoms receive alkyl- or element-based labels. We use hydrogen atoms during molecular reconstruction but exclude them from the final motif assignments. It is worth highlighting that other motif-based architectures differ in how they define motifs. For example, MotiL \cite{Liu2025-hn} allows overlapping motifs, while HierVAE~\cite{pmlr-v119-jin20a} uses a frequency-dependent motif vocabulary. In contrast, hierMOF-seg defines motifs independently of their frequency of occurrence, enabling consistent segmentation and straightforward combination of different datasets. Its non-overlapping partition also simplifies the hierarchical architecture by assigning each atom to a single motif representation, avoiding multiple memberships during information aggregation. Examples of linker motif assignments and their chemical ambiguities are discussed in Supplementary Information Fig.~S8.

\begin{figure*}[ht]
    \centering
    \includegraphics[width=0.99\linewidth]{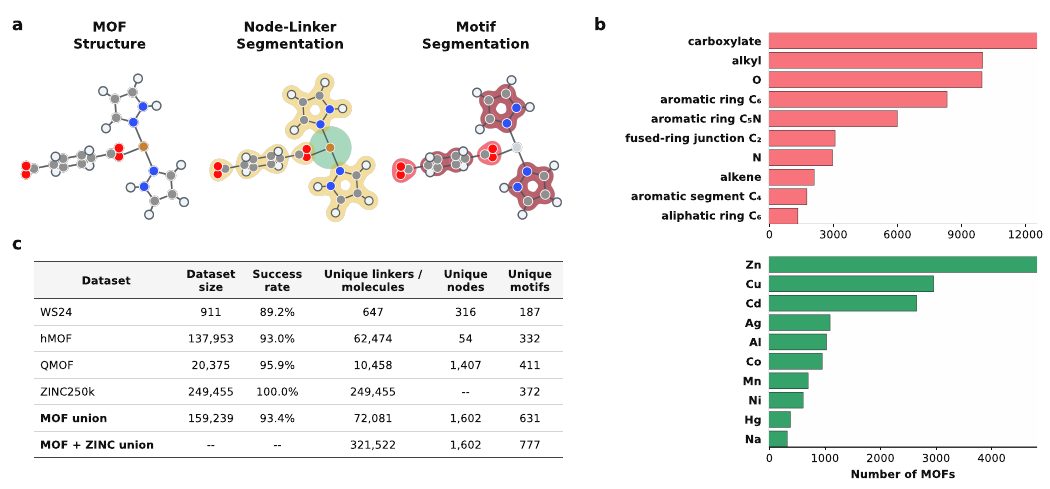}
    \caption{
        \textbf{hierMOF-seg decomposition and dataset statistics.}
        \textbf{a} Representative MOF structure illustrating its decomposition into inorganic nodes and organic linkers, followed by segmentation of the linkers into chemically defined motifs.
        \textbf{b} Distribution of the ten most frequent motif types (top) and metal elements (bottom) identified across the processed MOF datasets.
        \textbf{c} Dataset statistics for WS24, hMOF, QMOF, and ZINC250k, including dataset size, segmentation success rate, and numbers of unique linkers or molecules, nodes, and motifs. The final rows summarize the combined MOF datasets and their union with ZINC250k. Dataset size and success rate are omitted for this union because it combines MOF structures and individual molecules. ZINC250k contains molecules with predefined bond connectivity and therefore does not require geometry-based molecular reconstruction. 
    }
    \label{fig:MOFSEG}
\end{figure*}

Across MOF datasets (WS24, hMOF, and QMOF~\cite{Terrones2024-ef,Wilmer2012-dy,QMOF}), hierMOF-seg achieved an overall segmentation success rate of 93.4\%.  Successful segmentation required identifying nodes and linkers and reconstructing chemically valid linker molecular graphs for subsequent motif assignment. Failures primarily arose from missing or ambiguous bonds and problematic input geometries (Supplementary Information Table~S4). In contrast, a 100\% success rate was achieved on the ZINC250k dataset because its molecules have predefined bond connectivity, eliminating the need for geometry-based molecular reconstruction. On QMOF, hierMOF-seg improved the reported processing success rate by 8.0 percentage points compared with MOFSynth (from 87.9\% to 95.9\%)~\cite{Livas2024-su}. This difference may reflect MOFSynth's reliance on MOFid for linker fragmentation and subsequent SMILES identification, both of which are reported sources of failure, whereas hierMOF-seg uses its own periodic, graph-based segmentation and linker-reconstruction procedure. The identified linker and motif distributions (Fig.~\ref{fig:MOFSEG}b,c and Supplementary Information Figs.~S6 and~S7) are broadly consistent with previous analyses of common MOF building blocks~\cite{Bucior2019-ps}. The lower success rate for WS24 may reflect its diverse set of structures or issues with structure preprocessing, whereas the high QMOF success rate reflects the structural filtering applied during its construction. The small number of node records obtained for hMOF~\cite{Wilmer2012-dy} is expected from its generation using a restricted building-block library, while the broader chemistry of QMOF produced substantially greater node diversity.

\paragraph{hierMOF-enc: Encoding of MOF substructures}

In the second stage, we develop hierMOF-enc, comprising a property-regularized hierarchical autoencoder for organic linkers and a separate graph autoencoder for nodes (Fig.~\ref{fig:fig1}b). The models do not share parameters, latent spaces, decoders, or reconstruction objectives because they encode distinct chemical domains, enabling us to exploit diverse datasets. Although encoder-derived embeddings have been used for molecular property prediction~\cite{Han2025-gv,Zang2023-iv,nayak2020transformerbasedmoleculeencoding}, our approach pretrains representations on separate organic-linker and node chemical spaces and transfers them to MOF property prediction. The hierarchical linker autoencoder further provides both linker- and motif-level representations, mitigating the loss of local chemical information during coarse-graining for larger linkers such as those in Supplementary Information Figs.~S2 and ~S8. We show this is relevant for several properties, such as gas adsorption, which depends on framework-scale properties like pore geometry and accessible volume, as well as local binding interactions involving linker motifs and metal sites~\cite{Moreau2017-gy,Cui2023-ta}. During pretraining, the linker and node decoders reconstruct motif identities and attachment relationships, and elemental composition and connectivity, respectively. The linker representations are additionally regularized through property-prediction and contrastive objectives. Further architectural details and training objectives are provided in Methods Section \ref{sec:Autoncoders}.

\begin{figure*}[ht]
    \centering
    \includegraphics[width=0.99\linewidth]{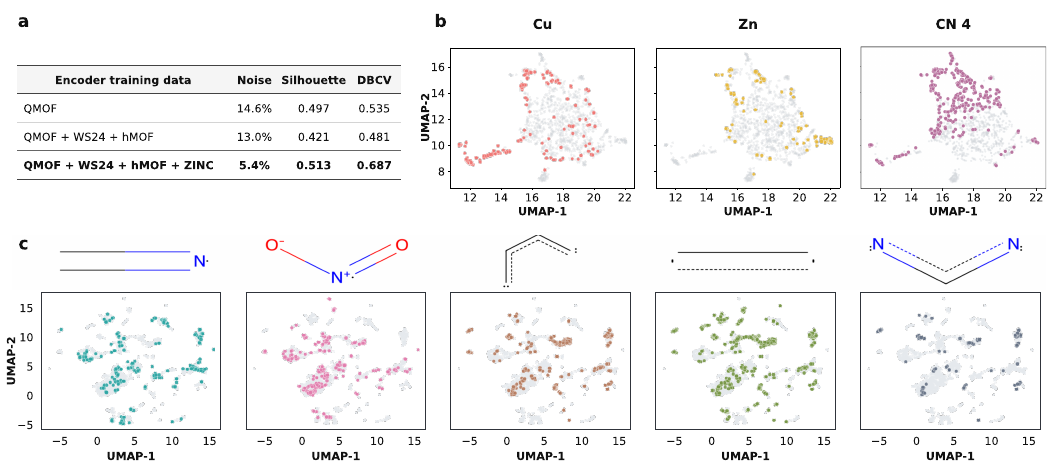}
    \caption{
        \textbf{hierMOF-enc latent representation visualization and analysis.}
        \textbf{a} Clustering quality of QMOF linker representations obtained using three encoder training datasets: QMOF, QMOF combined with WS24 and hMOF, and the combined dataset augmented with ZINC250k. Metrics include the fraction of points classified as noise, silhouette score, and density-based clustering validation (DBCV) score.
        \textbf{b} UMAP visualization of node latents, highlighting Cu-containing nodes, Zn-containing nodes, and nodes with coordination number four (CN 4). Remaining node embeddings are shown in grey.
        \textbf{c} UMAP visualization of linker latents obtained using the ZINC-inclusive encoder. Colors highlight linkers containing five representative hierMOF-seg motifs, while the remaining linker embeddings are shown in grey.
    }
    \label{fig:MOFENC}
\end{figure*}

We trained the node encoder on unique nodes identified by hierMOF-seg in the combined QMOF, WS24, and hMOF datasets. To investigate the effect of linker-training composition, we trained three linker encoders with the same architecture on progressively broader chemical spaces: 1) QMOF alone, 2) QMOF, WS24, and hMOF, and 3) these three datasets augmented with ZINC250k~\cite{10.1021/acscentsci.7b00572,10.1021/ci3001277}. We examined the resulting representations using Uniform Manifold Approximation and Projection (UMAP)~\cite{mcinnes2020umapuniformmanifoldapproximation}, followed by HDBSCAN density-based clustering, which identifies groups of similar embeddings and classifies unassigned points as noise~\cite{10.1007/978-3-642-37456-2_14}. Clustering quality was assessed using the noise fraction, silhouette score, and density-based clustering validation (DBCV), which measures cluster quality based on density separation (for details, see Supplementary Information Table~S7). Comparing the same QMOF linkers across the three encoders reveals a decreasing fraction of clustering noise with broader pretraining, with the ZINC-inclusive encoder achieving the highest silhouette and DBCV scores (Fig.~\ref{fig:MOFENC}a). These results show that extending pretraining to more diverse chemical spaces improves the organization of the learned representations.

Cu- and Zn-containing nodes occupy partly distinct regions of the latent space, while nodes with coordination number four overlap with both populations (Fig.~\ref{fig:MOFENC}b). This overlap is consistent with the occurrence of four-coordinate environments in both metals, whereas differences in elemental composition, connectivity, and other coordination environments explain their incomplete overlap. These patterns suggest that the learned representations capture both elemental identity and coordination topology. Likewise, linkers containing the highlighted hierMOF-seg motifs show characteristic distributions (Fig.~\ref{fig:MOFENC}c). Imine- and nitro-containing linkers occupy partly overlapping but distinct regions, whereas aromatic-segment and fused-junction motifs associated with multi-ring systems (see Supplementary Information Fig.~S2b) overlap more extensively. This overlap is consistent with their occurrence in related linkers and, frequently, within the same molecules. These trends are further supported by cluster-profile analyses, which associate node and linker representations with elemental composition and coordination numbers, and motif content, respectively (Supplementary Information Figs.~S10 and~S11). Beyond capturing chemical identity, the learned linker space also provides insight into dataset coverage. A joint UMAP of MOF-derived linkers and ZINC molecules reveals regions with limited WS24 representation, potentially reflecting underexplored chemistry or constraints associated with water stability (Supplementary Information Fig.~S12). Together, these patterns indicate that the global representations retain information associated with local chemical composition and substructure, supporting their use as chemically informed inputs for downstream property prediction with hierMOF.

\begin{figure*}[ht]
    \centering
    \includegraphics[width=0.99\linewidth]{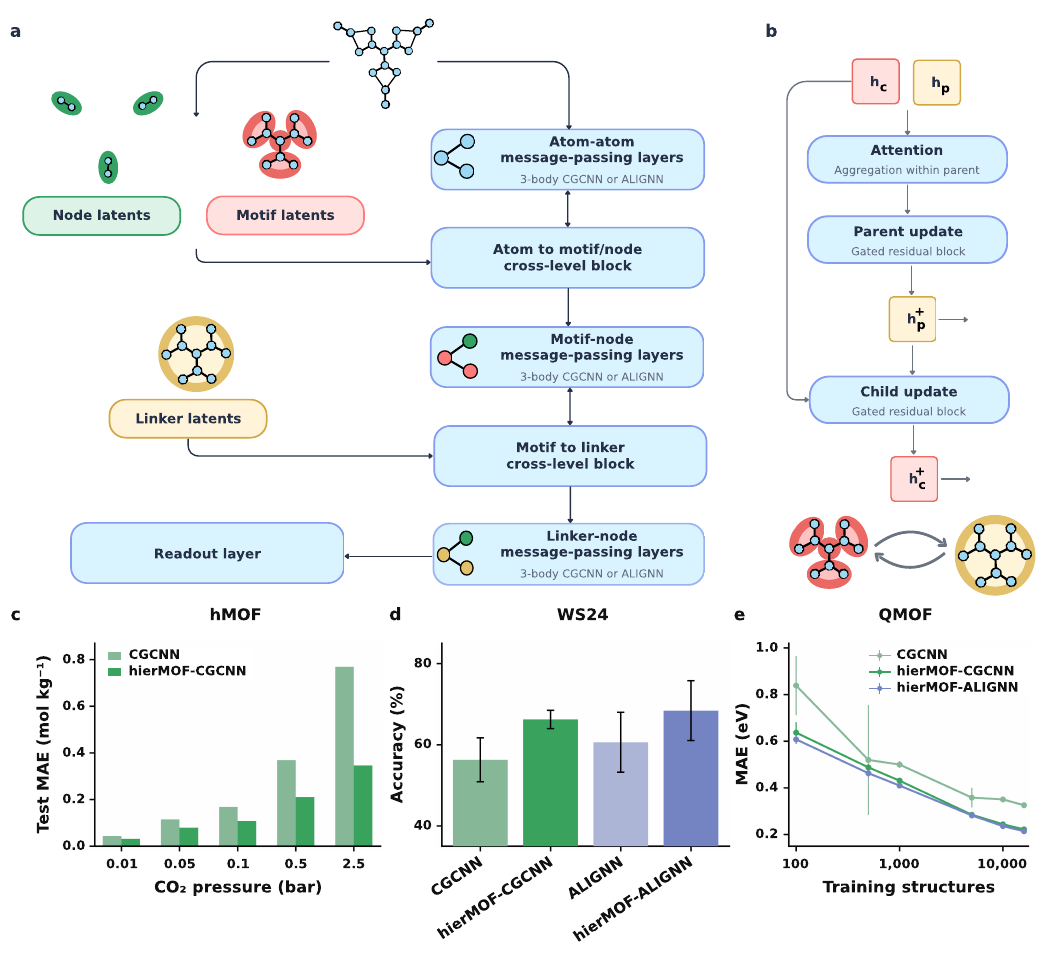}
    \caption{
        \textbf{hierMOF architecture and benchmarks.}
        \textbf{a} hierMOF information propagation through atomistic, motif-node, and linker-node message-passing layers, connected by cross-level blocks. 
        \textbf{b} Cross-level blocks with attention-based aggregation of child representations followed by gated residual updates of parent and child features.
        \textbf{c} Test mean absolute error (MAE) for CO$_2$ adsorption prediction at different pressures at 298~K for hierMOF-CGCNN and the CGCNN model reported by ~\citeauthor{Teng2025-lx}\cite{Teng2025-lx}.
        \textbf{d} Water-stability classification accuracy on the WS24 benchmark.
        \textbf{e} QMOF band-gap test MAE vs. training-set size for CGCNN, hierMOF-CGCNN, and hierMOF-ALIGNN. Error bars indicate standard deviation across three training runs.
    }
    \label{fig:MOFMOD}
\end{figure*}

\subsection{hierMOF: Modeling of MOFs with hierarchical representations}

Building on the learned MOF substructure representations, we developed hierMOF, a hierarchical graph neural network framework for property prediction based on chemically informed, multiscale MOF representations (Fig.~\ref{fig:MOFMOD}a,b). While hierarchical molecular GNNs typically combine atomic and motif representations to obtain a molecule-level representation~\cite{Zang2023-iv,pmlr-v119-jin20a}, hierMOF extends this hierarchy to four levels of structural organization, from atoms through motifs and complete linkers to the periodic MOF. This is achieved through three message-passing resolutions: (1) atoms, (2) motifs and nodes, and (3) linkers and nodes, followed by crystal-level readout. The hierarchy follows two coupled pathways: one progressively combines atoms into motifs and linkers, while the other maps atoms directly to node representations that participate at both higher resolutions. At the atomic level, vertices retain their original coordinates, while motif, linker, and node tokens are positioned at their centers of mass as determined by hierMOF-seg. Distance cutoffs then define spatial neighborhoods at each resolution, allowing interactions both within and between component classes. 

To incorporate the chemical information learned by hierMOF-enc, pretrained motif-, linker-, and node-level representations are projected into a shared hidden space and combined with learned class embeddings through gated fusion, while atomic states are initialized from element embeddings. The resulting representations are refined through hierarchical message passing, which integrates information both within and across structural resolutions. Importantly, this framework is independent of the underlying message-passing operator, allowing us to evaluate two variants: hierMOF-CGCNN, based on the Crystal Graph Convolutional Neural Network (CGCNN) with directed updates incorporating additional radial and three-body angular information~\cite{PhysRevLett.120.145301}, and hierMOF-ALIGNN, based on the Atomistic Line Graph Neural Network (ALIGNN), which employs alternating edge-gated graph and bond-angle line-graph convolutions~\cite{Choudhary2021-rn,CHOUDHARY2022111388}. Both variants share the same hierarchical tokenization, cross-level information exchange, and crystal-level readout, enabling us to assess the transferability of the hierarchical approach across different graph neural network architectures.

Information is exchanged bidirectionally between structural resolutions through attention-based aggregation and gated updates following the hierarchy defined by hierMOF-seg. Upward information transfer provides coarse representations with local chemical context, whereas downward transfer allows atom- and motif-level states to become conditioned on their larger linker or node environment. Propagation through higher-level representations also increases the effective receptive field of the model and enables information exchange between spatially separated regions without requiring larger cutoffs. Finally, motif, linker, and node representations are pooled separately and combined with a lattice descriptor derived from the volume-normalized Gram matrix of the unit-cell vectors, which encodes cell shape and relative dimensions, to predict crystal properties. The detailed architecture, information exchange operations, and readout are described in Methods Section \ref{HierGNN}. Unlike previous multiscale MOF models that primarily rely on distance-based structural resolutions~\cite{Li2025-oh,Cui2023-ta}, hierMOF combines chemically defined hierarchies, pretrained substructure representations, and bidirectional information exchange across periodic structural scales, enabling the model to capture both local chemical environments and their organization within the larger framework.


We evaluate hierMOF across three distinct property-prediction tasks, spanning different dataset sizes and chemical properties, to assess the benefits of hierarchical representations. We begin with gas-adsorption prediction using hMOF, the largest dataset considered in this work. Separate models were trained for individual adsorption targets spanning multiple gases, pressures, and temperatures. Across these conditions, hierMOF-CGCNN reduced prediction errors by approximately 28-55\% compared with previously reported atom-only CGCNN results~\cite{Teng2025-lx}, while hierMOF-ALIGNN achieved reductions of 15-23\% relative to available ALIGNN baselines (Fig.~\ref{fig:MOFMOD}c and Supplementary Information Table~S12). Notably, for CO$_2$ adsorption at 2.5~bar, hierMOF-CGCNN achieved a lower MAE using a subset of only 15,000 MOFs than the published CGCNN trained on the full hMOF dataset of approximately 138,000 structures, with errors of 0.499 and 0.769~mol kg$^{-1}$, respectively~\cite{Teng2025-lx} (Supplementary Information Table~S10). These substantial improvements suggest that information exchange between local chemical environments and larger framework components is beneficial even when extensive training data is available. Moreover, despite the generally stronger performance of conventional ALIGNN over CGCNN, both hierMOF variants achieved similar accuracy, suggesting that the hierarchical representations can compensate for differences in the underlying message-passing architectures.

Next, we test hierMOF on the WS24 dataset, which includes classification of MOFs into four classes: unstable, low kinetic stability, high kinetic stability, or thermodynamic stability. We adapt hierMOF for classification by replacing the scalar prediction output with a four-class head (for details, see Supplementary Information Table~S13). Both hierMOF variants improve classification accuracy over their atom-only baselines, with gains of approximately 10 and 8 percentage points for hierMOF-CGCNN and hierMOF-ALIGNN, respectively (Fig.~\ref{fig:MOFMOD}d). These results demonstrate the benefits of hierarchical representations even in data-limited settings, with hierMOF-ALIGNN achieving the highest overall accuracy.

Lastly, we evaluate hierMOF on predicting PBE band gaps for QMOF. On the full available training set, hierMOF-CGCNN achieves a test MAE of $0.224\pm0.004$~eV, a 31\% reduction compared with atom-only CGCNN ($0.325$~eV), while hierMOF-ALIGNN further reduces the MAE to $0.214\pm0.003$~eV (Supplementary Information Table~S11). The latter also outperforms previously reported ALIGNN and pretrained MOFTransformer results, both achieving an MAE of $0.224$~eV. Both hierMOF variants thus achieve state-of-the-art performance among the compared architectures, despite requiring substantially less extensive pretraining than models such as MOFTransformer~\cite{Kang2023-zy,Cao2023-pm}, which was pretrained on approximately one million porous-material structures. In contrast, hierMOF obtains high performance by transferring chemically relevant information at the level of nodes, motifs, and linkers. As hierMOF requires successful hierMOF-seg decomposition, the available dataset sizes are slightly reduced. Comparisons with published models are therefore approximate due to potential differences in dataset composition, except for CGCNN, for which we used our own implementation.

To further investigate data efficiency, we evaluate hierMOF across different QMOF training-set sizes. The scaling analysis in Fig.~\ref{fig:MOFMOD}e shows that hierarchical representations consistently improve prediction accuracy, with particularly pronounced gains in the low-data regime, consistent with our findings on WS24. This suggests that the chemically defined hierarchy provides a useful inductive bias, allowing the model to learn relevant structure-property relationships from fewer labeled MOFs than conventional atom-level message passing. Nevertheless, the improvements on the full dataset are smaller than those observed for gas adsorption, likely reflecting the greater importance of pore geometry and framework-scale organization for adsorption, whereas electronic properties depend more strongly on local atomic and coordination environments. These findings suggest that hierarchical representations are particularly beneficial for properties sensitive to the spatial organization and interactions of framework substructures.

To assess whether the observed improvements on the three datasets arise from the hierarchical representation rather than solely from pretrained features or the underlying message-passing backbone, we conducted a systematic ablation study on the full QMOF dataset and hMOF subset (see Supplementary Information Tables~S9 and~S10). For hMOF, hierMOF-CGCNN reduced the MAE by 55\% relative to the parameter-matched CGCNN, indicating that the improvement cannot be explained by model size alone. Notably, the QMOF ablations show that relying solely on pretrained node, motif, and linker representations, even without explicit atomic inputs or hierarchical information exchange, yielded better performance than atom-only CGCNN. Replacing pretrained hierMOF-enc representations with trainable embeddings led to the largest performance degradation among the ablations, although the hierarchical model still outperformed the atom-only CGCNN. The ablation results also show that removing the lattice descriptor or angular interactions further reduces accuracy, although their contributions are smaller than those of the hierarchical representation or pretrained features.

Interestingly, the pretraining ablations show that expanding hierMOF-enc training from QMOF linkers to those from all three MOF datasets (QMOF, WS24, and hMOF) did not improve performance, whereas incorporating ZINC molecules yielded the best results. This suggests that expanding the encoder training space beyond MOF linkers to a broader range of organic molecules can improve linker representations, highlighting the importance of chemical diversity for learning useful latent representations. Additionally, further ablations show that both motif- and linker-level representations contribute useful information. Together with the improvements obtained using both CGCNN- and ALIGNN-based variants, these results highlight the importance of hierarchical representations, complemented by chemical pretraining, in achieving state-of-the-art performance for MOF property prediction.

\subsection{hierMOF-int: Interpretable hierarchical substructure attribution}

An additional advantage of the hierMOF hierarchy is that its representations correspond directly to chemically identifiable MOF substructures. This enables attribution at the level of linkers, motifs, and nodes, rather than through global descriptors, spatial attention, or pore-level decompositions~\cite{Li2025-oh,10.1021/acs.jctc.6c00100}. Existing MOF interpretability methods based on these representations can provide useful structural insights but do not explicitly assign attribution scores to individual chemical building blocks. In particular, attention scores, which are often used for model interpretation, do not directly quantify a component's contribution to a prediction. Fragment-level attribution instead connects model predictions to chemically defined substructures and has been utilized to analyze structure-property relationships, diagnose model behavior under sparse or biased training data, and prioritize structural modifications in organic molecules~\cite{Wu2023,Jia2022,Roth2026}. hierMOF-int extends this interpretability approach to MOFs by exploiting the hierMOF hierarchy to attribute predictions to identifiable framework components.

To quantify the contributions of individual MOF substructures, hierMOF-int applies conditional Integrated Gradients (IG)~\cite{Sundararajan2017} to the pretrained hierMOF-enc representations entering hierMOF. For each linker, motif, or node, its representation is varied between the observed value and the corresponding hierarchy-level training mean, while all other model inputs remain fixed. Positive and negative attribution scores indicate whether the observed representation increases or decreases the predicted property relative to this reference. Because each substructure is evaluated conditionally within the original framework, the scores are not additive across substructures or hierarchy levels and reflect learned associations rather than causal effects.  Linker interventions retain the original motif inputs, while motif interventions retain the original parent-linker input. The mathematical formulation is provided in Methods Section~\ref{MOFINT}. To assess whether conditional IG provides a reliable description of model dependence, we validated its performance using several previously established criteria, including reproducibility across repeated runs, sensitivity to parameter randomization~\cite{Adebayo2018}, and numerical completeness \cite{Sundararajan2017} (details in Supplementary Information Tables~S14 and~S15). We subsequently apply conditional IG to the hierMOF-ALIGNN models trained on the QMOF and hMOF datasets, allowing their predictions to be examined directly in terms of linker, motif, and node contributions.

\begin{figure*}[ht]
    \centering
    \includegraphics[width=0.99\linewidth]{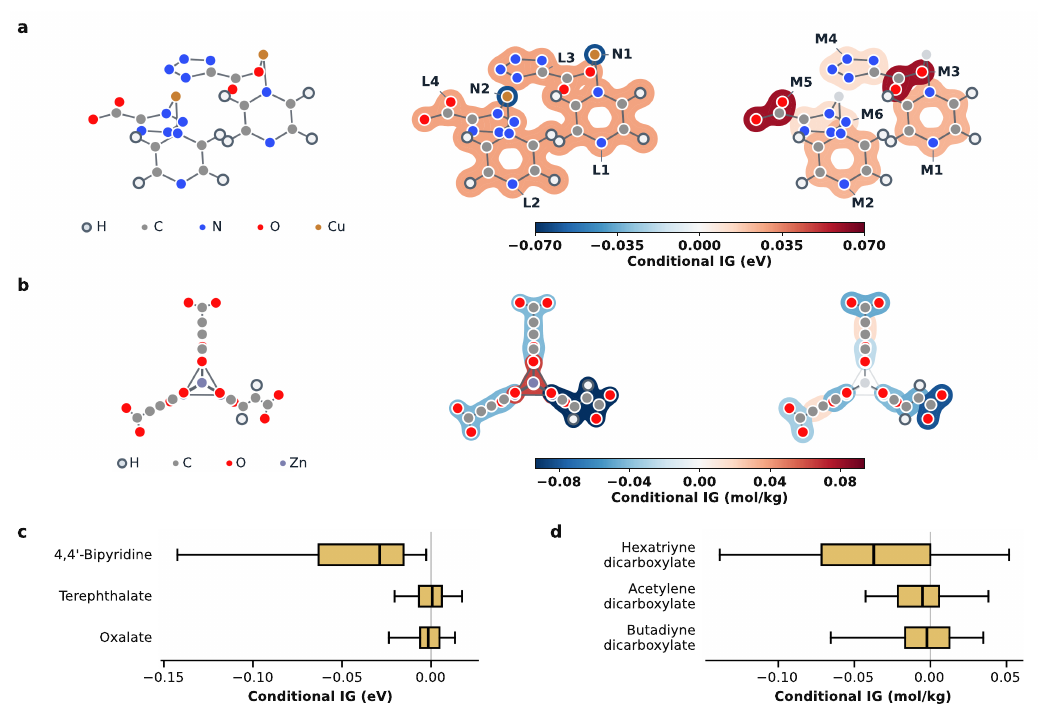}
    \caption{
        \textbf{Interpretation of hierMOF predictions using hierMOF-int.}
        \textbf{a,b)} Representative Cu-containing framework from QMOF (a) and Zn-containing framework from hMOF (b), with conditional IG attributions for nodes, linkers, and motifs in band-gap and gas adsorption prediction, respectively. Scores are expressed in the units of the predicted property, with red and blue indicating positive and negative conditional contributions relative to the reference representations.
        \textbf{c,d)} Distributions of linker-level conditional IG scores for selected linker types across QMOF band-gap predictions (c) and hMOF gas adsorption predictions (d). Boxes indicate the 25th–75th percentiles, central lines mark the median, and whiskers extend to the 10th–90th percentiles.
    }
    \label{fig:MOFINT}
\end{figure*}


The examples in Fig.~\ref{fig:MOFINT}a,b illustrate how linker, motif, and node representations contribute differently to predicted MOF properties. In the Cu-containing framework in Fig.~\ref{fig:MOFINT}a, the nodes have negative attributions of approximately $-0.07$ ~eV, indicating that they lower the predicted band gap relative to the corresponding training-mean reference. In contrast, motifs M3 and M5 have strong but positive contributions. Therefore, replacing Cu-node representations with their reference would increase the predicted band gap, whereas replacing motifs M3 or M5 would decrease it. A different pattern emerges for the Zn-containing framework in Fig.~\ref{fig:MOFINT}b, where linker and motif representations predominantly contribute negatively to the predicted gas adsorption, while the Zn-node contribution is positive. These examples illustrate that hierMOF-int distinguishes the contributions of different structural components to the predicted properties (for additional analysis and hierMOF-CGCNN scores, see Supplementary Information Figs.~S14 and~S15).

Across the test set, the attribution distributions reveal recurring associations for particular linker, motif, and node types. For QMOF, we find that linker representations show clear dataset-level trends, with 4,4'-bipyridine exhibiting predominantly negative conditional scores. This is consistent with electronic-structure studies showing that 4,4'-bipyridine can contribute directly to low-lying frontier states in semiconducting MOFs and, more generally, that linker chemistry can strongly influence MOF band gaps~\cite{Siemensmeyer2020,10.1021/ja405350u}. Complementary investigation of node and motif contributions on QMOF (Supplementary Information Fig.~S13) shows that node attributions vary widely across metal-containing groups, consistent with studies showing that changes in metal-node composition can modify MOF band gaps through changes in metal-dependent electronic states near the band edges, with the magnitude and direction of these effects depending on the surrounding framework~\cite{Ejegbavwo2020,Yan2023}. Motif contributions are generally smaller and more broadly distributed around zero.

Among the three most prevalent hMOF linkers, hexatriyne dicarboxylate exhibited the most negative median attribution for CO$_2$ uptake (Fig.~\ref{fig:MOFINT}d). Frameworks containing this linker also had the largest median pore-limiting diameter (5.25~\AA), potentially indicating weaker confinement and reduced overlap of adsorption potentials from opposing pore walls, which can lower CO$_2$ binding affinity. However, pore dimensions do not depend solely on linker length. Despite its longer backbone, butadiyne dicarboxylate occurred in frameworks with smaller median pore-limiting diameters than acetylenedicarboxylate (3.75 vs.\ 4.75~\AA), illustrating how framework topology and linker arrangement can influence pore accessibility and CO$_2$ adsorption. Local confinement may also explain why alkyl motifs showed higher attribution scores than alkyne motifs within the same linkers (Supplementary Information Fig.~S16 and Table~S16). Pore-facing alkyl groups may create tighter adsorption environments, increasing CO$_2$ interactions with the framework. Together, these findings highlight the interplay between linker chemistry and pore geometry and demonstrate how hierMOF-int can inform further investigation of structure-property relationships.

\section{Discussion}

Our results establish chemically defined hierarchical representations as a powerful inductive bias for MOF property prediction. Across three distinct benchmarks, hierMOF consistently outperforms conventional baseline models, achieving state-of-the-art predictive accuracy while demonstrating substantial improvements in data efficiency across different GNN architectures. By preserving the organization of atoms into motifs, linkers, and nodes, hierMOF enables information exchange across structural scales while incorporating chemical knowledge learned independently of the target properties. This approach can be viewed as a form of transfer learning, where pretrained substructure representations are transferred as features to downstream property-prediction models rather than transferring an entire pretrained model. Such representations can be reused across different frameworks and prediction tasks, reducing the dependence on task-specific labeled data. Importantly, the ablation results indicate that simply increasing the parameter count of atom-only architectures cannot reproduce the improvements, highlighting the importance of the hierarchical inductive bias alongside pretraining. Notably, hierMOF also improves QMOF band-gap prediction, where local atomic and coordination environments already provide substantial information, suggesting that framework-level organization contributes useful predictive information even for properties that are less influenced by the spatial arrangement of substructures. Beyond improving predictive accuracy, the explicit correspondence between learned representations and chemically identifiable MOF substructures enables hierMOF-int to provide insights into learned structure-property relationships.

Nevertheless, the hierMOF model could be further improved. Segmentation requires a discrete decomposition of the MOF, and ambiguous coordination environments or unusual bonding can lead to segmentation failures or imperfect assignments. The non-overlapping motif definitions provide consistency across datasets but represent only one possible chemical partition. Exploring more adaptive segmentation strategies could provide greater flexibility in defining chemically meaningful substructures, although preserving consistent substructure assignments across datasets would remain necessary. Furthermore, representing motifs, linkers, and nodes at their centers of mass compresses the internal geometry, while implicitly treating hydrogen atoms may limit the description of strongly directional interactions. Possible improvements include orientation- and shape-aware coarse representations, more robust treatment of coordination chemistry, and improved handling of directional interactions. Beyond these refinements, extending the hierarchy to incorporate pore geometry and framework topology as an additional structural level could enable hierMOF to explicitly capture pore connectivity and larger-scale geometric features relevant to adsorption and transport. Future work should also investigate transfer from more diverse organic molecules and metal coordination complexes, and evaluate generalization to chemically and topologically distinct MOFs.

The benefits demonstrated here suggest that hierarchical representations could also be applied to predicting other properties, including gas adsorption selectivity, molecular diffusion, and elastic moduli, which depend on the interplay between local interactions, pore accessibility, and framework organization~\cite{10.1021/acs.chemmater.3c01885, 10.1039/d1cs00558h}. Another promising direction is to extend hierMOF beyond property prediction toward the exploration and generation of novel MOF candidates, potentially leveraging hierMOF-int to guide structural modifications. Additionally, similar hierarchical organization is also present in covalent organic frameworks, porous organic polymers, and other materials, suggesting broader opportunities for transferable, chemically informed representation learning.

\section{Methods}


Duplicate molecules were merged prior to generating the combined dataset splits, and dataset-specific motif identifiers were remapped to a shared vocabulary based on their chemical labels. Further details on hierMOF-seg, including numerical bonding criteria, periodic unwrapping, bond-order and charge inference, motif rules, and failures, are provided in the Supplementary Information (Table~S4 and Fig.~S1). We performed all experiments using the deep learning framework JAX~\cite{jax2018github}.

\subsection{hierMOF-enc}
\label{sec:Autoncoders}


A linker \(L\) is initially represented as an atomic graph
\(
\mathcal G_L^{\mathrm a}
=
(\mathcal V_L^{\mathrm a},\mathcal E_L^{\mathrm a}),
\)
whose vertices (\(\mathcal V_L^{\mathrm a}\)) contain atomic numbers, hydrogen counts, and RDKit-derived descriptors (formal charge, valence, aromaticity, and ring membership), while edges (\(\mathcal E_L^{\mathrm a}\)) encode bond descriptors~\cite{Landrum2016RDKit2025093}. First, atom and bond features are embedded into latent representations, denoted by \(\mathbf h_i^{\mathrm a}\) and \(\mathbf h_{ij}^{\mathrm b}\), respectively, with the atomic representations subsequently refined through a message-passing network (MPN).

Each motif \(u\) comprises a set of atoms \(\mathcal A_u\subseteq\mathcal V_L^{\mathrm a}\), whose representations are aggregated through learned attention pooling, a weighted average with feature-dependent weights determined by a learnable scoring vector \(\mathbf w\), defined as:
\begin{equation}
\widetilde{\mathbf h}_u^{\mathrm m}
=
\operatorname{AttnPool}_{\mathbf w_{\mathrm a}}
\left(
\{\mathbf h_i^{\mathrm a}\}_{i\in\mathcal A_u}
\right),
\label{eq:mofenc_atom_to_motif}
\end{equation}
\begin{equation}
\operatorname{AttnPool}_{\mathbf w}
\left(\{\mathbf h_i\}_{i\in I}\right)
=
\sum_{i\in I}
\frac{\exp(\mathbf w^{\mathsf T}\mathbf h_i)}
{\sum_{j\in I}\exp(\mathbf w^{\mathsf T}\mathbf h_j)}
\mathbf h_i.
\label{eq:si_attention_pool}
\end{equation}
The motifs form a linker graph
\(
\mathcal G_L^{\mathrm m}
=
(\mathcal V_L^{\mathrm m},\mathcal E_L^{\mathrm m}),
\)
with one vertex per motif (\(\mathcal V_L^{\mathrm m}\))  and edges connecting motifs linked by atomic bonds (\(\mathcal E_L^{\mathrm m}\)). The pooled atom representations are combined with trainable motif-type embeddings and propagated over this graph,
\begin{equation}
\{\mathbf z_u^{\mathrm m}\}
=
\operatorname{MPN}_{\mathrm m}
\left(
\mathcal G_L^{\mathrm m},
\left\{
W_{\mathrm a\rightarrow\mathrm m}
\widetilde{\mathbf h}_u^{\mathrm m}
+
W_{\mathrm type}E_{\mathrm m}(u)
\right\}
\right).
\label{eq:mofenc_motif_encoding}
\end{equation}
Here, \(E_m(u)\) denotes the trainable embedding vector associated with the type of motif \(u\), while \(W_{\mathrm a\rightarrow\mathrm m}\) and \(W_{\mathrm type}\) are learned projection matrices. The resulting motif latents \(\mathbf z_u^{\mathrm m}\) serve as pretrained motif-level inputs to hierMOF for downstream property prediction.

The linker representation combines an attention-weighted motif summary with mean-pooled atom, motif, and bond states, where the latter are obtained from an auxiliary atom-bond message-passing branch. The resulting features are concatenated, normalized through layer normalization (LN), and projected into the linker latent using a multilayer perceptron (MLP):
\begin{equation}
\begin{aligned}
\mathbf h_L
&=
\operatorname{AttnPool}_{\mathbf w_{\mathrm m}}
\left(
\{\mathbf z_u^{\mathrm m}\}_{u\in\mathcal V_L^{\mathrm m}}
\right),\\
\mathbf z_L
&=
\operatorname{MLP}_{L}
\left(
    \operatorname{LN}
    \left(
    [\mathbf h_L,
    \overline{\mathbf h}_L^{\mathrm a},
    \overline{\mathbf h}_L^{\mathrm m},
    \overline{\mathbf h}_L^{\mathrm b}]
    \right)
\right),
\end{aligned}
\label{eq:mofenc_linker_latent}
\end{equation}
where the overbars denote mean pooling over the corresponding atom (\(\overline{\mathbf h}_L^{\mathrm a}\)), motif (\(\overline{\mathbf h}_L^{\mathrm m}\)), and bond (\(\overline{\mathbf h}_L^{\mathrm b}\)) states.

The decoder predicts motif identities and attachment relationships along a breadth-first traversal rooted at the largest motif. The root motif is predicted from the linker representation, while subsequent motif types are predicted using the current motif state and an attention-pooled summary of previously encountered motifs. Training uses teacher forcing, with predictions evaluated along the ground-truth traversal, allowing each prediction to be learned without propagating errors from earlier decoding steps~\cite{jin2018junction}. Atom and motif states are computed from the complete input linker, while traversal masks select the states included in the partial-context summary and the eligible attachment candidates. Since incoming motifs are encoded separately for attachment prediction, reconstruction uses both the global linker representation $\mathbf z_L$ and local encoder states, with the total linker charge supplied as an additional decoder conditioning variable.

The attachment head scores candidate atom pairs, treating role-equivalent attachments as multiple valid targets. Two auxiliary regression heads enforce charge consistency and predict LogP, with the latter encouraging linker representations to capture physicochemical features relevant to guest-framework interactions. In each minibatch, linkers with similar standardized RDKit values (LogP, QED, and synthetic accessibility)~\cite{Landrum2016RDKit2025093} form positive pairs, while dissimilar linkers form negative pairs. The contrastive loss is applied to a projection of the charge-conditioned linker representation. The combined objective is
\begin{equation}
\begin{aligned}
\mathcal L_L ={}&
\gamma_{\mathrm{type}}\mathcal L_{\mathrm{type}}
+\gamma_{\mathrm{pair}}\mathcal L_{\mathrm{pair}}
+\gamma_{\mathrm{charge}}\mathcal L_{\mathrm{charge}}\\
&+\gamma_{\mathrm{prop}}\mathcal L_{\mathrm{logP}}
+\gamma_{\mathrm{con}}\mathcal L_{\mathrm{con}}.
\end{aligned}
\label{eq:mofenc_linker_loss}
\end{equation}
The motif-type (\(\mathcal L_{\mathrm{type}}\)) and attachment (\(\mathcal L_{\mathrm{pair}}\)) terms use categorical and multi-positive cross-entropy losses, respectively, while the charge-consistency (\(\mathcal L_{\mathrm{charge}}\)) and LogP (\(\mathcal L_{\mathrm{logP}}\)) terms use mean-squared error. The coefficients \(\gamma\) determine the weights of the individual loss terms. Together, these objectives encourage chemically informative representations beyond molecular graph reconstruction, with the supervised contrastive objective (\(\mathcal L_{\mathrm{con}}\)) shaping the latent space based on similarities in standardized RDKit values~\cite{Landrum2016RDKit2025093,kirchoff2024salsa,10.1021/acscentsci.7b00572} to capture physicochemical properties and structural complexity~\cite{Wildman1999,Bickerton2012,Ertl2009}. Neither observed nor predicted RDKit values are appended to the decoder input.

The inclusion of RDKit-derived chemical information, both as input features and through physicochemical regularization, is supported by improved density-based clustering of the learned representations (Supplementary Information Table~S7). Hydrogen atoms are represented implicitly via hydrogen-count features rather than as explicit graph nodes, thereby reducing graph complexity while retaining information about the local chemical environment. The Supplementary Information provides detailed definitions of the message-passing, attention, and decoder operations, along with feature definitions, attachment-role criteria, property standardization, and training hyperparameters (Table~S6).

For nodes \(N\), we employ a separate graph autoencoder operating on a graph \(\mathcal G_N\), whose vertices encode element identities and whose edges are represented by an undirected binary adjacency matrix. Because nodes do not exhibit the same motif hierarchy as organic linkers, the node autoencoder operates at a single graph resolution. A dense-adjacency MPN produces environment-dependent atom states \(\mathbf h_i^N\), after which a learnable graph token \(\mathbf h_{\mathrm{token}}^{(0)}\), connected bidirectionally to all valid node atoms, is updated and mapped to the node latent:
\begin{equation}
    \mathbf z_N
    =
    \operatorname{MLP}_{N}
    \left(
        \left[
        \operatorname{MPN}_{N}
        \left(
            \mathcal G_N,
            \{\mathbf h_i^N\}
            \cup
            \{\mathbf h_{\mathrm{token}}^{(0)}\}
        \right)
        \right]_{\mathrm{token}}
    \right).
    \label{eq:mofenc_node_latent}
\end{equation}
The decoder reconstructs element identities, atom-slot presence, node adjacency, and global coordination classes from \(\mathbf z_N\). The corresponding objective is
\begin{equation}
    \mathcal L_N
    =
    \lambda_{\mathrm{elem}}\mathcal L_{\mathrm{elem}}
    +
    \lambda_{\mathrm{edge}}\mathcal L_{\mathrm{edge}}
    +
    \lambda_{\mathrm{coord}}\mathcal L_{\mathrm{coord}}
    +
    \lambda_{\mathrm{pres}}\mathcal L_{\mathrm{pres}}.
    \label{eq:mofenc_node_loss}
\end{equation}
Element identity (\(\mathcal L_{\mathrm{elem}}\)) is trained with masked categorical cross-entropy, adjacency (\(\mathcal L_{\mathrm{edge}}\)) with weighted binary cross-entropy over the upper triangle, and coordination (\(\mathcal L_{\mathrm{coord}}\)) and slot presence (\(\mathcal L_{\mathrm{pres}}\)) with sigmoid binary cross-entropy. The coefficients \(\lambda\) determine the relative weights of the individual loss terms, with their values and other architectural and training hyperparameters provided in Supplementary Information Table~S5.
\subsection{hierMOF}
\label{HierGNN}

Atomic representations are initialized with learnable element embeddings, while motif and linker representations are obtained from the pretrained hierarchical linker encoder, and node representations are obtained from the separate node encoder. First, the hierMOF-enc pretrained representations are projected into a common hidden space using class-specific linear projections, then combined with learned component-class embeddings via gated fusion.

The resulting features are then refined through message-passing blocks operating at three structural resolutions: atoms, motifs and nodes, and linkers and nodes. These blocks employ either directed three-body CGCNN or ALIGNN convolutions, depending on the hierMOF variant. Each message-passing block is followed by cross-level information exchange based on the membership relations defined by hierMOF-seg. At each hierarchical level, a child token \(c\) is associated with a parent token \(p\), and \(\mathcal C(p)\) denotes the set of children belonging to \(p\). A learned compatibility score \(s(\mathbf h_c,\mathbf h_p)\), computed from the concatenated child and parent representations, defines an attention-weighted summary of the child states:

\begin{equation}
\begin{aligned}
    \alpha_{cp}
    &=
    \frac{\exp s(\mathbf h_c,\mathbf h_p)}
    {\sum_{c'\in\mathcal C(p)}
    \exp s(\mathbf h_{c'},\mathbf h_p)},\\
    \mathbf r_p
    &=
    \sum_{c\in\mathcal C(p)}
    \alpha_{cp}\mathbf h_c .
\end{aligned}
\label{eq:mofmod_cross_level_pool}
\end{equation}
Upward and downward messages are generated by separate two-layer MLPs with Softplus activations and modulated by feature-wise sigmoid gates, $\mathbf g_p$ and $\mathbf g_c$, computed from the current token states and transformed incoming messages:
\begin{equation}
\begin{aligned}
    \mathbf h_p^{+}
    &=
        \mathbf h_p
        +
        \mathbf g_p\odot
        \operatorname{MLP}_{\uparrow}(\mathbf r_p),\\
    \mathbf h_c^{+}
    &=
        \mathbf h_c
        +
        \mathbf g_c\odot
        \operatorname{MLP}_{\downarrow}
        (\mathbf h_p^{+}),
\end{aligned}
\label{eq:mofmod_bidirectional_update}
\end{equation}
where $\odot$ denotes element-wise multiplication. Atom-level convolutions are followed by exchange between atoms and their assigned motifs or nodes, and subsequently between motifs and their parent linkers. Following linker-node message passing, a final linker-to-motif update incorporates the refined linker context into motif states.

Finally, linker, node, and motif states are aggregated separately using normalized sigmoid-gated averaging, yielding $\overline{\mathbf h}^{\,\mathrm l}$, $\overline{\mathbf h}^{\,\mathrm n}$, and $\overline{\mathbf h}^{\,\mathrm m}$, respectively. This class-specific pooling prevents the more numerous motif tokens from dominating the crystal representation. Global cell geometry is represented by the normalized lattice Gram matrix,
\begin{equation}
    \widetilde G
    =
    \frac{AA^{\mathsf T}}{V^{2/3}},
    \qquad
    V=|\det A|,
    \label{eq:mofmod_lattice_gram}
\end{equation}
where $A$ contains the lattice vectors as rows. Volume normalization removes absolute scale while retaining cell shape and relative lattice lengths. The six independent components of $\widetilde G$ are processed using a two-layer MLP to obtain $\mathbf h^{\mathrm{lat}}$. The pooled substructure and lattice representations are concatenated to form the crystal-level representation:
\begin{equation}
    \mathbf h^{\mathrm{MOF}}
    =
    [\overline{\mathbf h}^{\,\mathrm l},
     \overline{\mathbf h}^{\,\mathrm n},
     \overline{\mathbf h}^{\,\mathrm m},
     \mathbf h^{\mathrm{lat}}],
    \qquad
    \widehat y = \operatorname{Readout}(\mathbf h^{\mathrm{MOF}}).
    \label{eq:mofmod_readout}
\end{equation}

The \(\operatorname{Readout}\) operation comprises a two-layer fusion MLP followed by the prediction head, which outputs a scalar for regression or class logits for classification. Both variants used two convolutional blocks at the atom, motif-node, and linker-node resolutions with neighbor cutoffs of \(4\), \(8\), and \(15\)~\AA, respectively. Full architecture and optimization hyperparameters are provided in Supplementary Information Table~S8. All models were trained using  AdamW~\cite{loshchilov2019decoupledweightdecayregularization}. The learning rate followed a warm-up and cosine-decay schedule, and the checkpoint with the lowest validation loss was retained. In all experiments, structures that passed the segmentation checks were randomly divided into training, validation, and test sets in an 80:10:10 ratio.

\subsection{hierMOF-int}
\label{MOFINT}

The attribution is performed with respect to the hierMOF-enc representations entering hierMOF. For \(T\) tokens, these normalized representations are collected in \(H=(\mathbf h_1,\ldots, \mathbf h_t, \ldots,\mathbf h_T)\), where each \(\mathbf h_t\) corresponds to a  motif, linker, or node before projection and message passing. For a selected token \(t\), the reference \(H_t^0\) is constructed by replacing its representation with \(\boldsymbol{\mu}_{\tau(t)}\), the mean of the normalized training-set representations for its component class \(\tau(t)\) (motif, linker, or node):
\begin{equation}
    H_t^0
    =
    \left(
    \mathbf h_1,\ldots,\mathbf h_{t-1},
    \boldsymbol{\mu}_{\tau(t)},
    \mathbf h_{t+1},\ldots,\mathbf h_T
    \right).
    \label{eq:mofint_reference}
\end{equation}

Writing the normalized hierMOF prediction as \(F(H,\Omega)\), where \(\Omega\) collects the fixed atomic inputs, coordinates, periodic cell, connectivity, and hierarchical assignments, the interpolation between the reference and observed representations is defined as
\begin{equation}
    H_t(\alpha)
    =
    H_t^0
    +
    \alpha\left(H-H_t^0\right),
    \qquad
    \alpha\in[0,1].
    \label{eq:mofint_ig_path}
\end{equation}
The conditional attribution of token \(t\) is then defined in physical property units as
\begin{equation}
    A_t
    =
    \sigma_y
    \left(
    \mathbf h_t-\boldsymbol{\mu}_{\tau(t)}
    \right)^{\mathsf T}
    \int_0^1
    \nabla_{\mathbf h_t}
    F\left(H_t(\alpha),\Omega\right)
    \,\mathrm d\alpha,
    \label{eq:mofint_conditional_ig}
\end{equation}
where \(\sigma_y\) is the standard deviation used to normalize the target property.

During attribution, model parameters, atomic inputs, geometry, and hierarchical assignments are held fixed. Periodicity is handled through the hierMOF neighbor graph and therefore does not introduce independent token representations. Conditional IG is applied to one substructure at a time using a 256-point midpoint integration and training-set mean representations as reference. For detailed evaluation results, see Supplementary Information Tables~S14 and~S15.

\section*{Code Availability}
The code used to train the models is available on GitHub at \url{https://github.com/msan9908/hierMOF} and Zenodo \url{https://doi.org/10.5281/zenodo.24281754}.

\section*{Author contributions}
M.S. performed the experiments, analyzed the results, and wrote the manuscript. J.Z. supervised the project, reviewed the manuscript, provided resources, and acquired funding.

\section*{Conflicts of interest}
There are no conflicts to declare.

\subsection*{Acknowledgments}
The authors acknowledge Leo Duck and Han-Gyoung You for their assistance in testing the hierMOF framework. Funded by the European Union. Views and opinions expressed are, however, those of the author(s) only and do not necessarily reflect those of the European Union or the European Research Council Executive Agency. Neither the European Union nor the granting
authority can be held responsible for them.
This work was funded by the ERC (StG SupraModel) - 101077842.

\clearpage
\begin{appendices}

\end{appendices}
\bibliography{references}

\clearpage

\end{document}


\title{\begin{centering}\textbf{Supplementary Information: Chemical Substructure Representations for Hierarchical Learning in Metal-Organic Frameworks}\\[.5em]
\end{centering}}

\author[1,2]{\fnm{Micha\l} \sur{Sanocki}}

\author*[1,2]{\fnm{Julija} \sur{Zavadlav}}\email{julija.zavadlav@tum.de}

\affil[1]{\orgdiv{Multiscale Modeling of Fluid Materials, Department of Engineering Physics and
Computation}, \orgname{TUM School of Engineering and Design, Technical University of Munich}, \country{Germany}}

\affil[2]{\orgdiv{Atomistic Modeling Center, Munich Data Science Institute}, \orgname{Technical University of
Munich}, \country{Germany}}
\date{\vspace{20pt} \today}

\maketitle
\clearpage
\section{MOF segmentation and motif assignment}
\label{sec:si_mof_segmentation}

Node seeds were obtained from the elements in the QMOF node descriptors. If this produced no node atoms, the predefined forbidden-in-linker element set was used; this set contains metals and selected noble-gas and main-group elements. All remaining atoms were initially assigned to the linker set. Periodic connectivity was determined from minimum-image distances and covalent
radii from \texttt{pymatgen}. An edge between atoms \(i\) and \(j\) was proposed
when
\[
    d_{ij}\leq\alpha\left(r_i^{\mathrm{cov}}+r_j^{\mathrm{cov}}\right),
\]
with \(\alpha=1.30\), reduced to \(1.25\) for oxygen bonded to elements other than C, N, S, or P. Hydrogen used a maximum parent distance of
\(1.20~\text{\AA}\); the corresponding limits for F, Cl, Br, and I were \(1.70\), \(2.10\), \(2.20\), and \(2.60~\text{\AA}\), respectively. Each terminal atom was assigned to at most one heavy-atom parent. Before extracting linker components, H-H and O-O edges were removed, oxygen retained only its nearest preferred C, N, S, or P neighbor when available, and maximum degrees for B, C, N, O, F, P, S, Cl, Br, I, and H were set to \(4,4,3,2,1,5,6,1,1,1,\) and \(1\), respectively.

Connected components of the pruned linker-induced graph defined provisional linkers. A single-atom component was assigned to a bonded node when possible, otherwise merged with a bonded linker; if neither existed, the nearest periodic neighbor determined the assignment. Each retained component was unwrapped by traversing its molecular graph with minimum-image displacement vectors. RDKit connectivity and bond orders were then tested in the charge order
\(0,-1,\ldots,-4,+1,\ldots,+4\), and the first molecule passing sanitization, valence, atom-map, and connectedness checks was accepted. During molecular inference, maximum parent distances for H, F, Cl, Br, and I were \(1.35\), \(1.80\), \(2.20\), \(2.40\), and \(2.60~\text{\AA}\), respectively; these repair the inferred molecular graph, whereas the preceding limits define the periodic segmentation graph. If excessive valence remained, up to three long non-terminal edges could be removed, provided the component remained connected. An oxygen atom bearing zero, one, or two hydrogens and identified as water or hydroxide was reassigned to the node set. Failure of molecular inference or any subsequent partition check caused the complete MOF to be excluded.
Node fragments were the connected components of the final node-induced graph. Metal-containing nodes were represented by an element-labeled, single-bond topology over exactly the assigned node atoms. For metal-free nodes, bond orders were inferred using charges \(0,+1,\ldots,+6,-1,\ldots,-6\), accepting the first sanitized result. 

\begin{figure*}[ht]
    \centering
    \includegraphics[width=\textwidth]{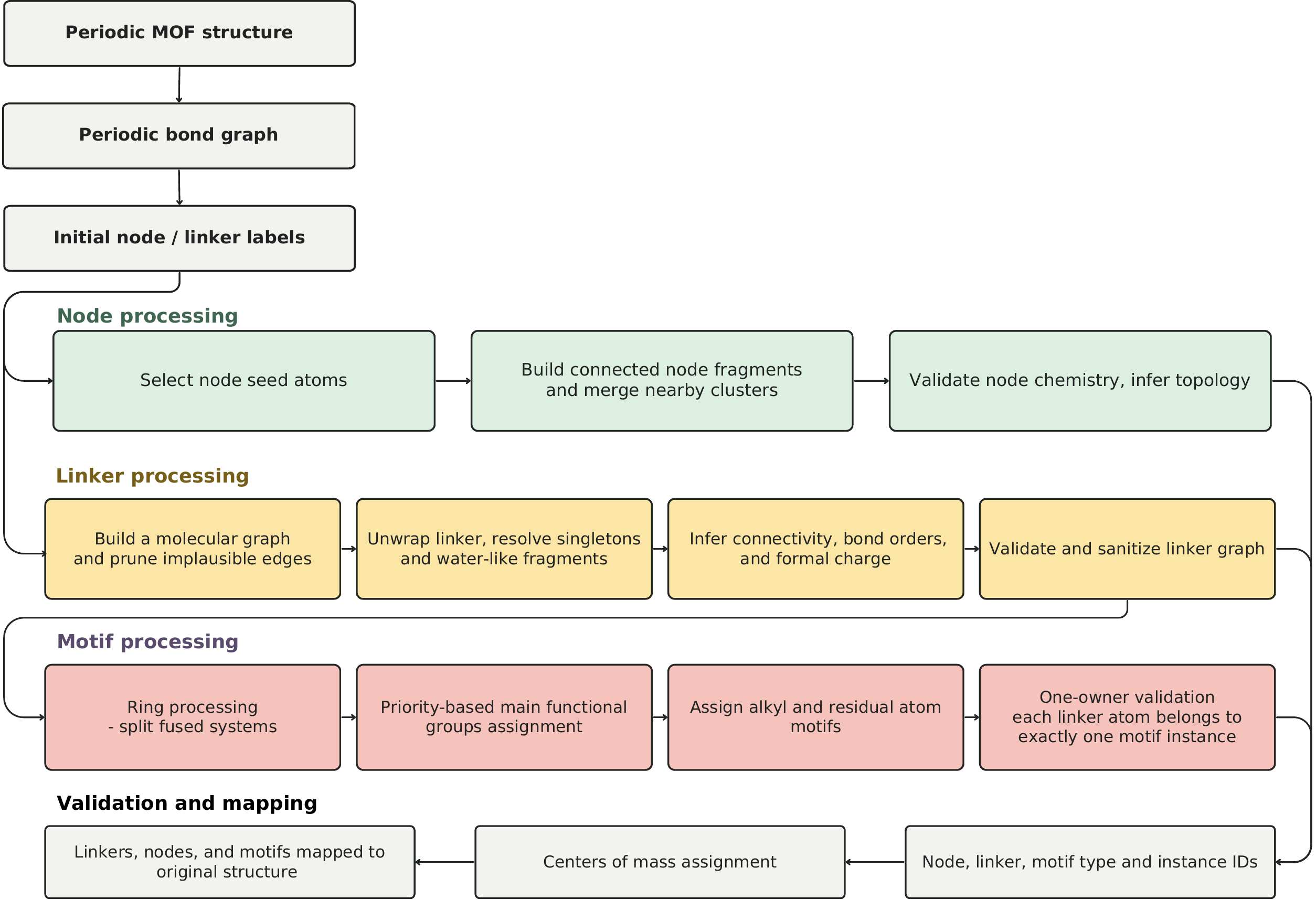}
    \caption{\textbf{hierMOF-seg workflow for assigning nodes, linkers, and linker motifs in periodic MOFs}. A periodic bond graph provides the basis for initial node/linker labels. The node branch assembles connected fragments and validates node chemistry and topology. The linker branch constructs and unwraps molecular graphs, resolves small fragments, infers connectivity, bond orders, and formal charges, then validates the graph. Motifs are assigned within validated linkers by processing rings, assigning priority-based functional groups, and labeling residual atoms. One-owner validation ensures that every linker atom belongs to exactly one motif instance. The final stage assigns centers of mass and type/instance IDs, then maps the results back to atoms in the original structure. Type IDs identify chemical categories; instance IDs distinguish their occurrences in a MOF.}
    \label{fig:MOFSEG}
\end{figure*}

Linker rings were assigned before non-ring motifs. Isolated rings were labeled by aromaticity, size, composition, formal charge, and cyclic topology. Overlapping rings were divided into subring-derived segments and shared junctions so that the partition remained disjoint. Ring atoms are separated according to their exact ring-membership signatures. Oversized junctions and ring paths are split into connected motifs containing at most six atoms; isolated rings remain intact. 
Candidate groups are assigned using deterministic chemical priorities. Nitro, nitrile, imine, carboxylate, and related distinctive groups take precedence over generic amine, alcohol, or unsaturated-carbon assignments, followed by alkyl and element-based fallback labels. During structure-to-motif mapping, explicit hydrogen atoms absent from the processed linker graph were associated with the motif of their bonded heavy-atom parent. Explicit hydrogen vertices retained in the processed linker graph could instead have their own group rows. Hydrogen atoms were not assigned exported structure-level motif IDs.

\begin{figure*}[ht]
    \centering
    \includegraphics[width=\textwidth]{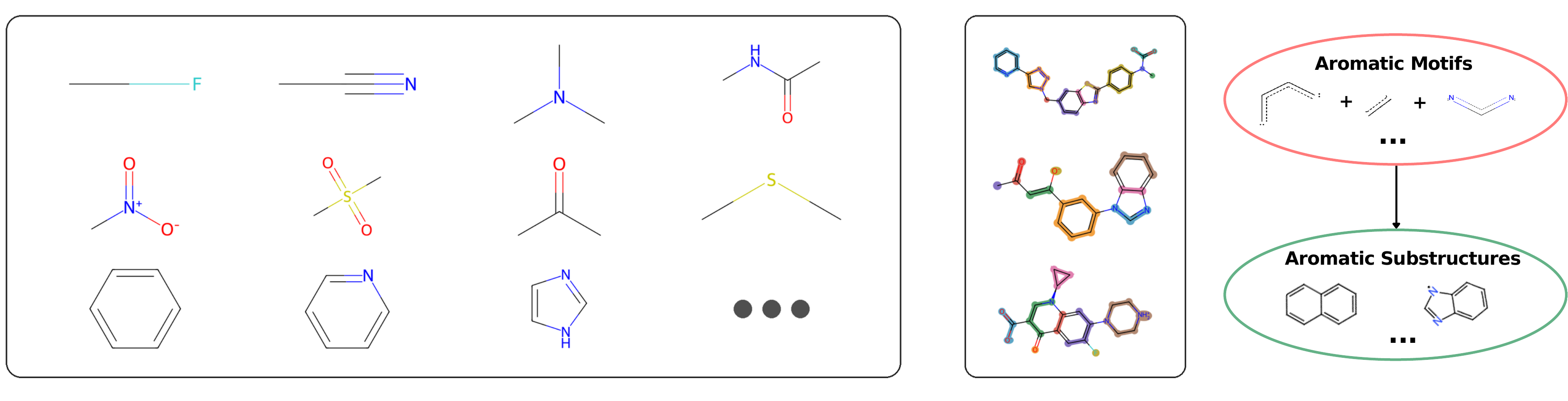}
    \caption{\textbf{Representative chemical motifs and aromatic substructures identified by hierMOF-seg.}
    Left: Examples of chemically defined motif types, including functional groups, aromatic rings, and residual fragments.
    Center: Representative organic linkers partitioned into non-overlapping, color-coded motifs.
    Right: Illustration of how smaller aromatic motifs combine to represent more complex aromatic substructures, preserving finer structural information rather than treating multi-ring systems as single motifs.}
    \label{fig:mofseg_schema}
\end{figure*}

The stored atomistic adjacency and atom-to-motif assignments preserve the connections between motifs. Node, linker, and motif type identifiers were assigned deterministically from sorted dataset-wide keys, while fragment and motif instance identifiers were local to each MOF, and motif identifiers were assigned only to linker atoms. The QMOF, hMOF, and WS24 datasets contain 411, 332, and 187 unique motif types, respectively, as listed in their \texttt{.json} files. Different charge states of otherwise identical motifs are treated as distinct motif types. This was done to account for partially charged motifs, which can be especially important for MOFs. Type IDs identify motif categories, whereas instance IDs distinguish individual occurrences within a MOF. Centers of mass were calculated using ASE atomic masses after graph-based periodic unwrapping and were wrapped back into the reference cell.

\begin{figure*}[ht]
    \centering
    \includegraphics[width=\textwidth]{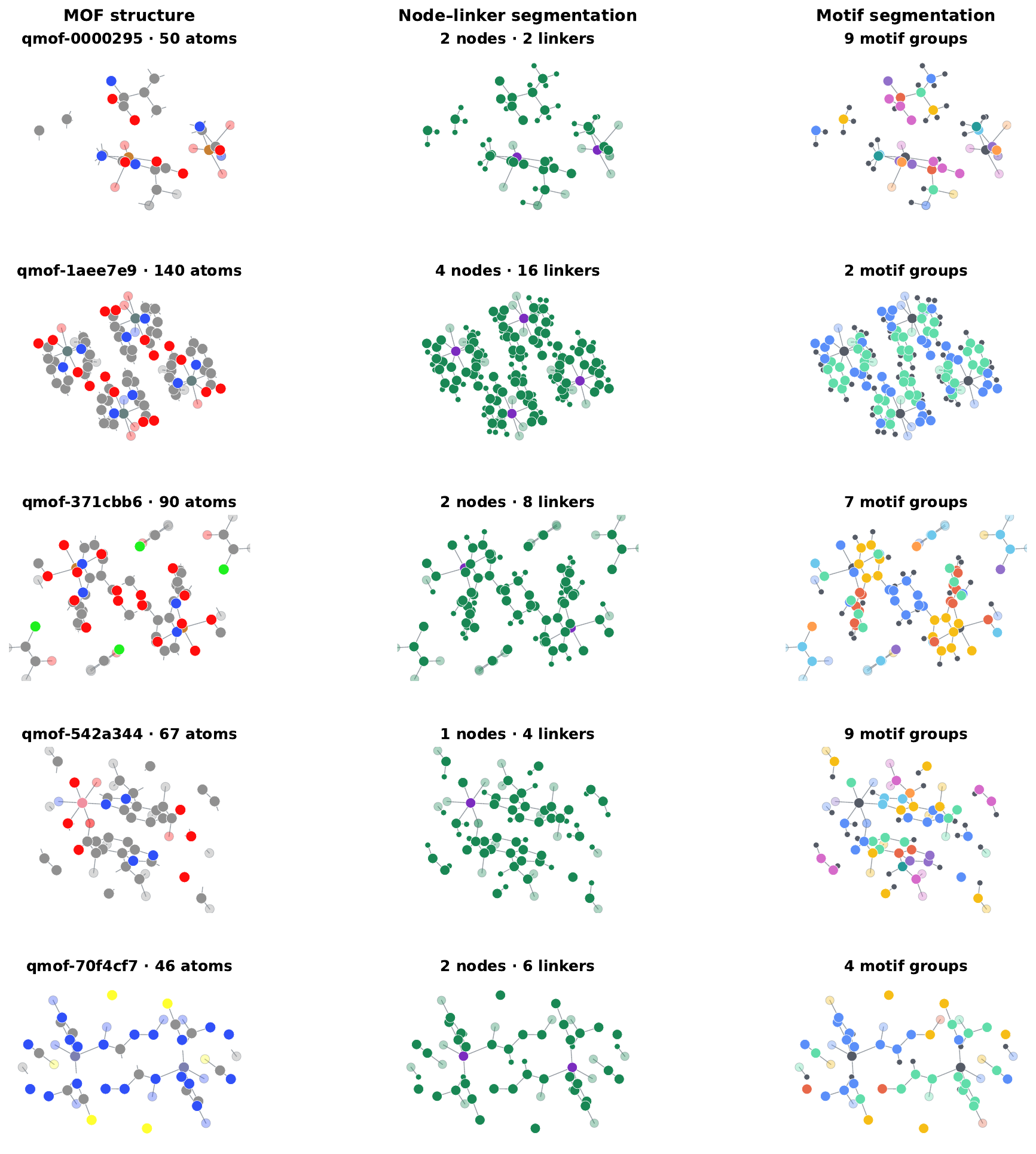}
    \caption{MOF structures and hierarchical segmentation for five selected QMOF structures. From left to right: element-colored structure, node-linker segmentation (node atoms in purple and linker atoms in green), and motif segmentation (colors distinguish motif groups). Translucent atoms indicate periodic images used to show bonds crossing unit-cell boundaries.}
    \label{fig:qmof-segmentation-01-05}
\end{figure*}

\begin{figure*}[ht]
    \centering
    \includegraphics[width=\textwidth]{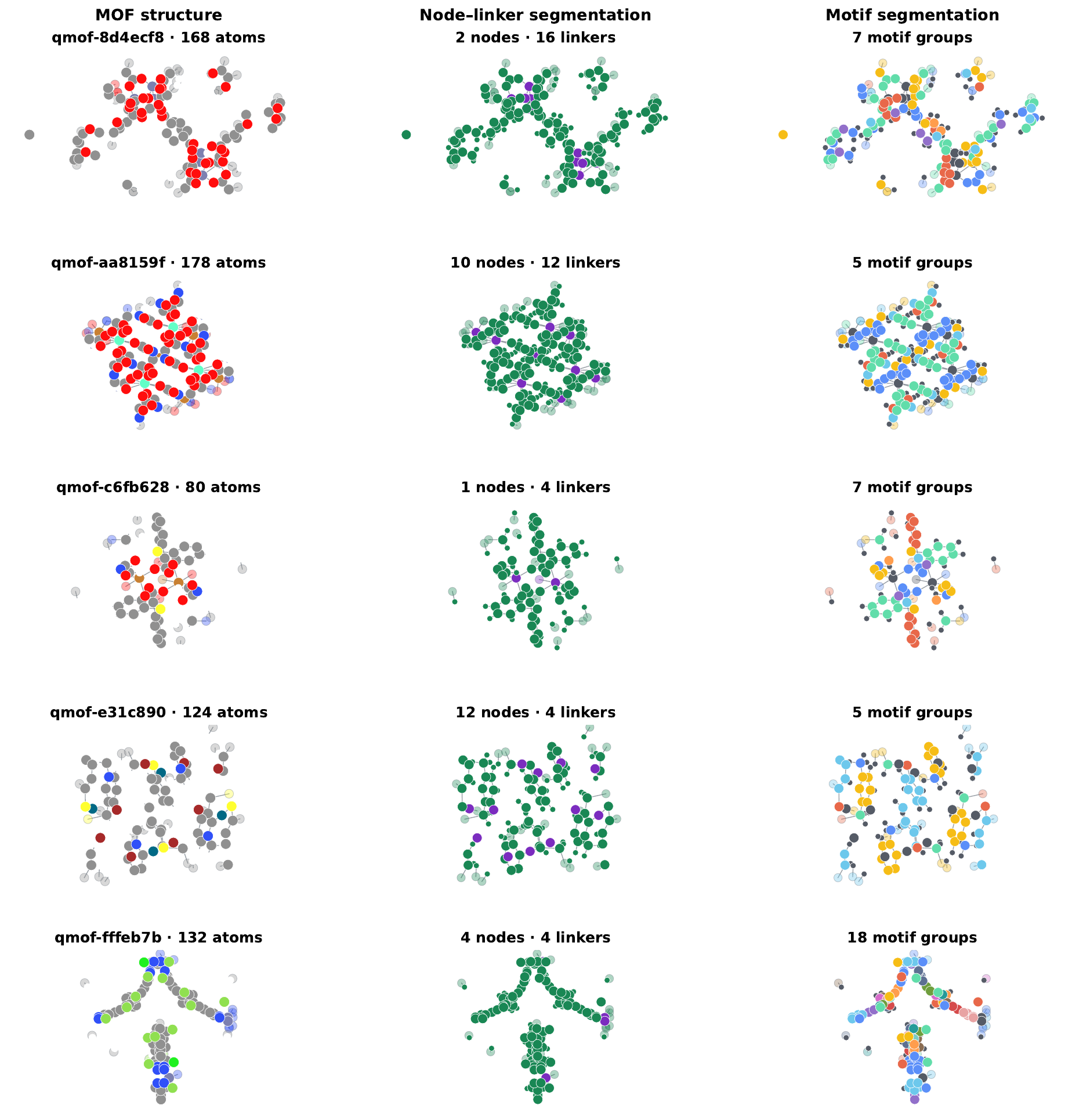}
    \caption{MOF structures and hierarchical segmentation for five additional selected QMOF structures. From left to right: element-colored structure, node-linker segmentation (node atoms in purple and linker atoms in green), and motif segmentation (colors distinguish motif groups). Translucent atoms indicate periodic images used to show bonds crossing unit-cell boundaries.}
    \label{fig:qmof-segmentation-06-10}
\end{figure*}
\clearpage
\subsection{hierMOF-seg Deduplication}
\label{sec:si_mofseg_deduplication}

\begin{figure}[p]
    \centering
    \includegraphics[width=0.85\linewidth]{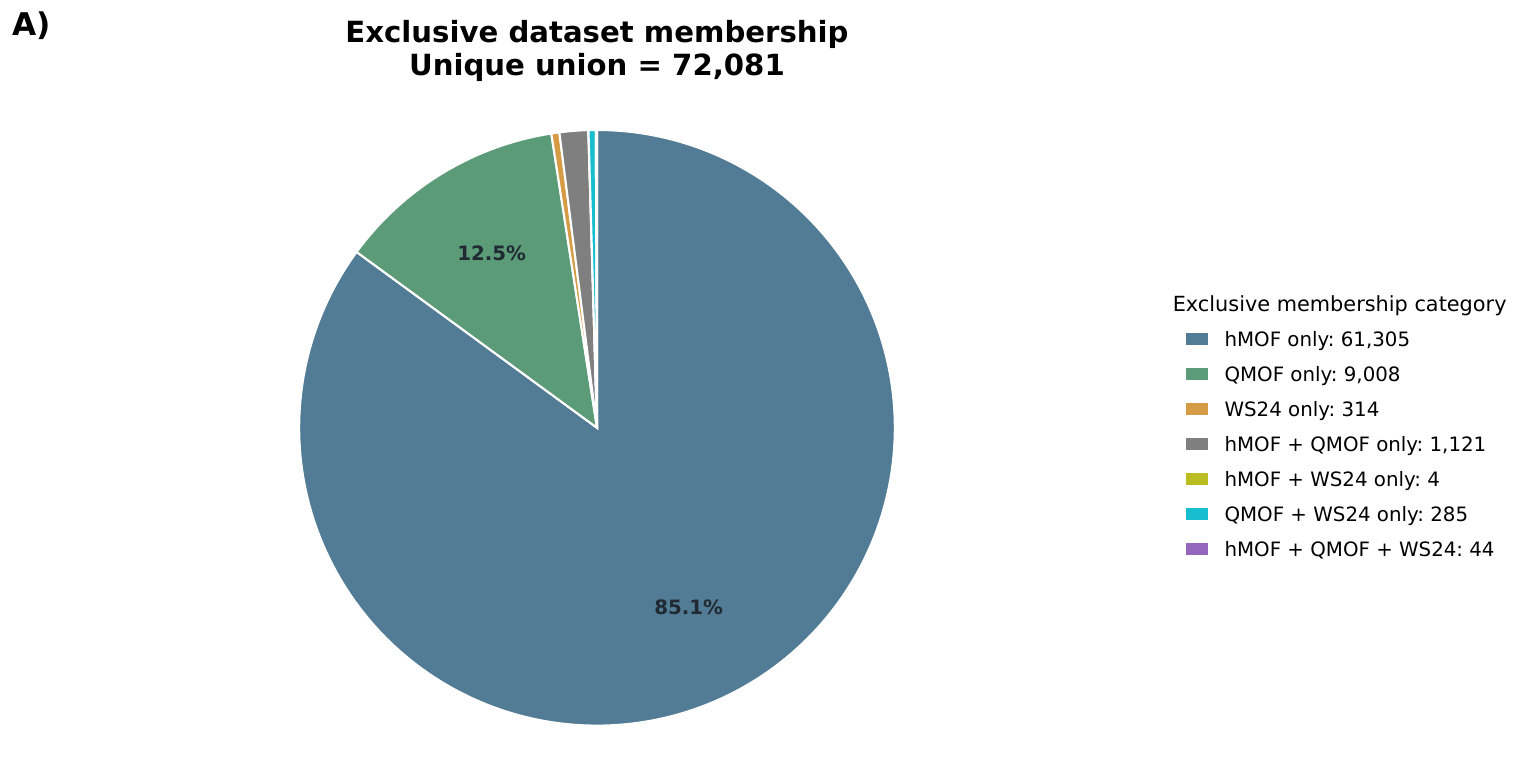}

    \vspace{0.5em}

    \includegraphics[width=0.85\linewidth]{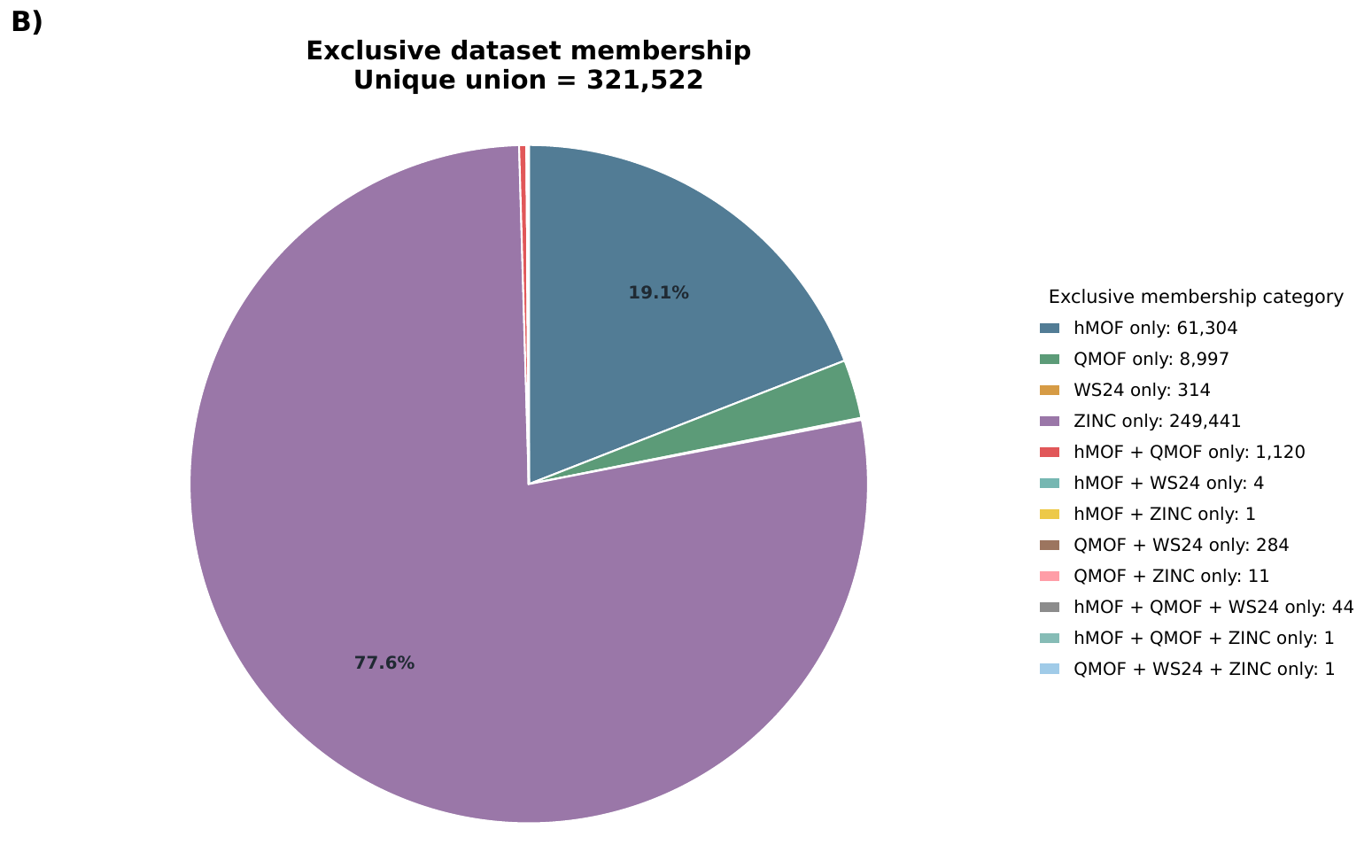}

    \caption{Exclusive membership of canonical linker structures across the datasets.
    (A) hMOF, QMOF, and WS24, with a unique union of 72,081 linkers.
    (B) The same datasets with ZINC250k included, giving a unique union of 321,522
    canonical structures. Counts are shown in the legends.}
    \label{fig:linker-overlap}
\end{figure}

\begin{table}[htbp]
    \centering
    \caption{Linker and node counts before and after deduplication.
    Registry entries are distinct records in the source maps, rather
    than physical occurrences in MOF structures. The sum adds the
    dataset-specific counts; the union counts each unique structure
    once across all three datasets.}
    \label{tab:mofseg_deduplication}
    \begin{tabular}{lrrrr}
        \hline
        Dataset &
        \shortstack{Linker registry\\entries} &
        \shortstack{Unique\\linkers} &
        \shortstack{Node registry\\entries} &
        \shortstack{Unique\\nodes} \\
        \hline
        WS24 & 3,507   & 647    & 374   & 316   \\
        hMOF & 176,604 & 62,474 & 67    & 54    \\
        QMOF & 28,264  & 10,458 & 1,708 & 1,407 \\
        \hline
        Sum   & 208,375 & 73,579 & 2,149 & 1,777 \\
        Union & -    & 72,081 & -   & 1,602 \\
        \hline
    \end{tabular}
\end{table}

\begin{table}[ht]
\centering
\caption{Overlap of unique linker structures across hMOF, QMOF, and WS24. Linkers were deduplicated by canonical isomeric SMILES after removal of atom-map numbers. Each row gives an exclusive membership category; the total is the union of the three datasets.}
\label{tab:linker_overlap}
\begin{tabular}{lr}
\hline
Dataset membership & Number of unique linkers \\
\hline
hMOF only & 61,305 \\
QMOF only & 9,008 \\
WS24 only & 314 \\
hMOF and QMOF only & 1,121 \\
hMOF and WS24 only & 4 \\
QMOF and WS24 only & 285 \\
Shared by hMOF, QMOF, and WS24 & 44 \\
\hline
Total unique linkers (union) & 72,081 \\
\hline
\end{tabular}
\end{table}

\begin{table}[htbp]
    \centering
    \small
    \caption{Summary of hierMOF-seg results for WS24~\cite{Terrones2024-ef}, hMOF~\cite{Wilmer2012-dy}, QMOF~\cite{QMOF}, and ZINC250k~\cite{10.1021/acscentsci.7b00572,10.1021/ci3001277}. ZINC250k contains organic molecules rather than MOF structures, so its unique linker/molecule count denotes unique segmented molecules. Motif counts denote distinct motif labels; source-local numeric IDs are ignored when calculating unions. The MOF union combines the three MOF datasets, while the MOF + ZINC union is the deduplicated canonical union across all four datasets. Dataset size and success rate are left blank for the combined union because it mixes MOF structures and molecules. Overlap counts and deduplication details are provided in Supplementary Information Tables~\ref{tab:mofseg_deduplication} and~\ref{tab:linker_overlap}.}
    \label{tab:mofseg_dataset_summary}
    \begin{tabular}{lrrrrr}
        \hline
        Dataset & Dataset size & Success rate & Unique linkers/molecules & Unique nodes & Unique motifs \\
        \hline
        WS24       & 911     & 89.2\%  & 647     & 316   & 187 \\
        hMOF       & 137,953 & 93.0\%  & 62,474  & 54    & 332 \\
        QMOF       & 20,375  & 95.9\%  & 10,458  & 1,407 & 411 \\
        ZINC250k   & 249,455 & 100.0\% & 249,455 & -    & 372 \\
        \hline
        MOF union  & 159,239 & 93.4\%  & 72,081  & 1,602 & 631 \\
        MOF + ZINC union & - & - & 321,522 & 1,602 & 777 \\
        \hline
    \end{tabular}
\end{table}

Linkers were deduplicated using canonical isomeric SMILES after removal of atom-map numbers, retaining distinctions in charge, bonding, and stereochemistry. Nodes were deduplicated by exact graph isomorphism of normalized atom labels and unweighted adjacency. Atom-map suffixes were removed from node labels, while encoded charge labels were retained. Equivalent graphs with different atom orderings were therefore merged.

\begin{figure*}[ht]
    \centering
    \includegraphics[width=\textwidth]{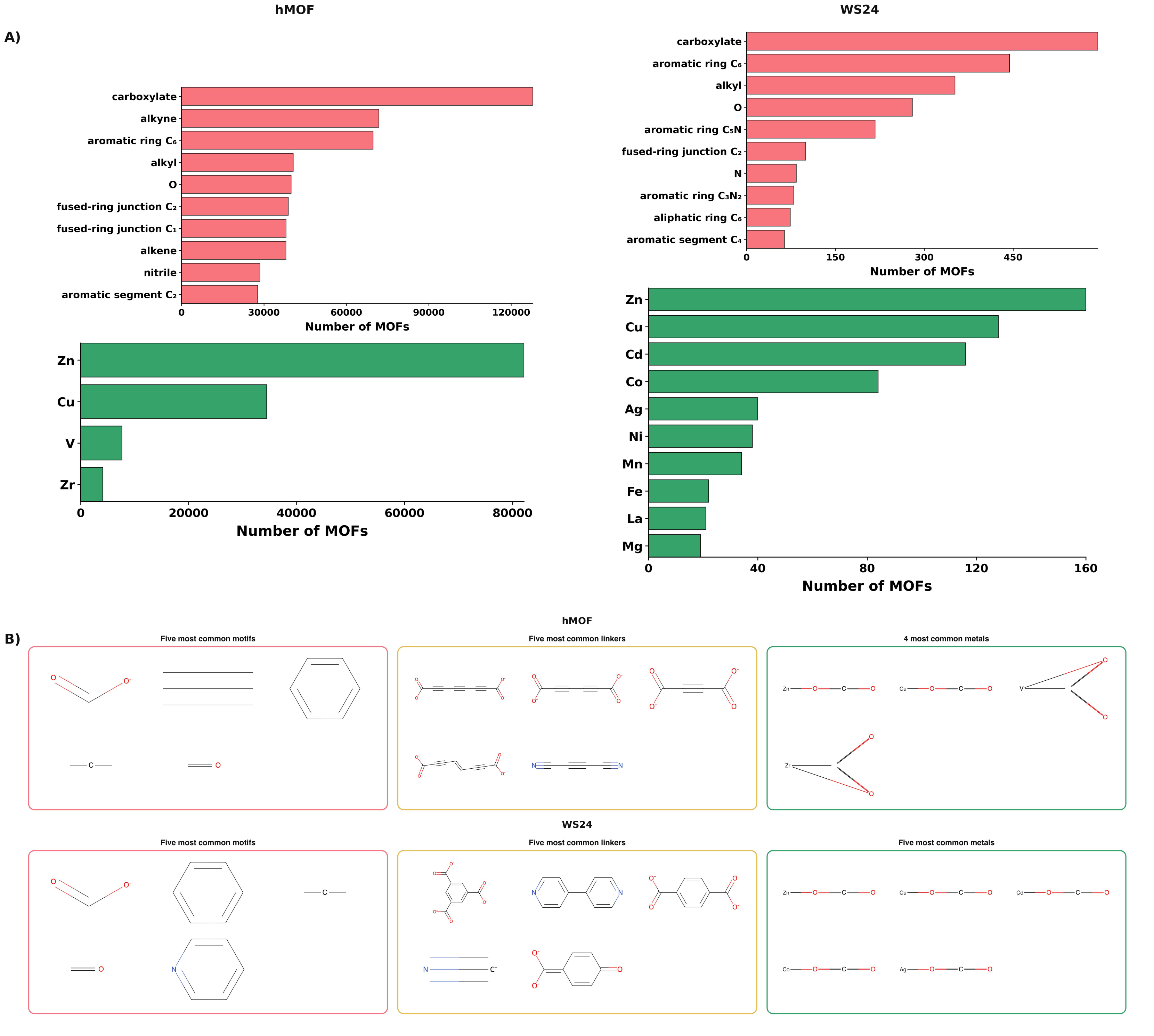}
    \caption{\textbf{Prevalence of common motifs and metals, with representative structures, in the hMOF and WS24 datasets.}
    \textbf{A)} Number of MOFs containing each of the most common assigned motifs and metal elements; each motif or metal is counted at most once per MOF. \textbf{B)} Representative drawings of the five most common motif types and canonical linker structures, and local metal–linker coordination environments. The hMOF linker examples correspond to building-block families in the original hypothetical hMOF library, including building blocks 10 and 25-27 in Supplementary Figure S2 of \citeauthor{Wilmer2012-dy}\cite{Wilmer2012-dy}.
    }
    \label{fig:ws24+hmof}
\end{figure*}

The counts in Tables~\ref{tab:mofseg_deduplication} and \ref{tab:linker_overlap} precede
train/validation/test splitting and, for linkers, motif-frequency
filtering. During framework prediction, repeated linker and node
occurrences remain separate instances and retrieve their initial
representations from the corresponding type libraries. Differences in common linker structures are illustrated in Fig.~\ref{fig:ws24+hmof}.

\begin{figure*}[ht]
    \centering
    \includegraphics[width=0.99\linewidth]{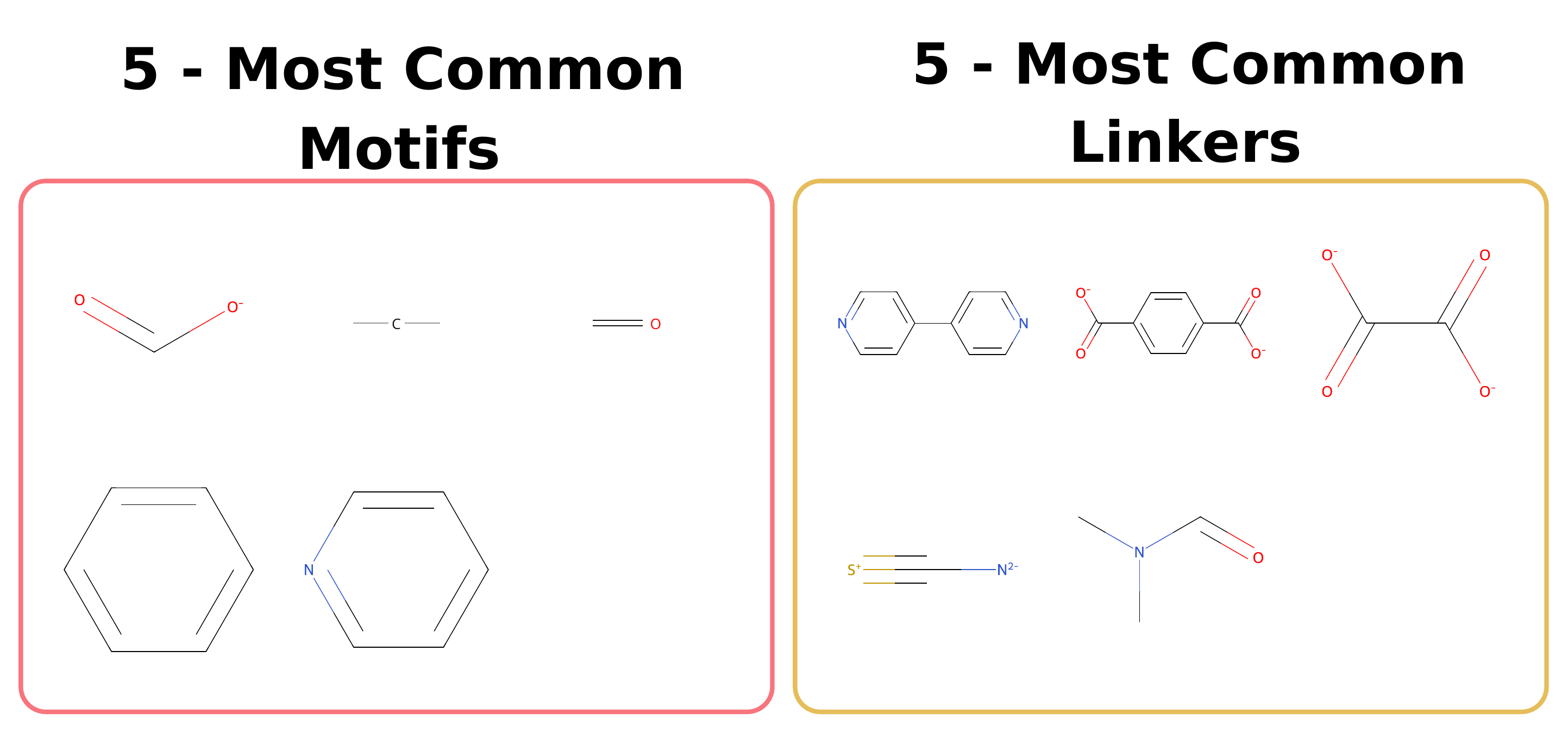}
    \caption{\textbf{hierMOF-seg segmentation results for QMOF.}
    Representative drawings of the five most common motif types and linker structures.}
        \label{fig:MOFSEG_qmof}
\end{figure*}

\clearpage

\subsection{hierMOF-seg failure analysis}
\label{sec:si_mofseg_failures}

The final hierMOF-seg runs processed 20,375 QMOF, 137,953 hMOF, and 911 WS24 structures \cite{QMOF, Terrones2024-ef,Wilmer2012-dy}, excluding 841, 9,597, and 98 structures, respectively. To identify the main sources of failure, each excluded structure was assigned to a single primary failure category based on the logged terminal failure (SI Table~\ref{tab:si_mofseg_failures}). These categories include failures in linker bond-order and charge inference, structural partitioning, and subsequent consistency checks. ``Unspecified linker inference'' denotes cases in which all tested bond-order and charge assignments failed without a more specific terminal cause being recorded.

\begin{table}[t]
\centering
\caption{Primary failure categories for structures that did not pass the
complete hierMOF-seg segmentation and validation pipeline. Each unsuccessful
structure is assigned to a single category.}
\label{tab:si_mofseg_failures}
\small
\begin{tabular}{p{0.25\textwidth}|p{0.45\textwidth}|ccc}
\toprule
Failure category & Description & QMOF & hMOF & WS24 \\
\midrule
Over-coordination/pruning &
No further bond corrections are possible. &
3 & 4,729 & 58 \\

Unspecified linker inference &
All tested bond-order and charge assignments fail. &
219 & 3,598 & 16 \\

Charge-inconsistent bond ordering &
No inferred bond-order assignment is consistent with the assigned charge. &
145 & 1,028 & 7 \\

Unassigned linker atoms &
One or more atoms assigned as linker are not included in any
reconstructed linkers. &
227 & 3 & 6 \\

Fragment atom-set inconsistency &
The reconstructed fragment does not preserve the expected atom set or
connectivity. &
156 & 95 & 3 \\

Terminal atom beyond cutoff &
No eligible neighboring atom is found within the bonding cutoff. &
79 & 106 & 6 \\

Atom correspondence lost &
Molecular standardization prevents an unambiguous mapping back to the input
atoms. &
4 & 38 & 1 \\

Metal-free node inference &
Bond and charge inference fails for a node fragment containing no metal atom. &
6 & 0 & 1 \\

Excessive valence &
The inferred molecular graph exceeds the permitted valence of at least one
element. &
2 & 0 & 0 \\
\bottomrule
\end{tabular}
\end{table}

\clearpage

\subsection{Segmentation ambiguity and chemical consistency}
\label{app:ambiguity}

\begin{figure*}[htbp]
    \centering
    \includegraphics[width=\textwidth]{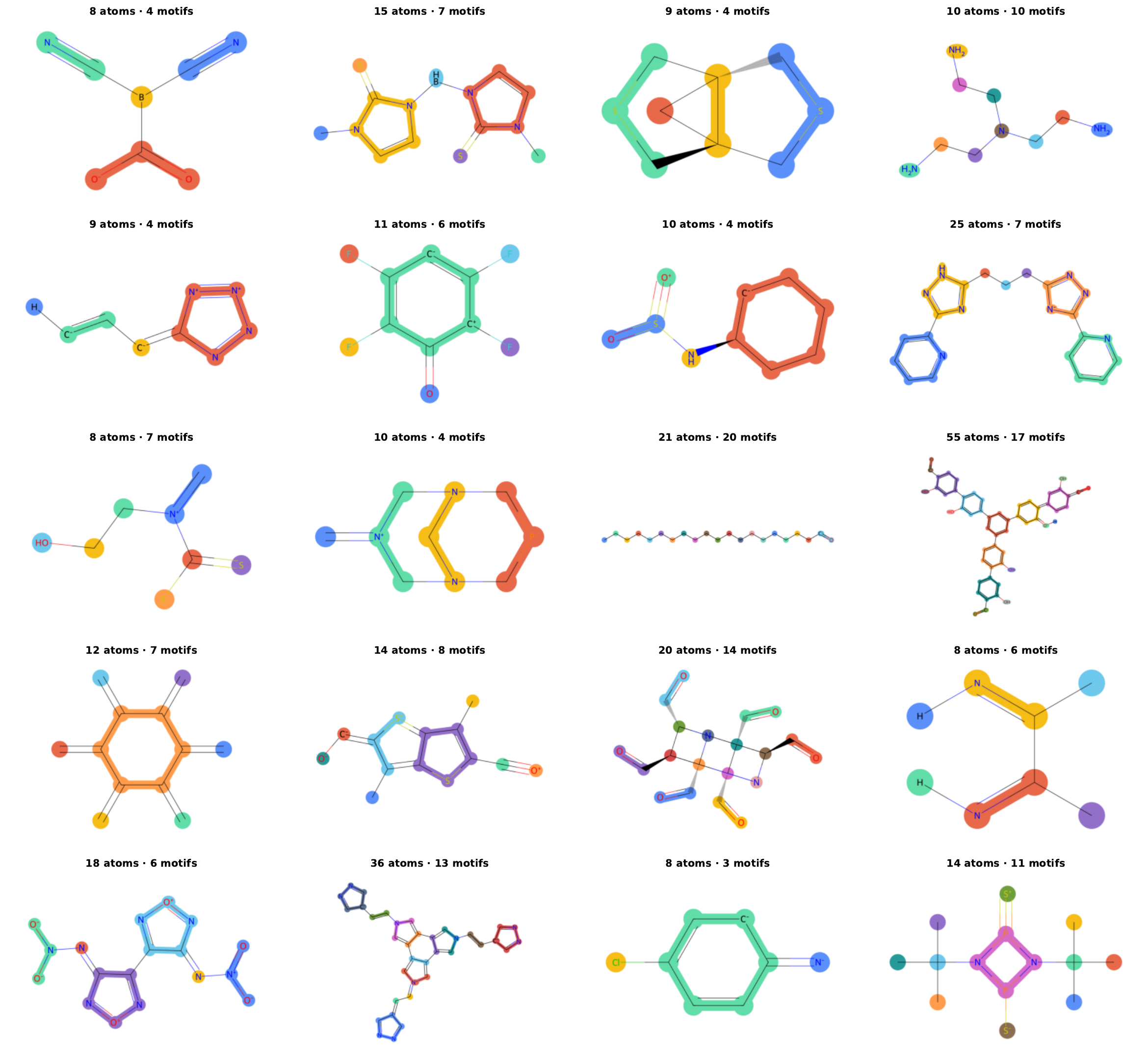}
    \caption{Twenty structurally diverse linkers selected from the QMOF dataset \cite{QMOF}. Colors identify the motif groups assigned to atoms within each linker.}
    \label{fig:qmof-linker-motifs}
\end{figure*}

The linker partitions in Figure~\ref{fig:qmof-linker-motifs} show that motif size and boundaries can vary, especially near fused or conjugated rings, linker junctions, and attachment sites. This is intended, since motifs are small recurring graph substructures, not complete functional groups. A small colored region should therefore not be interpreted as a failed attempt to identify a conventional functional group. At the same time, inferring motifs from structure alone is ambiguous. A structure-derived graph may not uniquely specify bond order, aromaticity, formal charge, protonation, tautomeric state, or the chemical role of a local substructure. This can give chemically similar or resonance-related structures different graph labels, and can make boundaries in fused or conjugated systems non-unique. Some input linkers consequently retain unusual formal-charge assignments or have motif labels that are structurally consistent but chemically non-intuitive. 

The unusual cases can arise from ambiguity in the input bond graph and from the rule-based motif boundaries used to partition it. Figure ~\ref{fig:qmof-linker-motifs} illustrates examples of linker partitions in the dataset. The input graph may contain hydrogen as its own atom, because the SMILES includes an explicit \texttt{[H]} atom. In that case, hydrogen is a separate graph vertex and can receive a separate group assignment. By contrast, \texttt{[NH]} normally denotes a nitrogen atom with one hydrogen. Differences in hydrogen representation can therefore change atom counts, group membership, and the resulting motif boundaries.

We acknowledge that the decomposition of MOF structures into nodes and organic linkers is not always chemically unambiguous, particularly around coordination environments. These uncertainties arise from MOF-specific framework segmentation rather than the subsequent partitioning of reconstructed linkers into motifs. Although such ambiguities may introduce noise into downstream representations, applying consistent segmentation rules and atom mapping across the dataset ensures comparable structural representations. Furthermore, hierMOF-enc and hierMOF may accommodate some of this residual variation by incorporating broader molecular and framework context.

\clearpage

\section{hierMOF-enc architecture details}
\label{app:mofenc_details}

This section provides additional architectural details of hierMOF-enc. Separate instances of MLPs, linear layers, and message-passing blocks have independent parameters unless stated otherwise.

\subsection{Hierarchical linker encoder}

The encoder first constructs atom and bond representations:

\begin{align}
\mathbf h_i^{(0)}
&=\operatorname{LN}\!\left(
\operatorname{MLP}_{\mathrm a}
([E_Z(Z_i),E_{\mathrm H}(n_i^{\mathrm H})])
+W_{\mathrm a}\mathbf f_i^{\mathrm a}\right),\\
\mathbf h_{ji}^{\mathrm b}
&=\operatorname{MLP}_{\mathrm b}(E_B(b_{ji}))
+W_{\mathrm b}\mathbf f_{ji}^{\mathrm b}.
\end{align}
Here, \(E_Z\), \(E_H\), and \(E_B\) denote trainable embeddings of atomic number, hydrogen count, and bond type, respectively. The vectors \(\mathbf f_i^{\mathrm a}\) and \(\mathbf f_{ji}^{\mathrm b}\) contain RDKit-derived atom and bond descriptors, while \(W_{\mathrm a}\) and \(W_{\mathrm b}\) are learned projections. The symbol $[\cdot,\cdot]$ denotes concatenation and $\operatorname{LN}$ denotes layer normalization. Atom descriptors comprise Gasteiger partial charge, degree, valence, formal charge, aromaticity, ring membership, hybridization indicators, and hydrogen count. Bond descriptors comprise conjugation, ring membership, ring-size category, stereochemistry, and a rotatable-bond indicator~\cite{Landrum2016RDKit2025093}.

Each message-passing layer combines receiver, sender, and bond representations:
\begin{equation}
\mathbf m_{j\to i}^{(\ell)}
=\phi_{\mathrm m}^{(\ell)}
\left(\left[\mathbf h_i^{(\ell)},\mathbf h_j^{(\ell)}
,\mathbf h_{ji}^{\mathrm b}\right]\right).
\label{eq:si_linker_message}
\end{equation}
Incoming messages are averaged, combined with the current atom state, and incorporated through a normalized residual update:
\begin{equation}
\mathbf h_i^{(\ell+1)}
=\operatorname{LN}\!\left(
\mathbf h_i^{(\ell)}
+\phi_{\mathrm u}^{(\ell)}\!\left(\left[
\mathbf h_i^{(\ell)}
,\frac{\sum_{j\in\mathcal N(i)}\mathbf m_{j\to i}^{(\ell)}}
{\max(1,|\mathcal N(i)|)}\right]\right)\right).
\label{eq:si_linker_update}
\end{equation}
Here, $\ell$ indexes layers, $\mathcal N(i)$ contains the valid bonded neighbors of atom $i$, and $\phi_{\mathrm m}^{(\ell)}$ and $\phi_{\mathrm u}^{(\ell)}$ are learned message and update transformations. The final atom states, denoted $\mathbf h_i^{\mathrm a}$, enter the hierarchical motif encoder described in the main text.

Both atom-to-motif and motif-to-linker pooling use learned attention-weighted averaging:
\begin{equation}
\operatorname{AttnPool}_{\mathbf w}
\left(\{\mathbf h_i\}_{i\in I}\right)
=
\sum_{i\in I}
\frac{
    \exp(\mathbf w^{\mathsf T}\mathbf h_i)
}{
    \sum_{j\in I}
    \exp(\mathbf w^{\mathsf T}\mathbf h_j)
}
\mathbf h_i.
\end{equation}

Here, \(I\) contains the valid states being pooled. The operation is applied to atom representations within each motif and to motif representations within each linker:
\begin{equation}
\begin{aligned}
    \widetilde{\mathbf h}_u^{\mathrm m}
    &=
    \operatorname{AttnPool}_{\mathbf w_{\mathrm a}}
    \left(
        \{\mathbf h_i^{\mathrm a}\}_{i\in\mathcal A_u}
    \right),\\
    \mathbf h_L
    &=
    \operatorname{AttnPool}_{\mathbf w_{\mathrm m}}
    \left(
        \{\mathbf z_u^{\mathrm m}\}_{u\in\mathcal V_L^{\mathrm m}}
    \right).
\end{aligned}
\end{equation}
The two pooling operations use independently learned scoring vectors. In parallel, an auxiliary branch represents each undirected bond as a separate state initialized from its endpoint atom representations and bond features. Two rounds of atom-to-bond and bond-to-atom message passing produce a mean-pooled bond summary. This branch contributes to the linker readout without replacing the atom states used for motif pooling or attachment prediction.  Following motif-level message passing, the attention-pooled motif summary $\mathbf h_L$ is combined with mean-pooled atom, motif, and bond states to obtain the linker latent:
\begin{equation}
    \mathbf z_L
    =
    \operatorname{MLP}_{L}
    \left(
        \operatorname{LN}
        \left(
            [
                \mathbf h_L,
                \overline{\mathbf h}_L^{\mathrm a},
                \overline{\mathbf h}_L^{\mathrm m},
                \overline{\mathbf h}_L^{\mathrm b}
            ]\right)
    \right).
\end{equation}
The overbars denote mean pooling over valid states. Layer normalization is applied before the three-layer projection, which uses tanh activations and training-time dropout after the first two layers.

\subsection{Hierarchical linker decoder}
\label{app:mofenc_dec}

The decoder predicts motif identities and attachment relationships along a breadth-first traversal, rather than reconstructing the linker atom by atom. At step $t$, let $u_t$ denote the current motif and $\mathbf z_{u_t}^{\mathrm m}$ its encoded representation. The partial-context vector $\mathbf s_t$ summarizes the previously encountered motifs using two-head attention, with $\mathbf z_{u_t}^{\mathrm m}$ as the query and the traversal mask selecting the participating motif states. These states are obtained from the complete input linker during teacher-forced training.

The next-motif prediction also uses the charge-conditioned linker representation $\mathbf d_L=\operatorname{LN}([\mathbf z_L,q_L])$, where $\mathbf z_L$ is the deterministic learned linker latent, $q_L$ is the observed total charge, and $[\cdot,\cdot]$ denotes concatenation. The observed charge is therefore treated as a decoder conditioning variable rather than information that must be inferred solely from the latent representation. The charge term in the training objective is evaluated from this charge-conditioned decoder representation and is used as an auxiliary reconstruction-consistency term, rather than as an independent assessment of charge information encoded in \(\mathbf z_L\). The probability vector over the motif vocabulary is
\begin{equation}
\mathbf p_t^{\mathrm{type}} =
\operatorname{softmax}\!\left(
\operatorname{MLP}_{\mathrm{type}}
\left(\left[
\mathbf z_{u_t}^{\mathrm m}
, \mathbf s_t
, \mathbf d_L
\right]\right)
\right),
\label{eq:si_linker_type_decode}
\end{equation}
whose entry indexed by $c$ gives the probability that the next motif has type $c$. For the root prediction, the same prediction head is used with a learned projection of $\mathbf d_L$ in place of $\mathbf z_{u_t}^{\mathrm m}$ and a zero partial-context vector.

Attachment prediction scores candidate pairs $(i,j)$, where atom $i$ belongs to the incoming motif and atom $j$ belongs to the partial linker. Let $\mathbf h_i^{\mathrm{a,new}}$ denote the atom state obtained by encoding the incoming motif separately, and let $\mathbf h_j^{\mathrm a}$ denote the existing atom state obtained from the complete-linker encoder. Each state is augmented with the current motif context and an anchor-use count before layer normalization:
\begin{align}
\mathbf v_i^{\mathrm{new}}
&=
\operatorname{LN}_{\mathrm{new}}\!\left(\left[
\mathbf h_i^{\mathrm{a,new}}
, \mathbf z_{u_t}^{\mathrm m}
, 0
\right]\right),\\
\mathbf v_j^{\mathrm{part}}
&=
\operatorname{LN}_{\mathrm{part}}\!\left(\left[
\mathbf h_j^{\mathrm a}
, \mathbf z_{u_t}^{\mathrm m}
, a_j^{(t)}
\right]\right).
\end{align}
Here, $a_j^{(t)}$ counts the previous uses of atom $j$ as an attachment
anchor along the traversal. The normalization operators have separately learned
parameters.

The pair-scoring network receives
\begin{equation}
\begin{aligned}
\mathbf r_{ij}^{(t)} ={}&\bigl[
\mathbf v_i^{\mathrm{new}}
, \mathbf v_j^{\mathrm{part}}
, \left(
\mathbf v_i^{\mathrm{new}}\odot\mathbf v_j^{\mathrm{part}}
\right)\\
&, \left|
\mathbf v_i^{\mathrm{new}}-\mathbf v_j^{\mathrm{part}}
\right|
, \operatorname{LN}_{\mathrm{pair}}(\mathbf d_L)\bigr],
\end{aligned}
\label{eq:si_attachment_pair_rep}
\end{equation}
where $\odot$ and $|\cdot|$ denote elementwise multiplication and
absolute value, respectively. Thus, each pair is described by its
two context-augmented anchor states, their pairwise interactions,
and the normalized charge-conditioned linker representation.

Multiple positive attachments are identified among eligible candidates
by matching the reference attachment's anchor-role identifiers at both
endpoints, together with the partial-linker motif type and anchor-use
count. The anchor-role identifiers are assigned during preprocessing
and specify the attachment-role equivalence used by the loss.
The candidate scores are normalized over the available attachment pairs,
\begin{equation}
    p_{ij}^{(t)}
    =
    \frac{
        \exp(
            \operatorname{MLP}_{\mathrm{pair}}
            (\mathbf r_{ij}^{(t)})
        )
    }{
        \sum_{(k,l)\in\mathcal P_t}
        \exp(
            \operatorname{MLP}_{\mathrm{pair}}
            (\mathbf r_{kl}^{(t)})
        )
    },
    \label{eq:si_attachment_probability}
\end{equation}
where \(\mathcal P_t\) is the candidate-pair set. If several pairs are chemically equivalent, their probabilities are combined in the attachment loss,
\begin{equation}
    \mathcal L_{\mathrm{pair}}^{(t)}
    =
    -\log
    \left(
        \sum_{(i,j)\in\mathcal P_t^{+}}
        p_{ij}^{(t)}
    \right),
    \label{eq:si_attachment_loss}
\end{equation}
with \(\mathcal P_t^{+}\) denoting the equivalent correct attachments. The total linker objective is given in the main text. Because teacher-forced decoding uses encoder states computed from the complete input linker, reconstruction is not constrained by an exclusive whole-linker latent bottleneck. Reported reconstruction metrics therefore measure prediction accuracy along the reference traversal rather than autonomous reconstruction of complete molecules from the global latent alone.

\subsection{Inorganic-node encoder and decoder}

Before construction of the graph-token latent defined in the main text, node atoms are initialized from their element types and processed by the dense-adjacency message-passing stack,
\begin{align}
    \mathbf x_i^N
    &=
    \operatorname{LN}
    \left(
        \operatorname{MLP}_N
        \left(E_N(c_i^N)\right)
    \right),\\
    \{\mathbf h_i^N\}
    &=
    \operatorname{MPN}_N
    \left(
        \mathcal G_N,
        \{\mathbf x_i^N\}
    \right).
    \label{eq:si_node_encoder}
\end{align}

For decoding, the node latent is projected to \(K_N\) candidate atom slots,
\begin{equation}
    \{\mathbf r_k^{N,0}\}_{k=1}^{K_N}
    =
    \operatorname{Expand}_N(\mathbf z_N),
    \label{eq:si_node_expand}
\end{equation}
which are refined by message passing on a complete graph without self-edges,
\begin{equation}
    \{\mathbf r_k^{N}\}
    =
    \operatorname{MPN}_{\mathrm{dec}}
    \left(
        \mathcal G_{K_N}^{\mathrm{complete}},
        \{\mathbf r_k^{N,0}\}
    \right).
    \label{eq:si_node_slot_mpn}
\end{equation}
Independent heads predict the element distribution and slot presence,
\begin{align}
    \mathbf p_k^{\mathrm{elem}}
    &=
    \operatorname{softmax}
    \left(
        \operatorname{Linear}_{\mathrm{elem}}
        (\mathbf r_k^{N})
    \right),\\
    p_k^{\mathrm{pres}}
    &=
    \sigma
    \left(
        \operatorname{Linear}_{\mathrm{pres}}
        (\mathbf r_k^{N})
    \right).
    \label{eq:si_node_heads}
\end{align}

For adjacency reconstruction, the predicted element distribution
is converted to a soft element embedding,
\begin{equation}
    \widetilde{\mathbf e}_i
    =
    \sum_c
    p_{i,c}^{\mathrm{elem}}E_{\mathrm{dec}}^N(c),
    \label{eq:si_soft_element_embedding}
\end{equation}
where $c$ indexes the element classes, including the padding class,
$p_{i,c}^{\mathrm{elem}}$ is the predicted probability of class $c$
at slot $i$, and $E_{\mathrm{dec}}^N$ is a learned decoder embedding
table with parameters separate from the encoder embedding $E_N$.
The soft element embedding is concatenated with the refined slot
state to form
\begin{equation}
    \mathbf u_i
    =
    [\mathbf r_i^N,\widetilde{\mathbf e}_i],
\end{equation}
where $[\cdot,\cdot]$ denotes concatenation and $i$ indexes a candidate atom
slot. Independently parameterized two-layer MLPs produce query and key vectors of equal dimension, while a linear projection produces a scalar slot bias:

\begin{equation}
\begin{aligned}
    \mathbf q_i &= \operatorname{MLP}_Q(\mathbf u_i),\\
    \mathbf k_i &= \operatorname{MLP}_K(\mathbf u_i),\\
    b_i &= \mathbf w_b^{\mathsf T}\mathbf u_i.
\end{aligned}
\end{equation}

Together with a learned global adjacency bias $b_{\mathrm{adj}}$, these quantities define the symmetric adjacency logit and predicted edge probability:
\begin{equation}
    \ell_{ij}^{\mathrm{adj}}
    =
    \frac{1}{2}
    \left(
        \mathbf q_i^{\mathsf T}\mathbf k_j
        +
        \mathbf q_j^{\mathsf T}\mathbf k_i
    \right)
    +
    b_i+b_j+b_{\mathrm{adj}},
    \qquad
    p_{ij}^{\mathrm{adj}}
    =
    \sigma(\ell_{ij}^{\mathrm{adj}}).
    \label{eq:si_node_adjacency}
\end{equation}
Diagonal entries are suppressed. A separate decoder head predicts the global coordination classes from $\mathbf z_N$. The corresponding reconstruction losses are combined as specified by the node objective in the main text.
\clearpage

\subsection{hierMOF-enc training}
\label{app:mofenc_train}

Linkers were deduplicated before splitting, using canonical isomeric SMILES after removing atom-map numbers, while retaining distinctions in charge, bonding, and stereochemistry. Duplicate structures within and across datasets were treated as a single training record. Existing hierMOF-seg partitions were preserved, with explicit atom and motif correspondences maintained between source and combined records. Motif frequencies were calculated separately for each deduplicated training catalog. The maximum capacities were 256 atoms, 256 directed atom edges, 64 motifs, 128 directed motif edges, and
64 traversal steps per molecule. Each incoming motif was limited to 64 atoms, and at most 32 candidate anchors in the partial linker.

Retained linker records were randomly divided into training, validation, and test subsets in an 80:10:10 ratio. Linker encoders were trained for 120 epochs. The checkpoint with the lowest validation reconstruction loss was selected for subsequent analysis. Property standardization used means and standard deviations calculated exclusively from the retained training subset. All three linker encoders used a deterministic 64-dimensional latent representation. The decoder was conditioned only on the observed total linker charge in addition to the learned representation. Neither observed nor predicted LogP, QED, or synthetic-accessibility values were appended to its input. Auxiliary LogP prediction and the property-based contrastive objective provided supervision during training but were not included in the checkpoint-selection criterion.

Node encoders were trained separately using the settings in Table~\ref{tab:si-node-hyperparameters}. Node records were deduplicated by exact graph isomorphism of normalized atom labels and unweighted adjacency, independently of atom ordering and source identifiers. Coordination-class targets were combined for identical input graphs. This yielded 1,407 unique graphs for QMOF and 1,602 for the combined QMOF, WS24, and hMOF catalog. Both catalogs were randomly split in an approximately 80:10:10 ratio. Node training ran for 70 epochs without early stopping, and the checkpoint minimizing the weighted total validation reconstruction loss was selected. Incomplete final batches were omitted during node training and loss evaluation.

\begin{table}[t]
\centering
\small
\caption{Node-encoder architecture and training hyperparameters. These settings are separate from those of the linker encoders.}
\label{tab:si-node-hyperparameters}
\begin{tabular}{p{0.34\textwidth}|p{0.58\textwidth}}
\toprule
Parameter & Value \\
\midrule
Hidden dimension & 128 \\
Latent dimension & 128 \\
Atom embedding dimension & 64 \\
Atom message-passing layers & 4 \\
Graph-level aggregation & Learned graph token with one additional message-passing layer \\
Decoder dropout & 0.1 \\
Batch size & 6 \\
Epochs & 70 \\
Optimizer & Adam with Nesterov momentum and decoupled weight decay \\
Optimizer parameters & $\beta_1=0.9$, $\beta_2=0.99$, $\epsilon=10^{-8}$ \\
Weight decay & $10^{-4}$ \\
Peak learning rate & $5\times10^{-4}$ \\
Learning-rate schedule & 1,000-step warmup; linear polynomial decay toward 5\% of peak at step 30,000 \\
Gradient clipping & Disabled \\
Reconstruction weights & Atom type 1; adjacency 2; coordination 10; atom presence 10 \\
Checkpoint selection & Minimum weighted total validation reconstruction loss \\
\bottomrule
\end{tabular}
\end{table}

\begin{table}[t]
\centering
\small
\caption{Architecture and training hyperparameters of the hierMOF-enc linker encoder.}
\label{tab:si-mofenc-hyperparameters}
\begin{tabular}{p{0.28\textwidth}|p{0.64\textwidth}}
\toprule
Parameter & Value \\
\midrule
Representation dimensions & Atom 256; motif 128; latent 64 \\
Embedding dimensions & Atom 128; motif 64 \\
Message-passing layers & Atom 2; motif 1 \\
Dropout & 0.1 \\
Batch size & 32 \\
Epochs & 120 \\
Optimizer & Nesterov Adam with decoupled weight decay $10^{-4}$ \\
Optimizer parameters & $\beta_1=0.9$, $\beta_2=0.99$, $\epsilon=10^{-8}$ \\
Learning-rate schedule & Peak $2\times10^{-3}$; 10\% warmup; polynomial decay to 5\% of peak \\
Gradient clipping & Global norm 1.0 \\
Reconstruction weights & $\gamma_{\mathrm{type}}=100$; $\gamma_{\mathrm{pair}}=100$; $\gamma_{\mathrm{charge}}=10$ \\
LogP prediction loss & $\gamma_{\mathrm{logP}}=10$ \\
Contrastive loss & $\gamma_{\mathrm{con}}=10$  \\
\bottomrule
\end{tabular}
\end{table}

\clearpage

\subsection{hierMOF-enc latent space analysis}
\label{app:latent}

\begin{figure}[htbp]
\centering
\includegraphics[width=\textwidth]{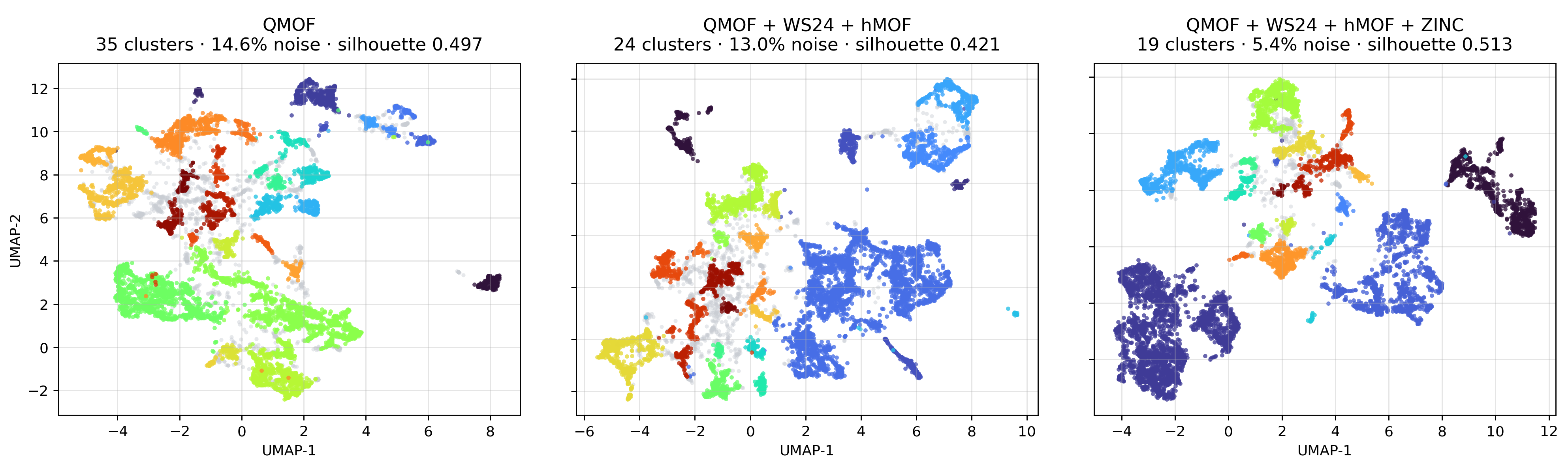}
\caption{Independent two-dimensional UMAP views of the QMOF linker records encoded by the three hierarchical encoders. Point colors denote HDBSCAN labels obtained in the separate intermediate-dimensional UMAP space described in the text. Cluster colors are model-specific.}
\label{fig:si-fixed120-qmof-umap}
\end{figure}
Three linker encoders sharing the same architecture and training hyperparameters, apart from dataset-dependent motif vocabularies, were trained on progressively broader datasets: (i) QMOF, (ii) QMOF, WS24, and hMOF, and (iii) QMOF, WS24, hMOF, and ZINC250k \cite{QMOF, Terrones2024-ef,Wilmer2012-dy, 10.1021/acscentsci.7b00572,10.1021/ci3001277}. The comparison uses the QMOF linkers represented by all three encoders (Figure~\ref{fig:si-fixed120-qmof-umap}). For each encoder, PCA of standardized learned latent vectors determines the number of components required to explain 95\% of the variance (Table~\ref{tab:si-mofenc-qmof-clustering}). UMAP then maps the full standardized latent vectors—not the retained PCA scores—to that number of dimensions for HDBSCAN clustering~\cite{10.1007/978-3-642-37456-2_14}. A separate UMAP projection of the full standardized latent vectors provides the two-dimensional visualization. For the three-encoder comparison, UMAP used \(n_{\mathrm{neighbors}}=30\), \(\mathrm{min\_dist}=0.1\), and Euclidean distance, while HDBSCAN used Euclidean distance and \(\mathrm{min\_cluster\_size}=\max\!\left(10,\operatorname{round}(0.005N)\right)\).

\begin{figure}[htbp]
\centering
\includegraphics[width=0.88\textwidth,height=0.78\textheight,keepaspectratio]{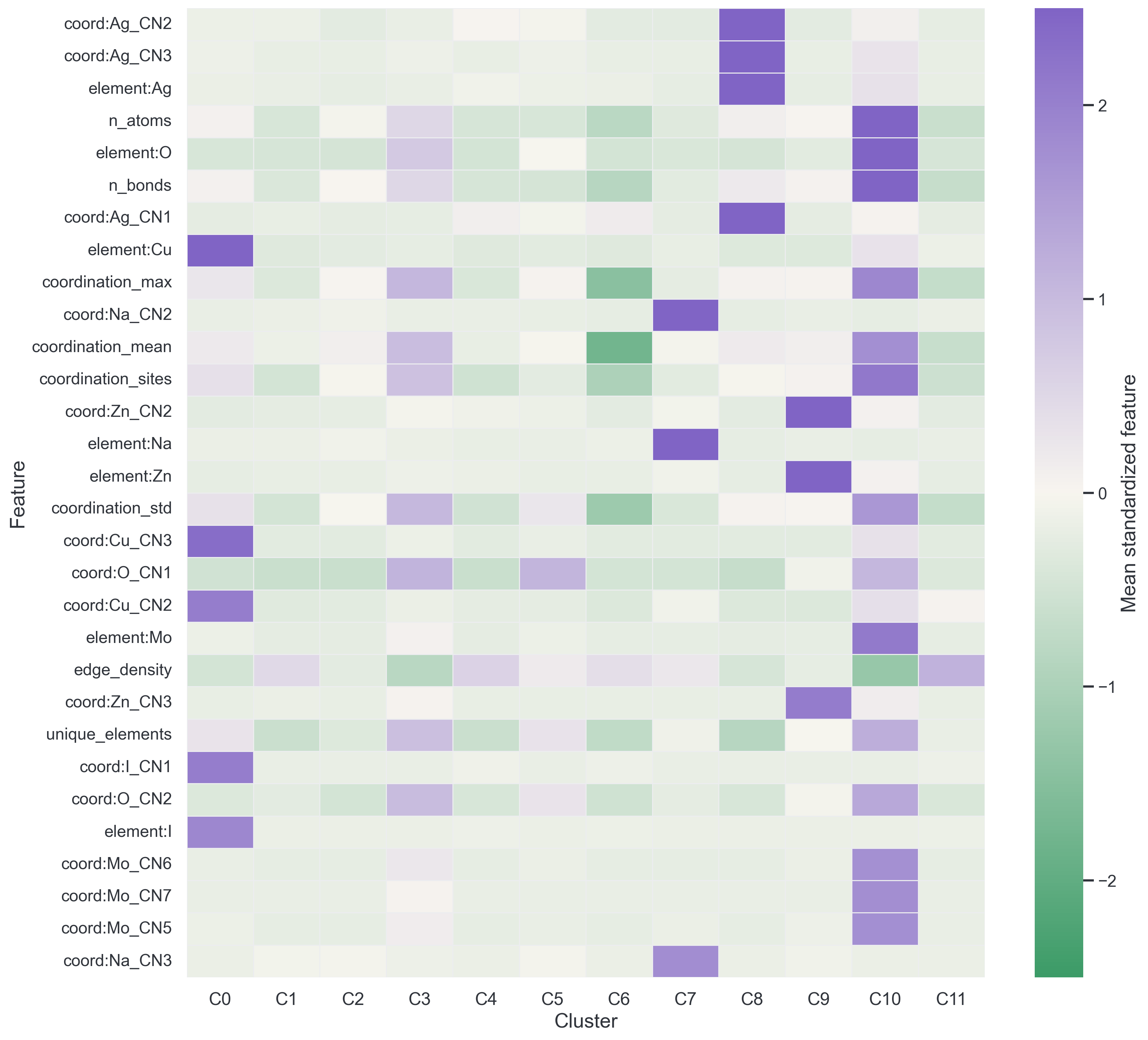}
\caption{Node-latent cluster profiles. Columns are the 12 K-means clusters
and rows are metadata features; colors show within-cluster mean standardized
feature values. Cluster identifiers are arbitrary.}
\label{fig:si-node-cluster-profiles}
\end{figure}

\begin{figure}[htbp]
\centering
\includegraphics[width=0.88\textwidth,height=0.78\textheight,keepaspectratio]{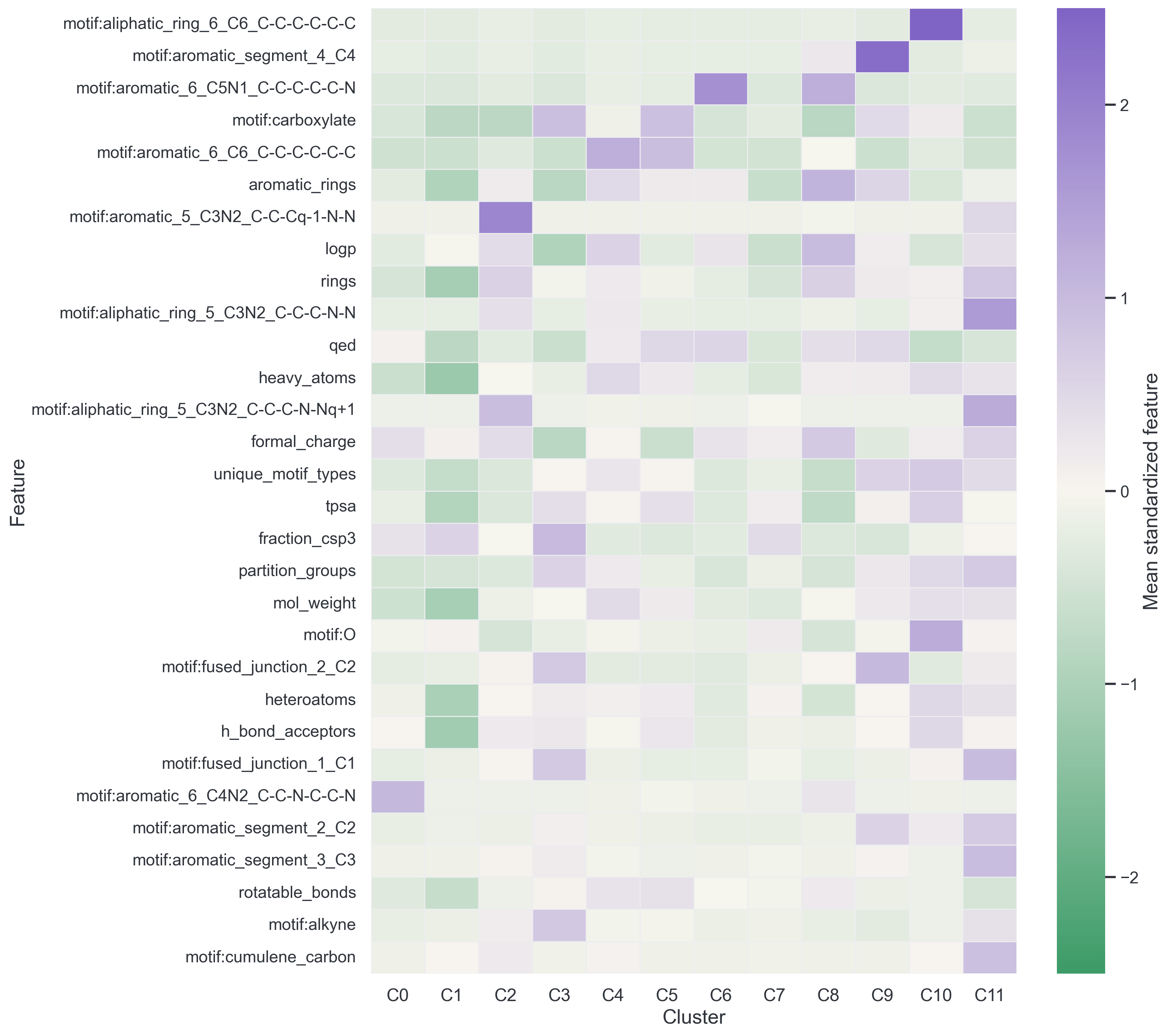}
\caption{Linker-latent cluster profiles. Columns are the 12 K-means clusters
and rows are metadata features, including hierMOF-seg motif-count labels and
whole-linker descriptors. Colors show within-cluster mean standardized
feature values; cluster identifiers are arbitrary.}
\label{fig:si-linker-cluster-profiles}
\end{figure}

The QMOF comparison shows that identical linkers occupy structured but model-dependent regions of latent space (Fig. \ref{fig:si-fixed120-qmof-umap} and Table~\ref{tab:si-mofenc-qmof-clustering}). As training expands from QMOF to the multi-dataset models, the fraction of noise points decreases. The ZINC model also has a higher silhouette than both the QMOF-only and joint linker models. This pattern supports the value of broader multi-dataset pretraining for producing more organized latent representations, consistent with the downstream performance gains. The node and linker heatmaps (Figs.~\ref{fig:si-node-cluster-profiles} and \ref{fig:si-linker-cluster-profiles}) show clusters correlated to elemental/coordination features and hierMOF-seg motifs, respectively. This indicates that the encoders organize node and linker space into distinct and relevant latent regions rather than random points.  To assess the benefits of incorporating RDKit-derived chemical information, we trained an additional encoder without RDKit-derived input features or property-based training objectives. Although this variant achieved a slightly higher silhouette score, it exhibited a substantially higher noise fraction and a lower DBCV score. These results support retaining RDKit-derived chemical information in hierMOF-enc. However, since atom and bond descriptors, property prediction, and contrastive regularization were removed simultaneously, this comparison does not isolate their individual contributions.

\begin{table}[htbp]
\centering
\caption{Density-clustering summary for the same matched QMOF linker records encoded by the three hierMOF-enc models with an additional ablation of RDKit features. HDBSCAN is applied after UMAP to the number of output dimensions selected by the 95\% cumulative explained variance (CEV) criterion. Noise is calculated as $100N_{-1}/N$, where $N_{-1}$ is the number of records assigned HDBSCAN label $-1$ (not assigned to a retained density cluster), and $N$ is the total number of analyzed records, including noise points. Silhouette and density-based clustering validation (DBCV) scores are computed in the intermediate UMAP space, with silhouette evaluated on non-noise points. No RDKit refers to the version without RDKit-derived input features or property-based training objectives.}
\label{tab:si-mofenc-qmof-clustering}
\begin{tabular}{lrrrrr}
\toprule
Encoder training data & PCA dimensions & HDBSCAN clusters & Noise & Silhouette & DBCV \\
\midrule
QMOF & 17 & 35 & 14.6\% & 0.497 & 0.535 \\
QMOF-WS24-hMOF & 20 & 24 & 13.0\% & 0.421 & 0.481 \\
QMOF-WS24-hMOF-ZINC & 30 & 19 & 5.4\% & 0.513 & 0.687 \\
QMOF-WS24-hMOF-ZINC no RDKit & 27 & 40 & 17.9\% & 0.571 & 0.538 \\
\bottomrule
\end{tabular}
\end{table}

\begin{figure}[htbp]
\centering
\includegraphics[width=\textwidth]{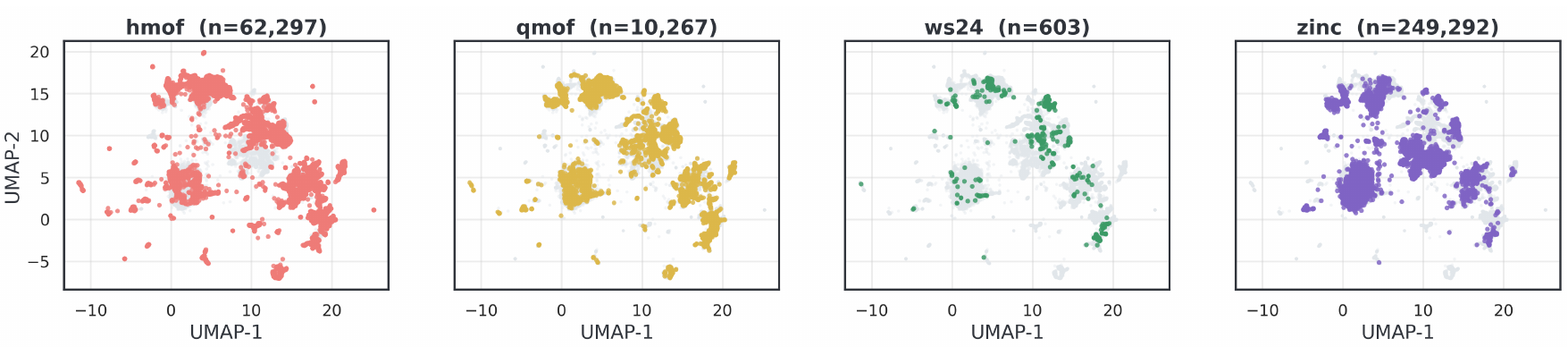}
\caption{Joint molecular representation space obtained using the hierMOF-enc encoder trained on QMOF, WS24, hMOF, and ZINC. The panels highlight hMOF, QMOF, WS24, and ZINC membership, respectively, against the remaining molecules in grey. All panels share the same two-dimensional UMAP coordinates, obtained from encoder representations. Memberships can overlap because a molecular structure may occur in multiple datasets.
}
\label{fig:si-joint-four-source-space}
\end{figure}

To examine chemical-space coverage (see Figure~\ref{fig:si-joint-four-source-space}), we jointly projected MOF-derived linkers from QMOF, hMOF, and WS24 together with ZINC molecules using the encoder trained on all four sources. The source-highlighted projections show that some regions have substantial overlap, while linkers from some datasets occur only in certain regions. Although WS24 linkers occupy several parts shared with the other sources, substantial regions have little WS24 representation. These gaps identify chemical families whose water-stability behavior is poorly represented in the available labeled dataset.

One possible explanation is that some of these chemistries are less compatible with the formation of water-stable frameworks. Alternatively, they may remain underexplored through synthesis, water-stability testing, or reporting. These explanations cannot be distinguished from the projection alone, particularly given the smaller size of WS24 and its inclusion of different stability classes~\cite{Terrones2024-ef}. The sparsely covered regions therefore provide candidates for targeted investigation, rather than evidence that their constituent linkers necessarily produce unstable MOFs.

\clearpage
\section{hierMOF architecture details}
\label{app:mofmod_details}

\subsection{Hierarchy membership and feature initialization}

The cross-level blocks operate on the deterministic membership relations supplied by hierMOF-seg. At the lower interface, atoms belonging to organic linkers are assigned to motifs, while atoms belonging to inorganic fragments are assigned directly to nodes. At the upper interface, motifs are assigned to their parent linkers. These assignments define the parent-child relationships used for cross-level information exchange, while spatial interactions within each resolution are determined independently by the corresponding periodic neighbor graph.

Pretrained token representations are first projected into a common hidden space:
\begin{equation}
    \mathbf v_u
    = W_{\tau(u)}\mathbf h_u+\mathbf b_{\tau(u)}.
    \label{eq:si_token_projection}
\end{equation}
Here, $u$ indexes tokens, $\mathbf h_u$ is the input representation, and $\tau(u)$ identifies its component class: motif, linker, or node. The projection parameters are class-specific.  The projected representation  is concatenated with a learned component-class embedding and fused using a shared network:
\begin{equation}
    \mathbf f_u
    = \operatorname{GELU}\left(
        W_f[
            \operatorname{LN}(\mathbf v_u),
            \operatorname{LN}(E_{\tau}(\tau(u)))
        ]+\mathbf b_f
    \right).
    \label{eq:si_token_fusion_hidden}
\end{equation}
Here, $E_{\tau}$ contains the learned embedding of the component class, and $\operatorname{LN}$ denotes layer normalization.  

The initialized token state is obtained through gated residual fusion:
\begin{equation}
\begin{aligned}
\mathbf h_u^{(0)}
=
\operatorname{LN}
\Big(
    \mathbf v_u
    &+
    \sigma(W_g\mathbf f_u+\mathbf b_g)&\odot
    \tanh(W_r\mathbf f_u+\mathbf b_r)
\Big).
\end{aligned}
\end{equation}

Here, $\odot$ denotes element-wise multiplication. The fusion weights $W_f$, $W_g$, $W_r$ and their biases are shared across component classes. \(\mathbf h_u^{(0)}\) is used for subsequent message passing. The three layer-normalization operations have independent parameters. Padded token states are masked to zero.

\subsection{Cross-level blocks}

At each hierarchical level, a child token \(c\) is associated with a parent token \(p\), and \(\mathcal C(p)\) denotes the set of children belonging to \(p\). The compatibility score is
\begin{equation}
    s(\mathbf h_c,\mathbf h_p)
    = \mathbf w_{\mathrm{att}}^{\mathsf T}
    \operatorname{softplus}\left(
        W_{\mathrm{att}}[\mathbf h_c,\mathbf h_p]
        +\mathbf b_{\mathrm{att}}
    \right)+b_{\mathrm{att}}'.
    \label{eq:si_cross_score}
\end{equation}
The attention parameters are learned independently for each cross-level block. Scores are normalized over children sharing the same parent, giving the pooled child representation:
\begin{equation}
\begin{aligned}
    \alpha_{cp}
    &= \frac{\exp s(\mathbf h_c,\mathbf h_p)}
    {\sum_{c'\in\mathcal C(p)}\exp s(\mathbf h_{c'},\mathbf h_p)},\\
    \mathbf r_p
    &= \sum_{c\in\mathcal C(p)}\alpha_{cp}\mathbf h_c .
\end{aligned}
\label{eq:si_cross_pool}
\end{equation}

The upward message is transformed and incorporated through a feature-wise sigmoid gate:
\begin{equation}
\begin{aligned}
    \mathbf g_p
    &= \sigma\left(
        W_g^\uparrow
        [\mathbf h_p,
        \operatorname{MLP}_\uparrow(\mathbf r_p)]
        +\mathbf b_g^\uparrow\right),\\
    \mathbf h_p^+
    &= \operatorname{BN}\left(
        \mathbf h_p+\mathbf g_p\odot
        \operatorname{MLP}_\uparrow(\mathbf r_p)\right).
\end{aligned}
\label{eq:si_cross_upward}
\end{equation}
The updated parent then sends a transformed message to each child, using its current state after any preceding normalization. The downward update is
\begin{equation}
\begin{aligned}
    \mathbf g_c
    &= \sigma\left(
        W_g^\downarrow
        [\mathbf h_c,
        \operatorname{MLP}_\downarrow(\mathbf h_p^+)]
        +\mathbf b_g^\downarrow\right),\\
    \mathbf h_c^+
    &= \operatorname{BN}\left(
        \mathbf h_c+\mathbf g_c\odot
        \operatorname{MLP}_\downarrow(\mathbf h_p^+)\right).
\end{aligned}
\label{eq:si_cross_downward}
\end{equation}

Here, $\sigma$ is the sigmoid function, and $W_g$ and $\mathbf b_g$ are learned gate parameters. Each MLP contains two linear layers with a Softplus hidden activation. Message, gate, and normalization parameters are independent between directions and cross-level blocks. Masked batch normalization uses current-batch statistics during training, and uses saved running means and variances during evaluation. Each normalization acts on all valid tokens, including those receiving no residual message.

\subsection{Three-body directed crystal graph convolution}
\label{app:three_body_directed_cgcnn}

The same convolutional form is used at the atom, motif-node, and linker-node resolutions, with independent parameters and neighbor cutoffs for each graph. For receiver \(i\), up to \(M\) periodic neighbors within cutoff \(r_{\mathrm c}\) are retained as directed edges \(j\!\to\!i\). The zero-translation self-edge is excluded.

\subsubsection{Radial and angular features}
For receiver $i$, let $\mathcal N(i)$ denote its retained incoming
periodic edges. Indices $j$ and $k$ identify neighbor images.
The distance $d_{ji}$ is expanded in $Q$ Gaussian radial basis functions:
\begin{equation}
    e_{ji,q}
    =
    \exp\left(-\frac{(d_{ji}-\mu_q)^2}{\Delta^2}\right),
    \qquad q=1,\ldots,Q.
    \label{eq:si_radial_basis}
\end{equation}
Here, $\mu_q$ are uniformly spaced centers and $\Delta$ determines
their spacing and Gaussian width. The radial feature vector is
$\mathbf e_{ji}=[e_{ji,1},\ldots,e_{ji,Q}]$.

For two distinct incoming periodic edges $j\to i$ and $k\to i$,
the angle between their displacement vectors is expanded as
\begin{equation}
    \mathbf c(\theta_{jik})
    =
    [\cos(\theta_{jik}),\ldots,\cos(K_\theta\theta_{jik})].
    \label{eq:si_angle_basis}
\end{equation}
Here, $K_\theta$ is the number of cosine angular harmonics.
The triplet state is

\begin{equation}
\begin{aligned}
\mathbf t_{jik}
=
\operatorname{softplus}\Big(
W_o\operatorname{softplus}\big(
    &W_i\mathbf h_i
    +W_n(\mathbf h_j+\mathbf h_k)\\
    &+W_r(\mathbf e_{ji}+\mathbf e_{ki})
    +W_\theta\mathbf c(\theta_{jik})
    +\mathbf b
\big)
+\mathbf b_o
\Big).
\end{aligned}
\end{equation}

The matrices and biases are learned parameters. The neighbor
projection $W_n$ and radial projection $W_r$ are shared between
the two neighbor edges.

Contributions are smoothly weighted using the neighbor cutoff
$r_{\mathrm c}$:
\begin{equation}
    f_{\mathrm c}(d)
    =
    \begin{cases}
        \frac{1}{2}\left[\cos(\pi d/r_{\mathrm c})+1\right],
        & 0\leq d\leq r_{\mathrm c},\\
        0, & d>r_{\mathrm c}.
    \end{cases}
    \label{eq:si_cosine_cutoff}
\end{equation}
The angular context associated with edge $j\to i$ is
\begin{equation}
\begin{aligned}
    w_{jik}
    &= f_{\mathrm c}(d_{ji})f_{\mathrm c}(d_{ki}),\\
    \mathbf a_{ji}
    &=
    \frac{
        \sum_{k\in\mathcal N(i)\setminus\{j\}}
        w_{jik}\mathbf t_{jik}
    }{
        \max\left(
            1,\sum_{k\in\mathcal N(i)\setminus\{j\}}w_{jik}
        \right)
    }.
\end{aligned}
    \label{eq:si_angular_context}
\end{equation}
This context is combined with the receiver, sender and radial features in the directed message update.

\subsubsection{Directed message update}

The sender, receiver, radial, and angular representations are projected into filter and signal channels,
\begin{equation}
    \begin{bmatrix}
        \mathbf f_{ji}\\
        \mathbf s_{ji}
    \end{bmatrix}
    =
    \operatorname{BN}
    \left(
        W_{\mathrm e}
        \left[
            \mathbf h_i
            ,
            \mathbf h_j
            ,
            \mathbf e_{ji}
            ,
            \mathbf a_{ji}
        \right]
        +
        \mathbf b_{\mathrm e}
    \right).
    \label{eq:si_edge_channels}
\end{equation}
The directed message is
\begin{equation}
    \mathbf m_{j\to i}
    =
    \sigma(\mathbf f_{ji})
    \odot
    \operatorname{softplus}(\mathbf s_{ji}),
    \label{eq:si_directed_message}
\end{equation}
and the receiver state is updated as
\begin{equation}
    \mathbf h_i^{\prime}
    =
    \operatorname{softplus}
    \left(
        \mathbf h_i
        +
        \operatorname{BN}
        \left(
            \sum_{j\in\mathcal N(i)}
            \mathbf m_{j\to i}
        \right)
    \right).
    \label{eq:si_directed_update}
\end{equation}
Here, \(\operatorname{BN}\) denotes masked batch normalization.

\subsection{hierMOF-ALIGNN}

hierMOF-ALIGNN adapts the ALIGNN formulation~\cite{Choudhary2021-rn,CHOUDHARY2022111388} to the atom, motif-node, and linker-node graphs. It retains the pretrained token inputs, component-class fusion, membership-based cross-level exchange, separate token pooling, and lattice Gram embedding.  At each resolution, distances and angle cosines are expanded in Gaussian basis functions and embedded into learned edge and angle representations. Each ALIGNN block first updates the spatial graph, then updates its angular line graph, whose vertices represent directed spatial edges sharing a receiver. Angular neighborhoods are capped by retaining the nearest available partner edges. Whereas three-body CGCNN incorporates angular context directly into spatial messages, the ALIGNN implementation maintains edge and angle states that evolve across successive blocks. Each resolution-specific ALIGNN stack is followed by additional edge-gated graph convolutions without line-graph updates.

At the atomic resolution, cross-level exchange occurs between successive ALIGNN blocks and again after the additional graph convolutions. At the motif-node resolution, motif-linker exchange follows each ALIGNN block, before the additional graph convolutions. The linker-node stack and its graph convolutions are followed by the final linker-to-motif update. Both variants then use the same pooling and prediction modules.

\clearpage

\subsection{hierMOF hyperparameters}
Target values were standardized using training-set statistics. Pretrained latent features were standardized separately for each component class and feature using the training split. Periodic image bounds specify integer lattice translations in each direction. Structures exceeding the token capacity were excluded rather than truncated. The earlier CGCNN baseline and its ablations used a prediction-head width of 128, four hidden layers, 200 epochs, and weight decay $10^{-5}$. The scaling experiments used the same width, depth, and epoch count, with weight decay $5\times10^{-4}$.
\begin{table*}[t]
\centering
\caption{Hyperparameters of the hierMOF models for QMOF
band-gap prediction. Triples refer to the atom, motif-node, and
linker-node resolutions, respectively.}
\label{tab:mofmod_hyperparameters}
\small
\begin{tabular}{lcc}
\toprule
Parameter & hierMOF-CGCNN & hierMOF-ALIGNN \\
\midrule
\multicolumn{3}{l}{\textit{Representations and spatial graphs}} \\
Hidden dimension & 64 & 64 \\
Element-embedding dimension & 64 & 64 \\
Component-class embedding dimension & 32 & 32 \\
Convolutional blocks per resolution & $(2,2,2)$ & $(2,2,2)$ \\
Additional graph-only layers & $(0,0,0)$ & $(1,1,1)$ \\
Neighbor cutoffs (\AA) & $(4,8,15)$ & $(4,8,15)$ \\
Maximum spatial neighbors & $(36,24,24)$ & $(36,24,24)$ \\
\midrule
\multicolumn{3}{l}{\textit{Geometric features}} \\
CGCNN radial start and spacing (\AA) & $(0.1,0.1)$ & - \\
Cosine angular harmonics & $(8,8,8)$ & - \\
Triplet-network hidden dimensions & $(12,16,16)$ & - \\
ALIGNN distance basis functions & - & 80 \\
ALIGNN angle-cosine basis functions & - & 40 \\
ALIGNN feature-embedding dimension & - & 64 \\
Backbone activation & Softplus & SiLU \\
\midrule
\multicolumn{3}{l}{\textit{Pooling and prediction}} \\
Pooled dimension per token class & 64 & 64 \\
Lattice-embedding MLP widths & $(32,32)$ & $(32,32)$ \\
Fusion MLP widths & $(64,64)$ & $(64,64)$ \\
Prediction-head hidden width & 64 & 64 \\
Prediction-head hidden layers & 2 & 2 \\
Lattice and readout activation & GELU & GELU \\
Masked batch-normalization epsilon & $10^{-5}$ & $10^{-5}$ \\
\midrule
\multicolumn{3}{l}{\textit{Training}} \\
Training/validation/test split & $80:10:10$ & $80:10:10$ \\
Batch size & 4 & 4 \\
Training epochs & 220 & 220 \\
Optimizer & AdamW & AdamW \\
Adam coefficients $(\beta_1,\beta_2)$ & $(0.9,0.999)$ & $(0.9,0.999)$ \\
Adam epsilon & $10^{-8}$ & $10^{-8}$ \\
Peak learning rate & $5\times10^{-4}$ & $5\times10^{-4}$ \\
Final learning rate & $2.5\times10^{-5}$ & $2.5\times10^{-5}$ \\
Weight decay & $5\times10^{-4}$ & $5\times10^{-4}$ \\
\bottomrule
\end{tabular}
\end{table*}
\clearpage

\subsection{hierMOF Ablation Study}
\label{app:mofmod_ablation}

The node + linker and node + motif variants in Tables \ref{tab:mofmod_ablation_qmof} and \ref{tab:mofmod_ablation} remove motif and linker tokens, respectively, while retaining the atom graph. Learned token embeddings replace pretrained representations with chemical-type embeddings trained jointly with the predictor. The lattice ablation removes the global Gram descriptor but retains periodic neighbor construction. Pairwise interactions omit angular contributions at all resolutions. The full model uses node encoders pretrained on QMOF, WS24 and hMOF, with ZINC additionally included in linker and motif pretraining. The reduced-pretraining variants either use QMOF alone for both encoders or exclude ZINC from linker pretraining while retaining the node encoder. The downstream QMOF dataset is unchanged.

\begin{table*}[htbp]
    \centering
    \small
    \caption{Ablation study of hierMOF-CGCNN for QMOF band-gap prediction.
    All results use the same seed and training epochs.
    Relative MAE increases are calculated with respect to the full model. The atom-free hierMOF-CGCNN variant removes the atom-level graph while retaining hierarchical information exchange between substructures. In contrast, Flat substructure CGCNN operates on a single graph of nodes, linkers, and motifs without cross-level information exchange.}
    \label{tab:mofmod_ablation_qmof}
    \begin{tabular}{lrr}
        \toprule
        Configuration & Test MAE (eV) & MAE increase (\%) \\
        \midrule
        \textbf{Full hierMOF-CGCNN}             & \textbf{0.2240} & - \\
        QMOF-only encoder pretraining          & 0.2276 & 1.6 \\
        Node + motif hierarchy                & 0.2339 & 4.4 \\
        QMOF + WS24 + hMOF encoder pretraining & 0.2359 & 5.3 \\
        Node + linker hierarchy               & 0.2373 & 5.9 \\
        Pairwise interactions                 & 0.2431 & 8.5 \\
        Without lattice descriptor            & 0.2435 & 8.7 \\
        hierMOF-CGCNN (no atom graph)          & 0.2705 & 20.8 \\
        Flat substructures                     & 0.2917 & 30.2 \\
        Learned token embeddings               & 0.2987 & 33.3 \\
        CGCNN                                  & 0.3250 & 45.1 \\
        \bottomrule
    \end{tabular}
\end{table*}

\begin{table}[htbp]
    \centering
    \small
    \caption{Ablation study of hierMOF-CGCNN for CO$_2$ adsorption at 298~K and 2.5~bar on a 15,000-MOF hMOF subset. Test MAEs are in mol\,kg$^{-1}$, with relative increases over the full model. All variants use the same 80:10:10 split and one seed. CGCNN L approximately matches the parameter count of hierMOF-CGCNN, whereas CGCNN S retains its atom-level   graph settings without hierarchical or three-body interactions. hierMOF-CGCNN contains 492,024 trainable parameters. The atom-only CGCNN L uses five layers of width 140 and 492,059 parameters, approximately matching the full model. CGCNN S uses two layers of width 64 and 64,193 parameters, retaining the atom-level graph settings of hierMOF-CGCNN.}
    \label{tab:mofmod_ablation}
    \begin{tabular}{lrr}
        \toprule
        Configuration & Test MAE & MAE increase (\%) \\
        \midrule
        \textbf{Full hierMOF-CGCNN}             & \textbf{0.4988} & - \\
        hierMOF-enc pretraining without ZINC       & 0.5041 & 1.1 \\
        Without lattice descriptor            & 0.5123 & 2.7 \\
        Node + motif hierarchy                & 0.5263 & 5.5 \\
        Pairwise convolutions                 & 0.5492 & 10.1 \\
        Node + linker hierarchy               & 0.6173 & 23.8 \\
        Learned token embeddings              & 0.8070 & 61.8 \\
        CGCNN L                               & 1.1092 & 122.4 \\
        CGCNN S                               & 1.3583 & 172.3 \\
        \bottomrule
    \end{tabular}
\end{table}

\clearpage
\subsection{hierMOF benchmarking}
\label{app:mofmod_bench}

For hMOF adsorption prediction, separate models were trained for each
gas and pressure using an 80/10/10 training, validation, and test split.
Table~\ref{tab:hmof_literature_comparison} compares single-seed test
MAEs for CO$_2$ and CH$_4$ adsorption at 298~K with published results. Gas, temperature, and pressure identify separate tasks and were not supplied as additional input features.

\begin{table}[ht]
\centering
\caption{Comparison of reported QMOF band-gap prediction performance.}
\label{tab:qmof_bandgap_comparison}
\begin{tabular}{lcc}
\hline
Model & Architecture & MAE (eV) \\
\hline
CGCNN & Crystal GNN & $0.325$ \\
ALIGNN~\cite{Choudhary2021-rn,10.1039/d4dd00014e} & Angular GNN & $0.224$ \\
MOFormer~\cite{Cao2023-pm} & Transformer & $0.367 \pm 0.005$ \\
MOFTransformer$_{\mathrm{scratch}}$~\cite{Kang2023-zy, Cao2023-pm} & Transformer & $0.271$ \\
MOFTransformer$_{\mathrm{pretrain}}$~\cite{Kang2023-zy,Cao2023-pm} & Transformer & $0.224$ \\
MGT~\cite{10.1039/d4dd00014e} & Graph Transformer & $0.240$ \\
\hline
hierMOF CGCNN & Hierarchical GNN & \textbf{0.224} $\pm 0.004$ \\
hierMOF ALIGNN & Hierarchical GNN & \textbf{0.214}$ \pm 0.003$ \\

\hline
\end{tabular}
\end{table}

\begin{table}[ht]
\centering
\small
\caption{Comparison of reported test MAEs for adsorption prediction on the hMOF dataset. Evaluation subsets and splits differ between studies, and the values are therefore not strictly directly comparable. All values are in mol\,kg$^{-1}$.}
\label{tab:hmof_literature_comparison}
\begin{tabular}{lcccc}
\toprule
Model &
CO$_2$ 0.01 bar &
CO$_2$ 0.05 bar &
CO$_2$ 2.5 bar &
CH$_4$ 0.5 bar \\
\midrule
CGCNN~\cite{Teng2025-lx}
& 0.044 & 0.115 & 0.769 & 0.121 \\

MMAGNN~\cite{Teng2025-lx}
& 0.037 & 0.093 & 0.465 & - \\

ALIGNN~\cite{Teng2025-lx}
& 0.040 & - & 0.480 & - \\

CGCNN$_{\mathrm{pretrain}}$~\cite{Cao2023-pm}
& - & 0.110 & 0.645 & 0.099 \\

SOAP~\cite{Cao2023-pm}
& - & 0.115 & 0.666 & 0.106 \\

\midrule
\textbf{hierMOF-CGCNN}
& \textbf{0.0317}
& \textbf{0.0789}
& \textbf{0.3462}
& \textbf{0.0586} \\

\textbf{hierMOF-ALIGNN}
& \textbf{0.0339}
& \textbf{0.0797}
& \textbf{0.3682}
& \textbf{0.0639} \\

\bottomrule
\end{tabular}
\end{table}

\clearpage
\subsection{Water-stability classification}
hierMOF was adapted to WS24 by replacing the scalar regression output with four logits while retaining the backbone, pooling, and hierarchical representations. The classes, in order, are unstable, low kinetic stability, high kinetic stability, and thermodynamic stability. Class probabilities are obtained by softmax,
\begin{equation}
    p_{mc}=\frac{\exp(\ell_{mc})}
    {\sum_{k=1}^{4}\exp(\ell_{mk})}.
    \label{eq:si_ws24_softmax}
\end{equation}
Here, $m$ indexes MOFs, $c$ indexes the four classes, and $\ell_{mc}$ is the corresponding output logit. The predicted class is
\begin{equation}
    \widehat y_m=\operatorname*{arg\,max}_{c\in\{1,2,3,4\}}p_{mc}.
    \label{eq:si_ws24_prediction}
\end{equation}
Labels were treated categorically, without target standardization or an ordinal loss. Class imbalance was addressed by weighted cross-entropy, with label smoothing:
\begin{equation}
    \mathcal L_{\mathrm{cls}}^{(m)}
    =
    -w_{y_m}
    \sum_{c=1}^{4}
    \left[
        (1-\varepsilon)\mathds{1}(c=y_m)
        +\frac{\varepsilon}{4}
    \right]\log p_{mc}.
    \label{eq:si_ws24_loss}
\end{equation}
Here, \(y_m\) is the observed class of MOF \(m\), \(\mathds{1}\) is the indicator function, and \(\varepsilon\) is the label-smoothing coefficient. Class weights were calculated from training labels as \(w_c=N_{\mathrm{tr}}/(4n_c)\), where \(N_{\mathrm{tr}}\) is the training-set size and \(n_c\) is the number of training examples in class \(c\). The training loss is averaged over examples without normalization by the sum of their weights. Validation and test cross-entropy are unweighted and unsmoothed.

The runs reported in Table~\ref{tab:ws24_comparison} use class-stratified 80/10/10 splits and three seeds. All four models share the same partitions within each seed, while both partition membership and predictor initialization vary between seeds.  No experimental exposure conditions were supplied as input features. Accuracy is the fraction of correct predictions. Macro-F1 is the unweighted mean of the four class-specific F1 scores, and balanced accuracy is the mean class recall. Reported uncertainties are sample standard deviations across the three runs, not confidence intervals.

\begin{table}[t]
\centering
\small
\caption{WS24 water-stability classification performance.
Our results are mean $\pm$ sample standard deviation over three
seeds, using shared 80:10:10 train/validation/test splits within
each seed on retained MOFs. }
\label{tab:ws24_comparison}
\begin{tabular}{lccc}
\toprule
Model & Accuracy (\%) & Macro-F1 & Balanced accuracy (\%) \\
\midrule
hierMOF-ALIGNN
& $\mathbf{68.40 \pm 7.38}$
& $0.585 \pm 0.116$
& $56.66 \pm 8.73$ \\

hierMOF-CGCNN
& $66.23 \pm 2.25$
& $\mathbf{0.591 \pm 0.040}$
& $\mathbf{59.72 \pm 7.00}$ \\

ALIGNN
& $60.61 \pm 7.38$
& $0.502 \pm 0.086$
& $50.77 \pm 8.25$ \\

CGCNN
& $56.28 \pm 5.41$
& $0.482 \pm 0.065$
& $49.42 \pm 4.92$ \\

\bottomrule
\end{tabular}
\end{table}

\clearpage

\section{hierMOF-int}

hierMOF-int evaluates the dependence of trained hierMOF predictions on normalized pretrained motif, linker, and inorganic-node representations before projection and message passing. For \(T\) tokens, these representations are collected in \(H=(\mathbf h_1,\ldots,\mathbf h_T)\). Writing the normalized prediction as
\begin{equation}
    \widetilde y = F(H,\Omega),
\end{equation}
\(\Omega\) collects the fixed atomic inputs, coordinates, periodic cell, connectivity, and hierarchical assignments. Model parameters and all other inputs remain fixed during attribution, while downstream hidden states are recomputed whenever token representations change.

\subsection{Reference representations}

Reference vectors are computed as occurrence-weighted means of normalized training-set representations, separately for each component class \(c\):
\begin{equation}
    \boldsymbol{\mu}_c
    =
    \frac{1}{N_c}
    \sum_{n=1}^{N_c}
    \mathbf h_n^{(c)},
\end{equation}
where \(N_c\) is the number of training-token occurrences belonging to class \(c\), and \(\mathbf h_n^{(c)}\) is the normalized representation of its \(n\)-th occurrence. The classes are motifs, linkers, and nodes. References are not conditioned on chemical identity.

For a selected token \(t\), the reference \(H_t^0\) is constructed by replacing its representation with the corresponding component-class mean \(\boldsymbol{\mu}_{\tau(t)}\), where \(\tau(t)\) identifies its class:
\begin{equation}
    H_t^0
    =
    \left(
    \mathbf h_1,\ldots,\mathbf h_{t-1},
    \boldsymbol{\mu}_{\tau(t)},
    \mathbf h_{t+1},\ldots,\mathbf h_T
    \right).
\end{equation}
Each physical substructure has one input token in the reference cell. Periodicity is represented through graph connectivity rather than explicitly replicated tokens. All other token representations and structural inputs remain unchanged. Motif replacement retains the parent-linker input, while linker replacement retains the motif inputs. These representation-level interventions are not hierarchy-consistent chemical substitutions, and the mean references need not correspond to realizable fragments.

\subsubsection{Conditional Integrated Gradients}

Conditional Integrated Gradients (IG)~\cite{Sundararajan2017} interpolates between the reference and observed token representations:
\begin{equation}
    H_t(\alpha)
    =
    H_t^0
    +
    \alpha\left(H-H_t^0\right),
    \qquad
    \alpha\in[0,1].
\end{equation}
Only the selected token \(t\) changes along the path. The normalized attribution of feature \(\ell\) of token \(t\) is
\begin{equation}
    a_{t\ell}
    =
    \left(
    h_{t\ell}-\mu_{\tau(t),\ell}
    \right)
    \int_0^1
    \left.
    \frac{\partial F(H,\Omega)}
    {\partial h_{t\ell}}
    \right|_{H=H_t(\alpha)}
    \,\mathrm d\alpha,
\end{equation}
where \(h_{t\ell}\) and \(\mu_{\tau(t),\ell}\) denote the corresponding components of the observed representation and its reference mean.

The integral is approximated using \(K\)-point midpoint quadrature:
\begin{equation}
\begin{aligned}
    a_{t\ell}
    &\approx
    \frac{h_{t\ell}-\mu_{\tau(t),\ell}}{K}
    \sum_{j=1}^{K}
    \left.
    \frac{\partial F(H,\Omega)}
    {\partial h_{t\ell}}
    \right|_{H=H_t(\alpha_j)},\\
    \alpha_j
    &=
    \frac{j-\frac12}{K}.
\end{aligned}
\end{equation}
We use \(K=256\) midpoint steps. The normalized token attribution is obtained by summing over its input features and converted to physical property units using the target standard deviation \(\sigma_y\):
\begin{equation}
    a_t
    =
    \sum_{\ell}a_{t\ell},
    \qquad
    A_t
    =
    \sigma_y a_t.
\end{equation}
By IG completeness, \(a_t\) equals the difference between the normalized predictions \(F(H,\Omega)\) and \(F(H_t^0,\Omega)\), up to numerical integration error. This provides a numerical consistency check rather than an independent test of chemical faithfulness. Scores from separate conditional interventions are not additive. Joint IG is reported separately for comparison and replaces all eligible token representations with their component-class means.

\clearpage

\subsection{Assessment of hierMOF-int attributions}
\label{sec:si_mofint_validation}
\label{sec:si_mofint_alignn_validation}

Attributions were evaluated on the QMOF test sets for hierMOF-CGCNN and hierMOF-ALIGNN. Predictor weights, pretrained encoders, and saved normalization statistics were fixed. Reference representations were estimated solely from training inputs. 

\paragraph{Reproducibility and model dependence.}
Attribution seeds 0 and 1 were compared for the same checkpoint and test structures using matched physical-instance scores, pooled Spearman correlation, and sign agreement (Table~\ref{tab:si_mofint_behavior}). Parameter randomization \cite{Adebayo2018} replaced either the final linear output layer or all floating predictor parameters with zero-mean Gaussian draws scaled by each original tensor's standard deviation, with a 0.02 fallback for near-constant tensors. Encoders and saved running moments were retained. Output-layer randomization retained strong signed attribution agreement for CGCNN (\(\rho=0.901\)) but not ALIGNN (\(\rho=-0.079\)), showing that resetting the final layer alone does not necessarily remove attribution rankings inherited from the trained representation. Full-predictor randomization reduced agreement for both models (\(\rho=0.080\) and \(0.067\), respectively), supporting dependence on learned predictor parameters.

\begin{table}[ht]
\centering
\small
\setlength{\tabcolsep}{4.5pt}
\renewcommand{\arraystretch}{1.12}
\caption{Fixed-predictor repeats and conditional-IG parameter randomization.}
\label{tab:si_mofint_behavior}
\begin{tabular}{lrrrr}
\toprule
& \multicolumn{2}{c}{CGCNN} & \multicolumn{2}{c}{ALIGNN} \\
\cmidrule(lr){2-3}\cmidrule(lr){4-5}
Comparison & \(\rho\) & Sign (\%) & \(\rho\) & Sign (\%) \\
\midrule
Conditional IG repeat        & 1.000 & 100.0 & 1.000  & 100.0 \\
Output-layer randomization   & 0.901 & 92.6  & -0.079 & 50.6 \\
Full-predictor randomization & 0.080 & 53.1  & 0.067  & 52.0 \\
\bottomrule
\end{tabular}
\par\smallskip
\begin{minipage}{0.98\linewidth}
\footnotesize Comparisons use 75,276 CGCNN or 75,590 ALIGNN matched physical
instances. \(\rho\) is the pooled Spearman correlation of signed scores.
Sign agreement is the percentage of matched scores with the same sign.
\end{minipage}
\end{table}

\paragraph{Numerical integration.}
Conditional-IG completeness~\cite{Sundararajan2017} was assessed using
\begin{equation}
    r_t
    =
    \frac{
    \left|
    F(H,\Omega)-F(H_t^0,\Omega)-a_t
    \right|
    }{
    \max\left(
    \left|F(H,\Omega)-F(H_t^0,\Omega)\right|,
    10^{-6}
    \right)
    }.
\end{equation}
Here, \(a_t\) is the conditional-IG attribution in normalized target units. The tolerance was \(r_t\leq 0.05\). Joint IG instead compares the summed attributions with the prediction difference relative to the joint reference (Table~\ref{tab:si_mofint_numerical_summary}). Integration sensitivity was assessed at 128 and 256 steps for both predictors, and at 256 and 512 steps. A score was considered stable when
\begin{equation}
    \left|A_t^{(n)}-A_t^{(m)}\right|
    \leq
    10^{-4}\,\mathrm{eV}
    +
    0.05\left|A_t^{(n)}\right|,
    \qquad n>m,
\end{equation}
where \(n\) and \(m\) denote the numbers of integration steps.

\begin{table}[ht]
\centering
\small
\setlength{\tabcolsep}{4.5pt}
\renewcommand{\arraystretch}{1.12}
\caption{Numerical IG completeness.}
\label{tab:si_mofint_numerical_summary}
\begin{tabular}{llrrr}
\toprule
Predictor & Method & Steps & 95th residual (\%) & Within 5\% (\%) \\
\midrule
CGCNN & Conditional IG & 256 & 1.496 & 98.17 \\
CGCNN & Joint IG       & 256 & 3.653 & 96.42 \\
CGCNN & Conditional IG & 512 & 0.358 & 99.49 \\
CGCNN & Joint IG       & 512 & 0.540 & 99.18 \\
ALIGNN & Conditional IG & 256 & 1.133 & 98.76 \\
ALIGNN & Joint IG       & 256 & 1.522 & 98.21 \\
ALIGNN & Conditional IG & 512 & 0.464 & 99.46 \\
ALIGNN & Joint IG       & 512 & 0.526 & 99.49 \\
\bottomrule
\end{tabular}
\par\smallskip
\begin{minipage}{0.98\linewidth}
\footnotesize Residuals measure the discrepancy between the summed IG scores
and the corresponding prediction difference, not error against the target
property. Conditional IG has one observation per physical instance and joint
IG one per MOF.
\end{minipage}
\end{table}

\clearpage

\subsection{hierMOF-int analysis of QMOF band gap}
\label{app:mofint_results}

\begin{figure*}[ht]
    \centering
    \includegraphics[width=0.99\linewidth]{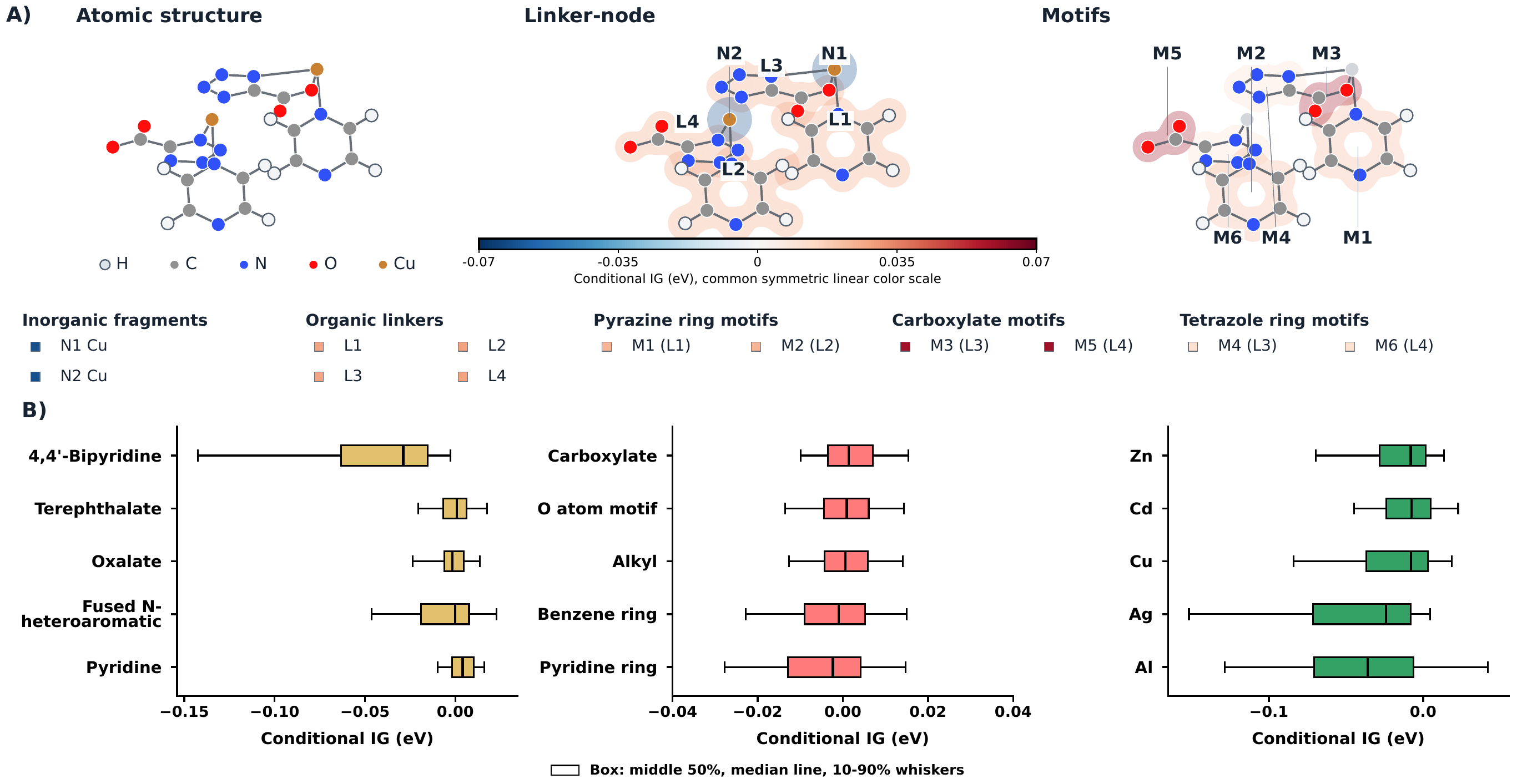}
    \caption{\textbf{Interpretation of MOF band-gap predictions using hierMOF-int for the hierMOF-ALIGNN model.} \textbf{A)} Example decomposition of a Cu framework \texttt{qmof-d64d83c}. Colored backgrounds show each substructure's conditional IG contribution to the band gap. The same symmetric linear color scale is used across all hierarchical levels. Labels identify individual inorganic fragments (N), organic linkers ((L)), and motifs ((M)), with each motif associated with its parent linker. Other hierarchy-level inputs remain fixed. \textbf{B)} Attribution distributions for the five most frequent eligible linkers (left), motif types (middle), and metals present in inorganic fragments (right), across test set MOFs. Node scores are pooled by present metal, not individual node topology. Boxes span the 25th to 75th percentiles, internal lines indicate medians and whiskers span the 10th to 90th percentiles. Scores are reference-dependent and nonadditive. 
    }
    \label{fig:mofint_example}
\end{figure*}
\begin{figure*}[ht]
    \centering
    \includegraphics[width=0.99\linewidth]{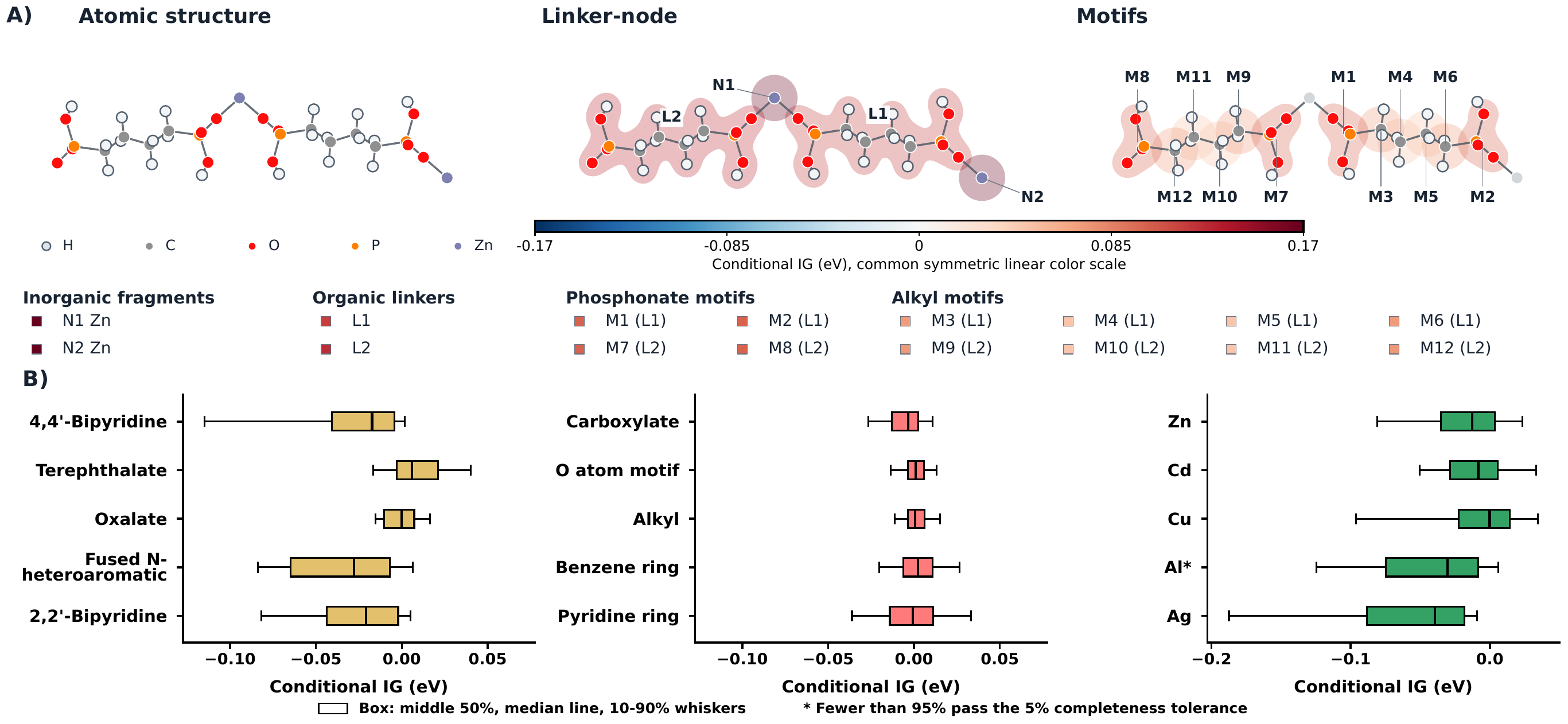}
    \caption{\textbf{Interpretation of MOF band-gap predictions using hierMOF-int for the hierMOF-CGCNN model.} \textbf{A)} Example decomposition of a Zn-phosphonate framework qmof-5a9cc08 into the atomic structure, organic linkers, nodes, and linker motifs. Colored backgrounds show the conditional IG contribution of each physical substructure to the band gap. The same symmetric linear color scale is used across all hierarchical levels. Labels identify individual inorganic fragments ((N)), organic linkers ((L)), and motifs ((M)), with each motif associated with its parent linker. Other hierarchy-level inputs remain fixed. \textbf{B)} Attribution distributions for the five most frequent eligible linkers (left), motif types (middle) and metals present in inorganic fragments (right), across test set MOFs. Node scores are pooled by present metal, not individual node topology. Boxes span the 25th to 75th percentiles, internal lines indicate medians and whiskers span the 10th to 90th percentiles.
    }
    \label{fig:mofint_example_cgcnn}
\end{figure*}

\begin{figure*}[p]
    \centering
    \includegraphics[
        page=1,
        width=0.99\linewidth
    ]{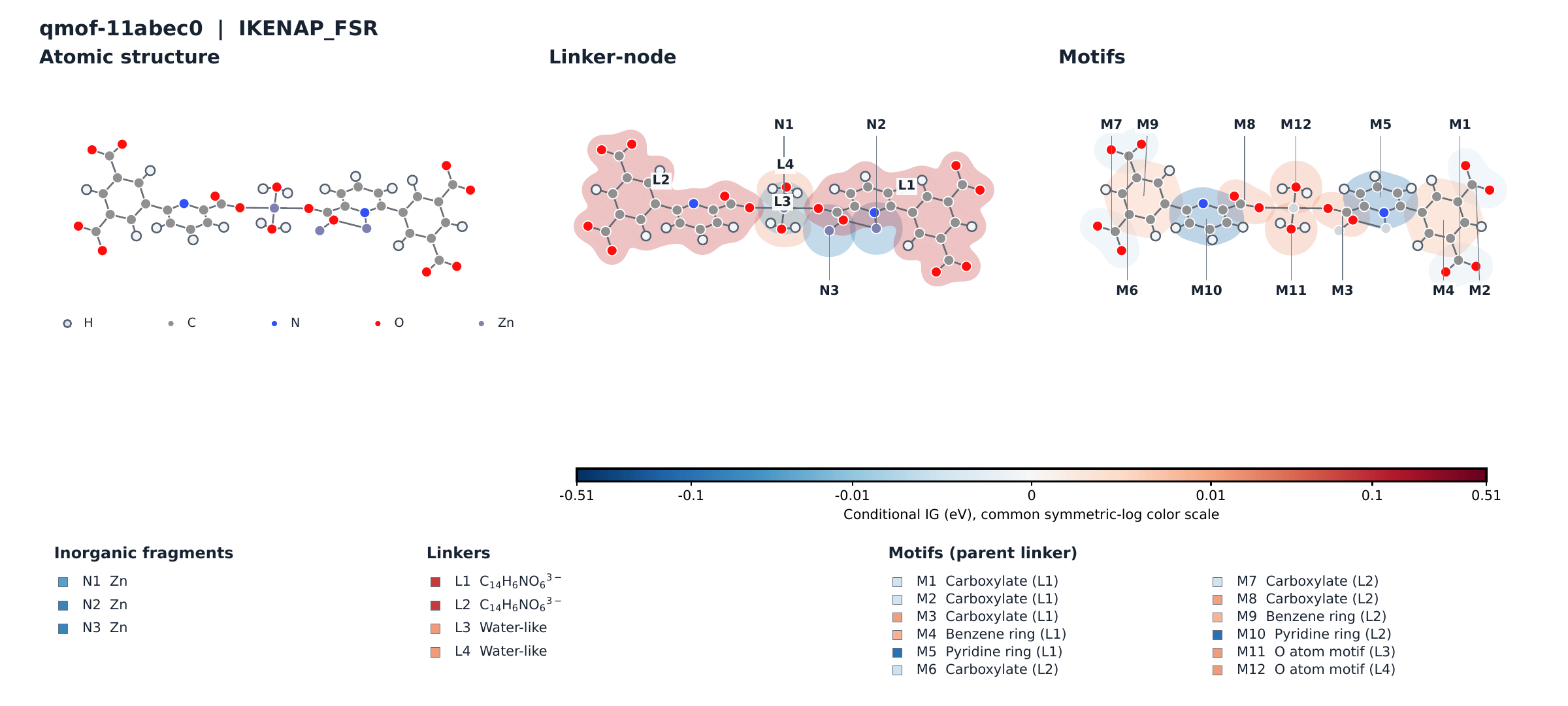}
    \caption{\textbf{hierMOF-int attribution maps for eight selected held-out QMOF structures evaluated using hierMOF-ALIGNN.}
    Each example shows the atomic structure (left), node-linker attributions (center), and linker-motif attributions (right). Blue and red indicate negative and positive conditional IG scores relative to the corresponding hierarchy-level training-mean reference, with other hierarchy inputs fixed. All examples share color scale from \(-0.51\) to \(+0.51\)~eV, linear within \(\pm 0.01\)~eV. Labels \(N\), \(L\), and \(M\) identify inorganic fragments, linkers, and motifs, respectively. Motif labels specify their parent linker. Scores are reference-dependent and nonadditive and should not be interpreted as causal chemical-substitution effects.}
    \label{fig:mofint_alignn_examples}
\end{figure*}

The reported QMOF analyses used 1,954 structures (full test set). Illustrative structures were selected separately for readable geometry and prediction errors below 0.1~eV. Additional hierMOF-int analyses complement the hierMOF-ALIGNN results presented in the main text. Figure~\ref{fig:mofint_alignn_examples} shows conditional IG attribution maps for a broader set of QMOF structures, illustrating how linker, motif, and inorganic-node contributions vary across chemically distinct frameworks. Figure~\ref{fig:mofint_example} presents the corresponding analysis for hierMOF-CGCNN.

\begin{figure*}[p]
    \centering
    \includegraphics[page=2,width=0.99\linewidth]
        {figures/MOFINT_ALIGNN_examples.pdf}
\end{figure*}

\begin{figure*}[p]
    \centering
    \includegraphics[page=3,width=0.99\linewidth]
        {figures/MOFINT_ALIGNN_examples.pdf}
\end{figure*}

\begin{figure*}[p]
    \centering
    \includegraphics[page=4,width=0.99\linewidth]
        {figures/MOFINT_ALIGNN_examples.pdf}
\end{figure*}

\begin{figure*}[p]
    \centering
    \includegraphics[page=5,width=0.99\linewidth]
        {figures/MOFINT_ALIGNN_examples.pdf}
\end{figure*}

\begin{figure*}[p]
    \centering
    \includegraphics[page=6,width=0.99\linewidth]
        {figures/MOFINT_ALIGNN_examples.pdf}
\end{figure*}

\begin{figure*}[p]
    \centering
    \includegraphics[page=7,width=0.99\linewidth]
        {figures/MOFINT_ALIGNN_examples.pdf}
\end{figure*}

\begin{figure*}[p]
    \centering
    \includegraphics[page=8,width=0.99\linewidth]
        {figures/MOFINT_ALIGNN_examples.pdf}
\end{figure*}

\subsection{hierMOF-int analysis of hMOF adsorption}
\label{SI-app:mofint_hmof}

The reported hMOF analyses used 855 test structures. Illustrative structures were selected separately for readable geometry and prediction errors below 0.1~mol/kg. Conditional IG was evaluated for CO$_2$ uptake at 298~K and 2.5~bar using 256 midpoint steps and training-mean references for each token kind. Population plots weight physical instances equally and rank groups by distinct-MOF frequency (minimum 20 MOFs). Linkers are connected, carbon-containing graphs with at least six heavy atoms. Graph aliases are pooled only if hydrogen counts and formal charge agree. Node scores are pooled by metal presence and describe whole fragments.

For comparisons between predictors, we use the same held-out structures and retain scores that satisfy the 5\% relative completeness criterion. Mean alkyne scores correlate negatively with pore-limiting diameter (ALIGNN $\rho=-0.462$, CGCNN $\rho=-0.301$). For each parent linker containing both motif types, the alkyl-minus-alkyne contrast is the mean conditional IG score of its alkyl motifs minus the mean score of its alkyne motifs. We average these linker-level differences within each MOF and then weight MOFs equally in the reported mean. The mean contrast is positive for both predictors (Table~\ref{tab:si_hmof_design}). This means that alkyl motifs received higher attribution scores on average in this matched comparison.

\begin{table}[htbp]
\centering\small
\caption{Matched-context conditional-IG summaries. Pore correlations use
per-MOF mean alkyne scores. Alkyl-minus-alkyne contrasts are averaged within
parent linkers and then within MOFs. Both models use the same 64 MOFs and
the same completeness-passing instances for the contrast. Intervals use
3,000 whole-MOF bootstrap resamples.}
\label{tab:si_hmof_design}
\begin{tabular}{lrr}
\toprule
Statistic & CGCNN & ALIGNN \\
\midrule
Mean alkyl-minus-alkyne contrast (mol/kg) & 0.0678 & 0.1062 \\
95\% bootstrap interval & $[0.0463,0.0885]$ & $[0.0724,0.1428]$ \\
Alkyne score vs. pore diameter, Spearman $\rho$ & $-0.301$ & $-0.462$ \\
MOFs with both carboxylate score signs & 421/842 & 503/843 \\
\bottomrule
\end{tabular}
\end{table}

\begin{figure}[p]
\centering
\includegraphics[width=0.48\linewidth]{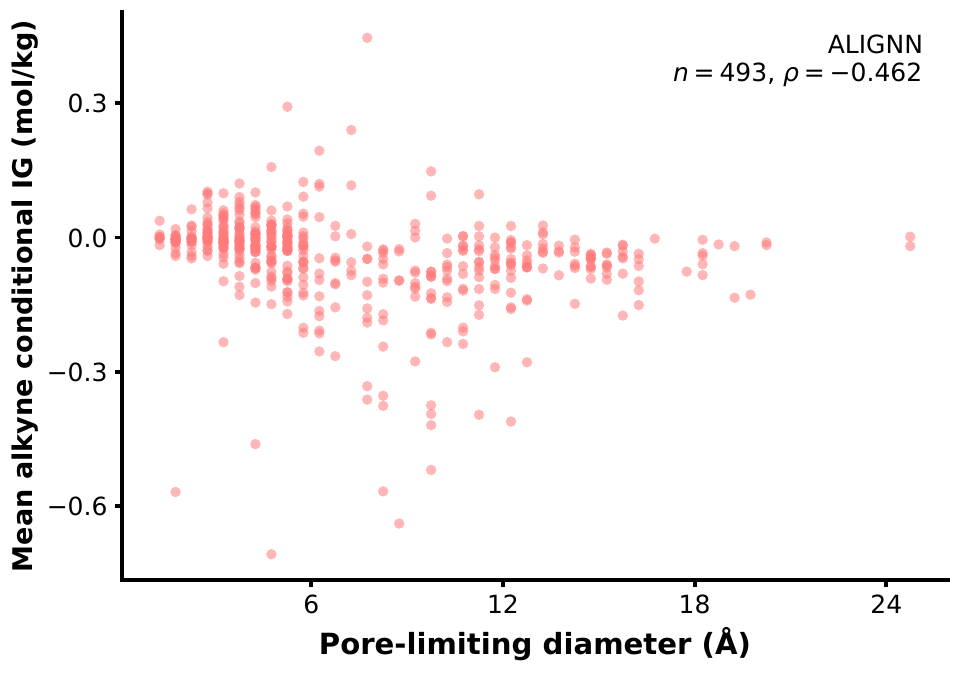}%
\includegraphics[width=0.48\linewidth]{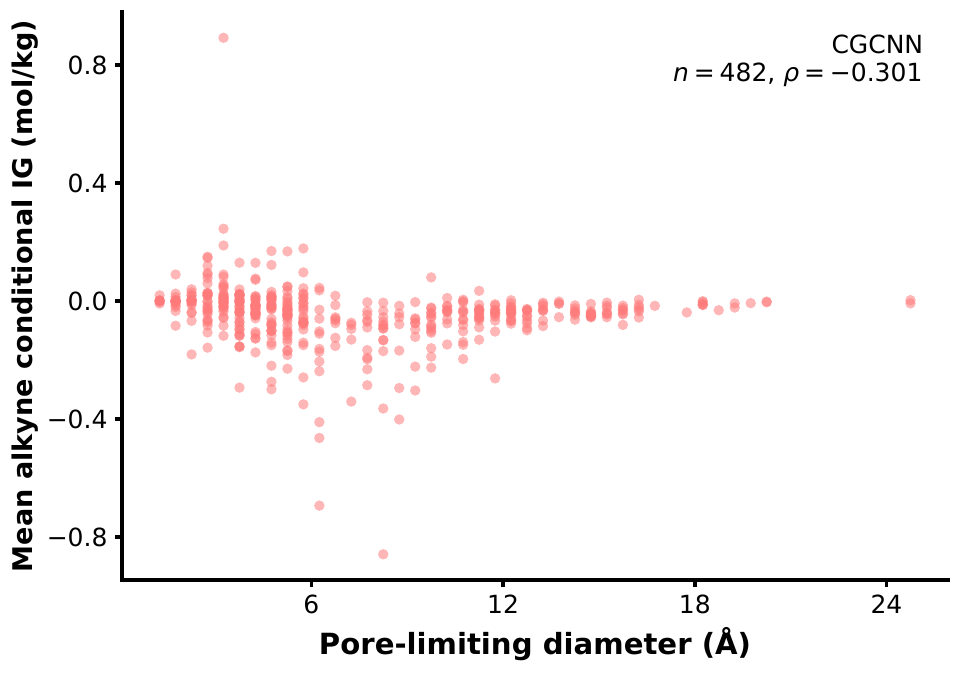}\par\medskip
\includegraphics[width=0.48\linewidth]{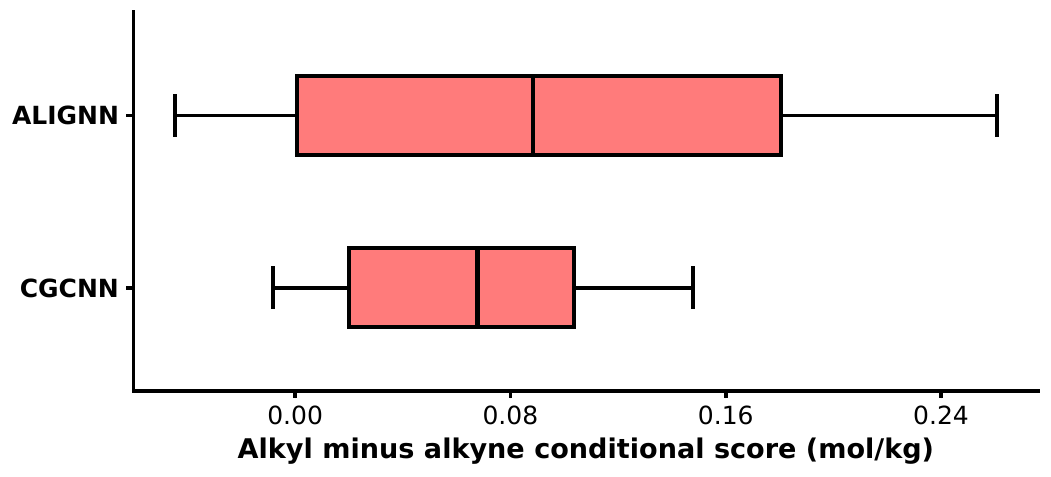}%
\includegraphics[width=0.48\linewidth]{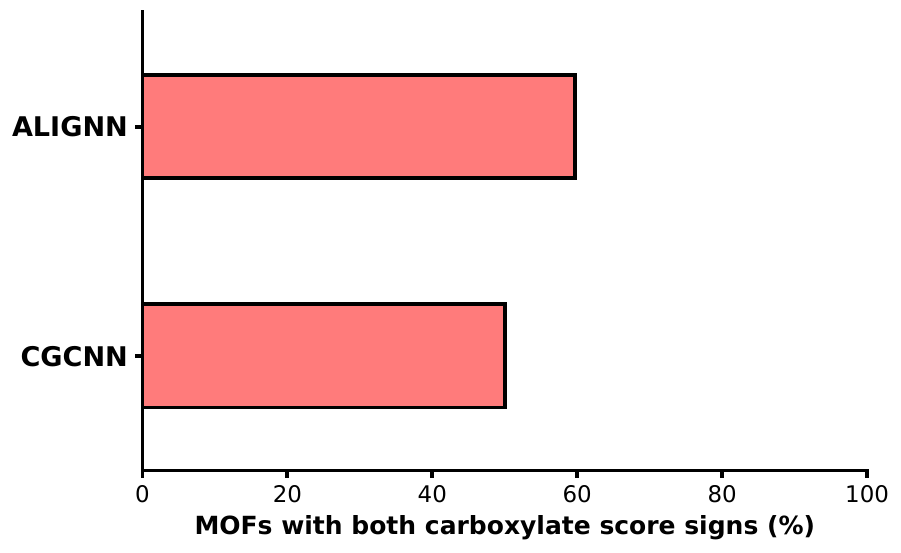}
\caption{Matched hMOF context analyses. Each point in the top row is one MOF. The bottom row shows the within-parent alkyl-minus-alkyne contrasts and the fraction of MOFs with opposite carboxylate signs among those with at least two qualifying scores exceeding 0.001~mol/kg in magnitude.}
\label{fig:si_hmof_context}
\end{figure}

\begin{figure}[p]
\centering
\includegraphics[width=0.91\linewidth]{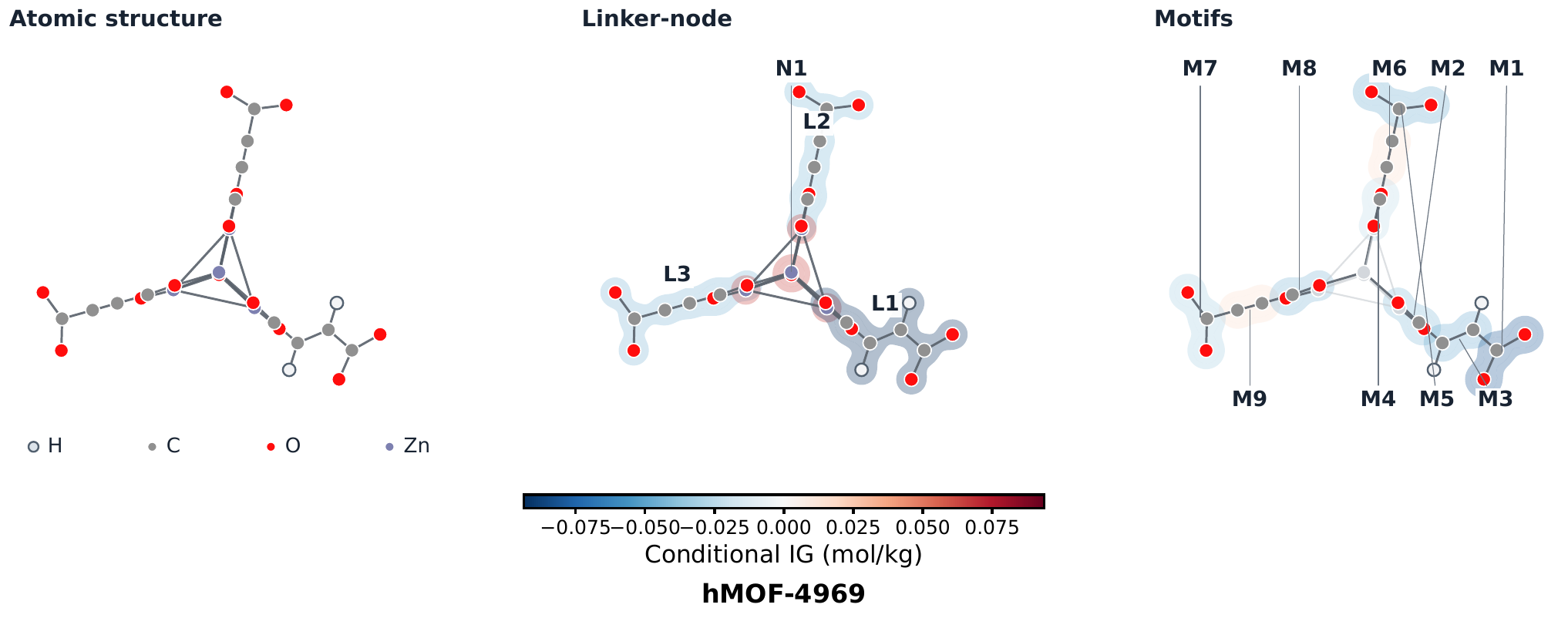}
\caption{Labeled ALIGNN attribution map for the main-text example
\texttt{hMOF-4969}. The uptake prediction error is below 0.1~mol/kg,
and all displayed scores meet the 5\% relative completeness criterion.
The finite projection preserves internal fragment bonds and omits six
interfragment periodic closure edges. Scores are reference-dependent
and nonadditive.}
\label{fig:si_hmof_examples}
\end{figure}

\clearpage
\bibliography{references}